%% file: paper.tex
\documentclass[10pt,twocolumn,letterpaper]{article}

\usepackage{cvpr}              %

\usepackage{graphicx}
\usepackage{amsmath}
\usepackage{amssymb}
\usepackage{booktabs}
\usepackage[ruled]{algorithm2e} %
\usepackage{animate}
\usepackage{caption}
\usepackage{url}
\usepackage{epsfig}
\usepackage{amsmath}
\usepackage{amssymb}
\usepackage{xcolor}
\usepackage{multirow}
\usepackage{siunitx}
\usepackage{mathtools}
\usepackage{subcaption}
\usepackage{soul}
\usepackage{balance}

\newcommand{\qheading}[1]{\noindent\textbf{#1}}

\usepackage{enumitem}

\usepackage[pagebackref,breaklinks,colorlinks]{hyperref}

\usepackage[capitalize]{cleveref}
\crefname{section}{Sec.}{Secs.}
\Crefname{section}{Section}{Sections}
\Crefname{table}{Table}{Tables}
\crefname{table}{Tab.}{Tabs.}

\def\cvprPaperID{8277} %
\def\confName{CVPR}
\def\confYear{2022}

\usepackage[symbol]{footmisc}

\begin{document}

\title{
    Neural 3D Video Synthesis from Multi-view Video
}

\vspace{-20pt}

\author{
Tianye Li$^{1,2,\ast}$ \qquad
Mira Slavcheva$^{2,\ast}$ \qquad
Michael Zollhoefer$^{2}$ \qquad
\\
Simon Green$^{2}$ \qquad
Christoph Lassner$^{2}$ \qquad
Changil Kim$^{3}$ \qquad
Tanner Schmidt$^{2}$ \qquad
\\
Steven Lovegrove$^{2}$ \qquad
Michael Goesele$^{2}$ \qquad
Richard Newcombe$^{2}$ \qquad
Zhaoyang Lv$^{2}$ \qquad
\\
$^1$University of Southern California \qquad
$^2$Reality Labs Research \qquad
$^3$Meta
}

\input{teaser_animated}

\input{sections/abstract}

\footnotetext{$\ast$ Equal contribution. TL's work was done during an internship at Reality Labs Research.}

\input{sections/introduction}

\input{sections/related_work_compact}

\input{sections/method}

\input{sections/experiments}

\input{sections/limitations}

\input{sections/conclusion}

\balance

{\small
\bibliographystyle{ieee_fullname}
\bibliography{bibliography}
}

\clearpage
\appendix

\input{appendix}

\end{document}

%% file: teaser_animated.tex
\twocolumn[{%
\renewcommand\twocolumn[1][]{#1}%
\maketitle

\vspace{-35pt}    %

\begin{center}
  \centering
  \animategraphics[
    autoplay,
    loop,
    every=1,             %
    width=\textwidth]
    {10}                  %
    {figures/teaseranim_depth_at_corner/img_} %
    {001}{040}           %
    \captionof{figure}{
        We propose a novel method for representing and rendering high quality 3D video.
        Our method trains a novel and compact dynamic neural radiance field (DyNeRF) in an efficient way.
        Our method demonstrates near photorealistic dynamic novel view synthesis for complex scenes including challenging scene motions and strong view-dependent effects.
        We demonstrate three synthesized 3D video, and show the associated high quality geometry in the heatmap visualization in each top right corner.
        \textbf{The embedded animations only play in Adobe Reader or KDE Okular. Please see the \href{https://neural-3d-video.github.io/resources/video.mp4}{full video} for the high-quality renderings and additional information}.
    }
  \label{fig:teaser}
    
\end{center}%
}]

\vspace{-35pt}

%% file: sections/abstract.tex
\begin{abstract}
\vspace{-5pt}  %

We propose a novel approach for 3D video synthesis that is able to represent multi-view video recordings of a dynamic real-world scene in a compact, yet expressive representation that enables high-quality view synthesis and motion interpolation.
Our approach takes the high quality and compactness of static neural radiance fields in a new direction: to a model-free, dynamic setting.
At the core of our approach is a novel time-conditioned neural radiance field that represents scene dynamics using a set of compact latent codes.
We are able to significantly boost the training speed and perceptual quality of the generated imagery by a novel hierarchical training scheme in combination with ray importance sampling.
Our learned representation is highly compact and able to represent a 10 second 30 FPS multi-view video recording by 18 cameras with a model size of only 28MB.
We demonstrate that our method can render high-fidelity wide-angle novel views at over 1K resolution, even for complex and dynamic scenes.
We perform an extensive qualitative and quantitative evaluation that shows that our approach outperforms the state of the art.
Project website: \url{https://neural-3d-video.github.io/}.
\end{abstract}

%% file: sections/introduction.tex
\section{Introduction}
Photorealistic representation and rendering of dynamic real-world scenes are highly challenging research topics, yet with many important applications that range from movie production to virtual and augmented reality.
Dynamic real-world scenes are notoriously hard to model using classical mesh-based representations, since they often contain thin structures, semi-transparent objects, specular surfaces, and topology that constantly evolves over time due to the often complex scene motion of multiple objects and people.

In theory, the 6D plenoptic function $P(\mathbf{x}, \mathbf{d}, t)$ is a suitable representation for this rendering problem, as it completely explains our visual reality and enables rendering every possible view at every moment in time~\cite{adelson1991plenoptic}.
Here, $\mathbf{x}\in\mathbb{R}^3$ is the camera position in 3D space, $\mathbf{d} = (\theta, \phi)$ is the viewing direction, and $t$ is time.
Thus, fully measuring the plenoptic function requires placing an omnidirectional camera at every position in space at every possible time.

Neural radiance fields (NeRF) \cite{Mildenhall20eccv} offer a way to circumvent this problem: instead of directly encoding the plenoptic function, they encode the radiance field of the scene in an implicit, coordinate-based function, which can be sampled through ray casting to approximate the plenoptic function.
However, the ray casting, which is required to train and to render a neural radiance field, involves hundreds of MLP evaluations for \emph{each} ray.
While this might be acceptable for a static snapshot of a scene, directly reconstructing a dynamic scene as a sequence of per-frame neural radiance fields would be prohibitive as both storage and training time increase linearly with time.
For example, to represent a $10$ second, \SI{30}{FPS} multi-view video recording by 18 cameras, which we later demonstrate with our method, a per-frame NeRF would require about \SI{15000}{GPU} hours in training and about \SI{1}{GB} in storage.
More importantly, such obtained representations would only reproduce the world as a discrete set of snapshots, lacking any means to reproduce the world in-between.
On the other hand, Neural Volumes \cite{Lombardi19tog} is able to handle dynamic objects and even renders at interactive frame rates.
Its limitation is the dense uniform voxel grid that limits resolution and/or size of the reconstructed scene due to the inherent $O(n^3)$ memory complexity.

In this paper, we propose a novel approach for 3D video synthesis of complex, dynamic real-world scenes that enables high-quality view synthesis and motion interpolation while being compact.
Videos typically consist of a time-invariant component under stable lighting and a continuously changing time-variant component.
This dynamic component typically exhibits locally correlated geometric deformations and appearance changes between frames.
By exploiting this fact, we propose to reconstruct a dynamic neural radiance field based on two novel contributions.

First, we extend neural radiance fields to the space-time domain.
Instead of directly using time as input, we parameterize scene motion and appearance changes by a set of compact latent codes.
Compared to the more obvious choice of an additional `time coordinate', the learned latent codes show more expressive power, allowing for recording the vivid details of moving geometry and texture.
They also allow for smooth interpolation in time, which enables visual effects such as slow motion or `bullet time'.
Second, we propose novel importance sampling strategies for dynamic radiance fields.
Ray-based training of neural scene representations treats each pixel as an independent training sample and requires thousands of iterations to go through all pixels observed from all views.
However, captured dynamic video often exhibits a small amount of pixel change between frames.
This opens up an opportunity to significantly boost the training progress by selecting the pixels that are most important for training.
Specifically, in the time dimension, we schedule training with coarse-to-fine hierarchical sampling in the frames.
In the ray/pixel dimension, our design tends to sample those pixels that are more time-variant than others.
These strategies allow us to shorten the training time of long sequences significantly, while retaining high quality reconstruction results.
We demonstrate our approach using a multi-view rig based on 18 GoPro cameras.
We show results on multiple challenging dynamic environments with highly complex view-dependent and time-dependent effects.
Compared to the na\"ive per-frame NeRF baseline, we show that with our combined temporal and spatial importance sampling we achieve one order of magnitude acceleration in training speed, with a model that is 40 times smaller in size for 10 seconds of a 30 FPS 3D video.
In summary we make the following contributions:
\begin{itemize}[nosep,align=parleft,leftmargin=*]
\item We propose a novel dynamic neural radiance field based on temporal latent codes that achieves high quality 3D video synthesis of complex, dynamic real-world scenes.
\item We present novel training strategies based on hierarchical training and importance sampling in the spatiotemporal domain, which boost training speed significantly and lead to higher quality results for longer sequences.
\item We provide our datasets of time-synchronized and calibrated multi-view videos that covers challenging 4D scenes for research purposes\footnote{\scriptsize\url{https://github.com/facebookresearch/Neural_3D_Video}}.

\end{itemize}

%% file: sections/related_work_compact.tex
\section{Related Work}
Our work is related to several research domains, such as novel view synthesis for static scenes, 3D video synthesis for dynamic scenes, image-based rendering, and neural rendering approaches.
For a detailed discussion of neural rendering applications and neural scene representations, we refer to the surveys \cite{tewari2020neuralrendering} and \cite{tewari2021advances}.

\qheading{Novel View Synthesis for Static Scenes.}
Novel view synthesis has been tackled by explicitly reconstructing textured 3D models of the scene and rendering from arbitrary viewpoints.
Multi-view stereo~\cite{furukawa2009accurate, schonberger2016pixelwise} and visual hull reconstructions~\cite{laurentini1994visual, esteban2004silhouette} have been successfully employed.
Complex view-dependent effects can be captured by light transport acquisition methods~\cite{debevec2000acquiring, wood2000surface}.
Learning-based methods have been proposed to relax the high number of required views and to accelerate the inference speed for geometry reconstruction~\cite{kar2017learning, yao2018mvsnet, gu2020cascade} and appearance capture~\cite{bi2020deep, meka2019deep}, or combined reconstruction techniques~\cite{yariv2020multiview, niemeyer2020differentiable}.
Novel view synthesis can also be achieved by reusing input image pixels.
Early works using this approach interpolate the viewpoints~\cite{chen1993view}.
The Light Field/Lumigraph method~\cite{gortler1996lumigraph,levoy1996light,davis2012unstructured,Overbeck18tog} resamples input image rays to generate novel views.
One drawback of these approaches is that it require dense sampling for high quality rendering of complex scenes.
More recently, \cite{flynn2019deepview, mildenhall2019local, srinivasan2019pushing, zhou2018stereo, kalantari2016learning} learn to fuse and resample pixels from reference views using neural networks.
Neural Radiance Fields (NeRFs) \cite{Mildenhall20eccv} train an MLP-based radiance and opacity field and achieve state-of-the-art quality for novel view synthesis.
Other approaches~\cite{meshry2019neural, wiles2020synsin} employ an explicit point-based scene representation combined with a screen space neural network for hole filling.
\cite{lassner2020pulsar} push this further and encode the scene appearance in a differentiable sphere-based representation.
\cite{sitzmann2019deepvoxels} employs a dense voxel grid of features in combination with a screen space network for view synthesis.
All these methods are excellent at interpolating views for static scenes, but it is unclear how to extend them to the dynamic setting.

\qheading{3D Video Synthesis for Dynamic Scenes.}
Techniques in this category enable view synthesis for dynamic scenes and might also enable interpolation across time.
For video synthesis, \cite{kanade1997virtualized} pioneers in showing the possibility of explicitly capture geometry and textures.
\cite{Zitnick2004high} proposes a temporal layered representation that can be compressed and replayed at an interactive rate. 
Reconstruction and animation is particularly well studied for humans~\cite{carranza2003free, starck2007surface, guo2019relightables}, but is usually performed model-based and/or only works with high-end capture setups.
\cite{li2012temporally} captures temporally consistent surfaces by tracking and completion.
\cite{Collet2015fvv} proposes a system for capturing and compressing streamable 3D video with high-end hardware.
More recently, learning-based methods such as \cite{huang2018dvv} achieve volumetric video capture for human performances from sparse camera views.
\cite{bansal20204d} focus on more general scenes. They decompose them into a static and dynamic component, re-project information based on estimated coarse depth, and employ a U-Net in screen space to convert the intermediate result to realistic imagery.
\cite{bemana2020xfields} uses a neural network for space-time and illumination interpolation.
\cite{yoon2020novel} uses a model-based step for merging the estimated depth maps to a unified representation that can be rendered from novel views.
Neural Scene Flow Fields \cite{li2021neural} incorporates a static background model.
Space-time Neural Irradiance Fields \cite{xian2021space} employs video depth estimation to supervise a space-time radiance field.
\cite{gao2021freeviewvideo} recently proposes a time-conditioned radiance field, supervised by its own predicted flow vectors.
These works have limited view angle due to their single-view setting and require additional supervision, such as depth or flow.
\cite{park2021nerfies, pumarola2020d, tretschk2021nonrigid, du2021neural}
explicitly model dynamic scenes by a warp field or velocity field to deform a canonical radiance field.
STaR \cite{yuan2021starcvpr} models scenes of rigidly moving objects using several canonical radiance fields that are rigidly transformed.
These methods cannot model challenging dynamic events such as topology changes.
Several radiance field approaches have been proposed for modeling digital humans \cite{gafni2021dynamic, Raj2021PixelalignedVA, peng2021neural, 2021narf, liu2021neural}, but they can not directly be applied to general non-rigid scenes.
Furthermore, there have been efforts in improving neural radiance fields for in-the-wild scenes \cite{martinbrualla2020nerfw},
generalization across scenes.
HyperNeRF~\cite{park2021hypernerf} is a concurrent work on dynamic novel view synthesis, but they focus on monocular video in a short sequence.
Neural Volumes~\cite{Lombardi19tog} employs volume rendering in combination with a view-conditioned decoder network to parameterize dynamic sequences of single objects.
Their results are limited in resolution and scene complexity due to the inherent $O(n^3)$ memory complexity.
\cite{broxton2020siggraph} enable 6DoF video for VR applications based on independent alpha-textured meshes that can be streamed at the rate of hundreds of Mb/s.
This approach employs a capture setup with 46 cameras and requires a large training dataset to construct a strong scene-prior.
In contrast, we seek a unified space-time representation that enables continuous viewpoint and time interpolation, while being able to represent an entire multi-view video sequence of $10$ seconds in as little as $28$MB.

%% file: sections/method.tex
\newcommand{\DyNeRF}{DyNeRF}

\section{\protect\DyNeRF: Dynamic Neural Radiance Fields}
\label{representation}

We address the problem of reconstructing dynamic 3D scenes from time-synchronized multi-view videos with known intrinsic and extrinsic parameters.
The representation we aim to reconstruct from such multi-camera recordings should allow us to render photorealistic images from a wide range of viewpoints at arbitrary points in time.

Building on NeRF~\cite{Mildenhall20eccv}, we propose \emph{dynamic neural radiance fields (DyNeRF)} that are directly optimized from input videos captured with multiple video cameras. 
DyNeRF is a novel continuous space-time neural radiance field representation, controllable by a series of temporal latent embeddings that are jointly optimized during training.
Our representation compresses a huge volume of input videos from multiple cameras to a compact 6D representation that can be queried continuously in both space and time.
The learned embedding faithfully captures detailed temporal variations of the scene, such as complex photometric and topological changes, without explicit geometric tracking.

\input{figures/fig_network}

\input{sections/method_representation}

\input{sections/method_importance_sampling}

%% file: figures/fig_network.tex
\begin{figure}[t]

\centering
\includegraphics[width=1\linewidth]{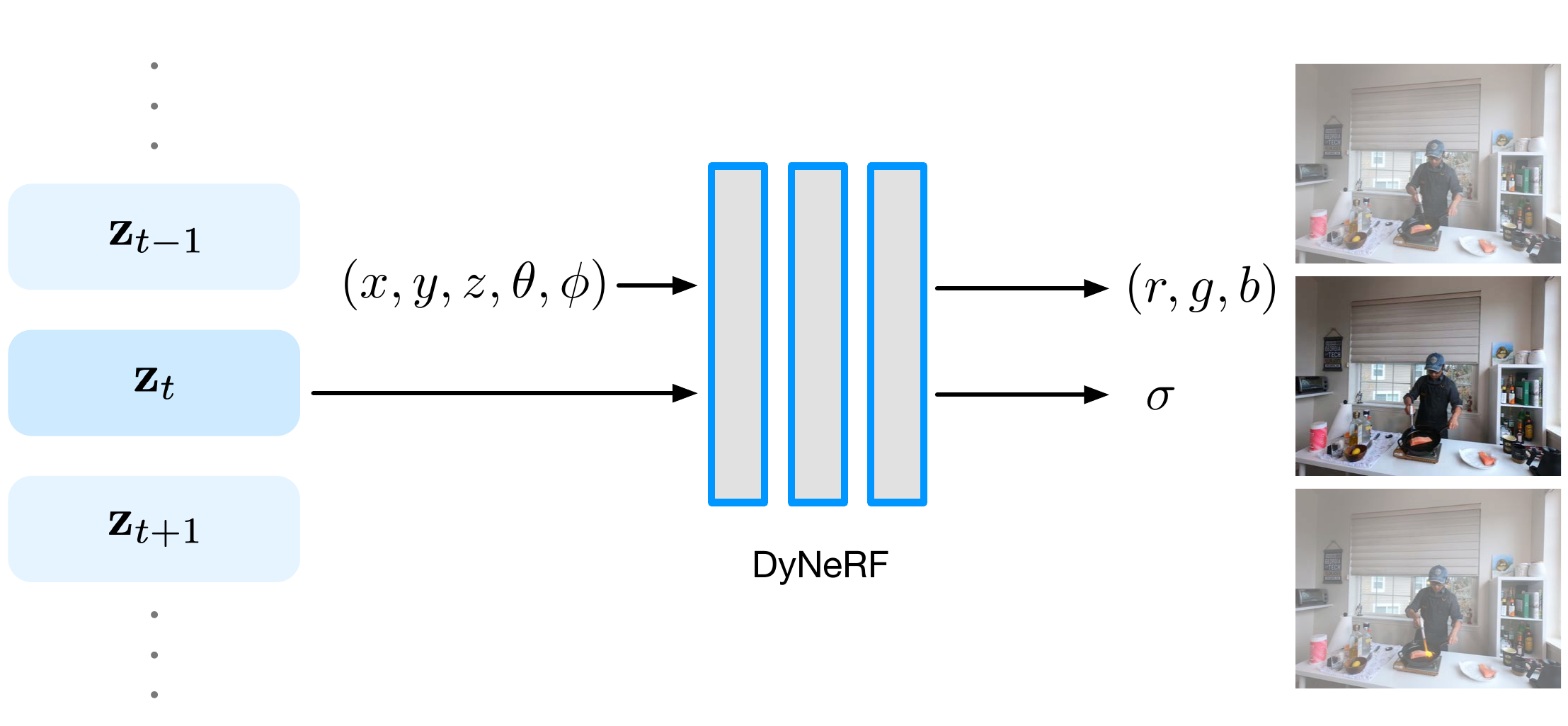}
\vspace{-0.8cm}
\caption{
We learn the 6D plenoptic function by our novel dynamic neural radiance field (DyNeRF) that conditions on position, view direction and a compact, yet expressive time-variant latent code.  
}
\label{fig:network}

\end{figure}

%% file: sections/method_representation.tex
\subsection{Representation}

The problem of representing 3D video comprises learning the 6D plenoptic function that maps a 3D position $\mathbf{x} \in \mathbb{R}^3$, direction $\mathbf{d} \in \mathbb{R}^2$, and time $t \in \mathbb{R}$, to RGB radiance $\mathbf{c} \in \mathbb{R}^3$ and opacity $\sigma \in \mathbb{R}$.
Based on NeRF \cite{Mildenhall20eccv}, which approximates the 5D plenoptic function of a static scene with a learnable function, a potential solution would be to add a time dependency to the function:
\begin{equation}
F_{\Theta}:
(
\mathbf{x},
\mathbf{d},
t
)
\longrightarrow
(
\mathbf{c},
\sigma
) \enspace ,
\label{eq:Nerf-T}
\end{equation}
which is realized by a Multi-Layer Perceptron (MLP) with trainable weights~$\Theta$.
The $1$-dimensional time variable $t$ can be mapped via positional encoding \cite{tancik2020fourier} to a higher dimensional space, in a manner similar to how NeRF handles the inputs $\mathbf{x}$ and $\mathbf{d}$.
However, we empirically found that it is challenging for this design to capture complex dynamic 3D scenes with challenging topological changes and time-dependent volumetric effects, such as flames.

\qheading{\textbf{Dynamic Neural Radiance Fields}.}
We model the dynamic scene by time-variant latent codes $\mathbf{z}_{t} \in \mathbb{R}^{D}$, as shown in Fig.~\ref{fig:network}.
We learn a set of time-dependent latent codes, indexed by a discrete time variable $t$:
\begin{equation}
F_{\Theta}:
(
\mathbf{x},
\mathbf{d},
\mathbf{z}_{t}
)
\longrightarrow
(
\mathbf{c},
\sigma
) \enspace .
\end{equation}
The latent codes provide a compact representation of the state of a dynamic scene at a certain time, which can handle various complex scene dynamics, including deformation, topological and radiance changes.
We apply positional encoding~\cite{tancik2020fourier} to the input position coordinates to map them to a higher-dimensional vector.
However, no positional encoding is applied to the time-dependent latent codes.
Before training, the latent codes $\{\mathbf{z}_{t}\}$ are randomly initialized independently across all frames.

\qheading{\textbf{Rendering}.}
We use volume rendering techniques to render the radiance field given a query view in space and time.
Given a ray $\mathbf{r}(s) = \mathbf{o} + s\mathbf{d}$ with the origin $\mathbf{o}$ and direction $\mathbf{d}$ defined by the specified camera pose and intrinsics, the rendered color of the pixel corresponding to this ray $\mathbf{C}(\mathbf{r})$ is an integral over the radiance weighted by accumulated opacity~\cite{Mildenhall20eccv}:
\begin{equation}
\mathbf{C}^{(t)}(\mathbf{r})
=
\int_{s_n}^{s_f}
T(s)
\sigma(\mathbf{r}(s), \mathbf{z}_t)
\mathbf{c}(\mathbf{r}(s), \mathbf{d}, \mathbf{z}_t)) \,ds \,
\enspace .
\end{equation}
where $s_n$ and $s_f$ denote the bounds of the volume depth range and the accumulated opacity $T(s) = \text{exp}(
- \int_{s_n}^{s} 
\sigma(\mathbf{r}(p), \mathbf{z}_t))\,dp
).
$
We apply a hierarchical sampling strategy as ~\cite{Mildenhall20eccv} with stratified sampling on the coarse level followed by importance sampling on the fine level.

\qheading{\textbf{Loss Function}.}
The network parameters $\Theta$ and the latent codes $\{\mathbf{z}_{t}\}$ are simultaneously trained by minimizing the $\ell_2$-loss between the rendered colors $\hat{\mathbf{C}}(\mathbf{r})$ and the ground truth colors $\mathbf{C}(\mathbf{r})$, and summed over all rays $\mathbf{r}$ that correspond to the image pixels from all training camera views $\mathcal{R}$ and throughout all time frames $t \in \mathcal{T}$ of the recording:
\begin{equation}
\mathcal{L} = 
\sum_{t \in \mathcal{T}, \mathbf{r} \in \mathcal{R}}
\sum_{j \in \{c, f\}}
\left\|
\hat{\mathbf{C}}_{j}^{(t)}(\mathbf{r})
-
\mathbf{C}^{(t)}(\mathbf{r})
\right\|^{2}_{2}
\enspace .
\end{equation}
We evaluate the loss at both the coarse and the fine level, denoted by $\hat{\mathbf{C}}_{\text{c}}^{(t)}$ and $\hat{\mathbf{C}}_{\text{f}}^{(t)}$ respectively, similar to NeRF.
We train with a stochastic version of this loss function, by randomly sampling ray data and optimizing the loss of each ray batch.
Please note that our dynamic radiance field is trained with this plain $\ell_2$-loss without any special regularization.

%% file: sections/method_importance_sampling.tex
\subsection{Efficient Training}

\input{figures/fig_importance_sampling}

An additional challenge of ray casting--based neural rendering on video data is the large amount of training time required.
The number of training iterations per epoch scales linearly with the total number of pixels in the input multi-view videos.
For a 10 second, 30 FPS, 1 MP multi-view video sequence from 18 cameras, there are about 7.4 billion ray samples in one epoch, which would take about half a week to process using 8 NVIDIA Volta class GPUs.
Given that each ray needs to be re-visited several times to obtain high quality results, this sampling process is one of the biggest bottlenecks for ray-based neural reconstruction methods to train 3D videos at scale.

However, for a natural video a large proportion of the dynamic scene is either time-invariant or 
only contains a small time-variant radiance change at a particular timestamp across the entire observed video. Hence, uniformly sampling rays causes an imbalance between time-invariant observations and time-variant ones. This means it is highly inefficient \emph{and} impacts reconstruction quality: time-invariant regions reach high reconstruction quality sooner and are uselessly oversampled, while time-variant regions require additional sampling, increasing the training time.

To explore temporal redundancy in the context of 3D video, we propose two strategies to accelerate the training process (see Fig.~\ref{fig:importance_sampling_viz}):
(1) hierarchical training that optimizes data over a coarse-to-fine frame selection and (2) importance sampling that prefers rays around regions of higher temporal variance.
In particular, these strategies form a different loss function by paying more attention to the ``important'' rays in time frame set $\mathcal{S}$ and pixel set $\mathcal{I}$ for training:
\begin{equation}
\mathcal{L}_{\text{efficient}} = 
\sum_{t \in \mathcal{S}, \mathbf{r} \in \mathcal{I}}
\sum_{j \in \{c, f\}}
\left\|
\hat{\mathbf{C}}_{j}^{(t)}(\mathbf{r})
-
\mathbf{C}^{(t)}(\mathbf{r})
\right\|^{2}_{2}
\enspace .
\end{equation}
These two strategies combined can be regarded as an adaptive sampling approach,
contributing to significantly faster training and improved rendering quality.

\qheading{Hierarchical Training.} %
Instead of training DyNeRF on all video frames, we first train it on keyframes, which we sample all images equidistantly at fixed time intervals~$K$, i.e. $\mathcal{S} = \{t \mid t = nK, n \in \mathbb{Z}^{+}, t\in \mathcal{T}\}$. 
Once the model converges with keyframe supervision, we use it to initialize the final model, which has the same temporal resolution as the full video.
Since the per-frame motion of the scene within each segment (divided by neighboring keyframes) is smooth, we initialize the fine-level latent embeddings by linearly interpolating between the coarse embeddings.
Finally, we train using data from all the frames jointly, $\mathcal{S} = \mathcal{T}$, further optimizing the network weights and the latent embeddings.
The coarse keyframe model has already captured an approximation of the time-invariant information across the video.  %
Therefore, the fine full-frame training only needs to learn the time-variant information per-frame.

\qheading{Ray Importance Sampling.} We propose to sample rays $\mathcal{I}$ across time with different importance based on the temporal variation in the input videos. For each observed ray $\mathbf{r}$ at time $t$, we compute a weight $\omega^{(t)}(\mathbf{r})$.
In each training iteration we pick a time frame $t$ at random. 
We first normalize the weights of the rays across all input views for frame $t$, and then apply inverse transform sampling to select rays based on these weights.

To calculate the weight of each ray, we propose three implementations based on different insights. 
\begin{itemize}[nosep,align=parleft,leftmargin=*]
\item \textbf{Global-Median (DyNeRF-ISG)}: We compute the weight of each ray based on the residual difference of its color to its the global median value across time.
\item \textbf{Temporal-Difference (DyNeRF-IST)}:
We compute the weight of each ray based on the color difference in two consecutive frames.
\item \textbf{Combined Method (DyNeRF-IS$^\star\!$}): Combine both strategies above.
\end{itemize}

We empirically observed that training DyNeRF-ISG with a high learning rate leads to very quick recovery of dynamic detail, but results in some jitter across time.
On the other hand, training DyNeRF-IST with a low learning rate produces a smooth temporal sequence which is still somewhat blurry.
Thus, we combine the benefits of both methods in our final strategy, DyNeRF-IS$^\star$ (referred as DyNeRF in later sections), which first obtains sharp details via DyNeRF-ISG and then smoothens the temporal motion via DyNeRF-IST.
We explain the details of the three strategies in the \textit{Supp. Mat.}
All importance sampling methods assume a static camera rig.

%% file: figures/fig_importance_sampling.tex
\begin{figure}[t]
\centering
\includegraphics[width=0.95\linewidth]{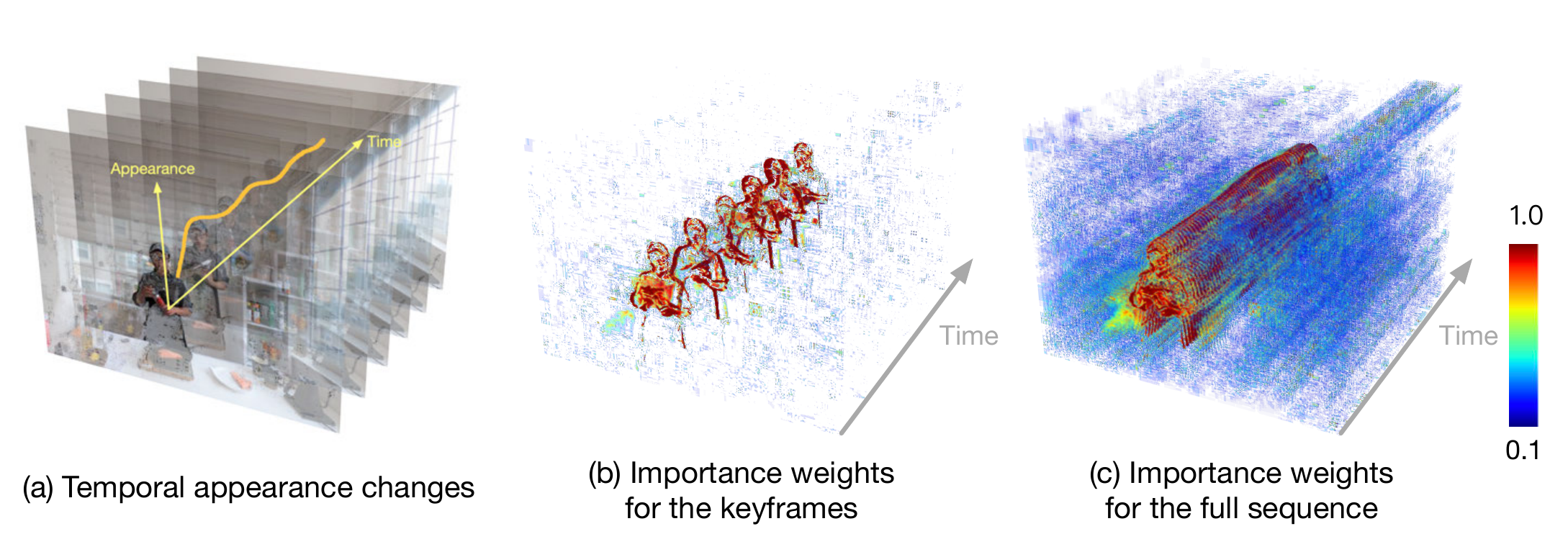}
\caption{\textbf{Overview of our efficient training strategies.} We perform hierarchical training first using keyframes (b) and then on the full sequence (c). At both stages, we apply the ray importance sampling technique to focus on the rays with high time-variant information based on weight maps that measure the temporal appearance changes (a). We show a visualized example of the sampling probability based on global median map using a heatmap (red and opaque means high probability).
}
\label{fig:importance_sampling_viz}
\end{figure}

%% file: sections/experiments.tex
\section{Experiments}

We demonstrate our approach on a large variety of captured daily events with challenging scene motions, varying illuminations and self-cast shadows, view-dependent appearances and highly volumetric effects.
We performed detailed ablation studies and comparisons to various baselines on our multi-view data and immersive video data \cite{broxton2020siggraph}.

\qheading{\textbf{Supplementary materials.}} We strongly recommend the reader to watch our \textit{supplemental video} to better judge the photorealism of our approach at high resolution, which cannot be represented well by the metrics.
We demonstrate interactive playback of our 3D videos in commodity VR headset \textit{Quest 2} in the \textit{supplemental video}.
We further provide comprehensive details of our capture setup, dataset descriptions, comparison settings, more ablations studies on parameter choices and failure case discussions.

\input{sections/experiments_setting}

\input{sections/experiments_results}

%% file: sections/experiments_setting.tex
\subsection{Evaluation Settings}

\qheading{\textbf{Plenoptic Video Datasets.}} 
We build a mobile multi-view capture system using 21 GoPro Black Hero 7 cameras.
We capture videos at a resolution of $2028 \times 2704$ (2.7K) and frame rate of \SI{30}{FPS}.
The multi-view inputs are time-synchronized. We obtain the camera intrinsic and extrinsic parameters using COLMAP \cite{schoenberger2016sfm}.
We employ 18 views for training, and 1 view for qualitative and quantitative evaluations for all datasets except one sequence observing multiple people moving, which uses 14 training views.
For more details on the capture setup, please refer to the \textit{Supp. Mat.}

Our captured data demonstrates a variety of challenges for video synthesis, including
(1) objects of high specularity, translucency and transparency,
(2) scene changes and motions with changing topology (poured liquid),
(3) self-cast moving shadows,
(4) volumetric effects (fire flame),
(5) an entangled moving object with strong view-dependent effects (the torch gun and the pan), 
(6) various lighting conditions (daytime, night, spotlight from the side), and
(7) multiple people moving around in open living room space with outdoor scenes seen through transparent windows with relatively dark indoor illumination.
Our collected data can provide sufficient synchronized camera views for high quality 4D reconstruction of challenging dynamic objects and view-dependent effects in a natural daily indoor environment, which, to our knowledge, did not exist in public 4D datasets.
We will release the datasets for research purposes.
\qheading{\textbf{Immersive Video Datasets.}} We also demonstrate the generality of our method using the multi-view videos from \cite{broxton2020siggraph} directly trained on their fisheye video input. %

\input{figures/fig_results}

\qheading{\textbf{Baselines.}} 
We compare to the following baselines:
\begin{itemize}[nosep,align=parleft,leftmargin=*]
\itemsep0em 
\item \textbf{Multi-View Stereo (MVS)}:
frame-by-frame rendering of the reconstructed and textured 3D meshes using commercial software RealityCapture\footnote{\url{https://www.capturingreality.com/}}.
\item \textbf{Local Light Field Fusion (LLFF)} \cite{mildenhall2019local}:
frame-by-frame rendering of the LLFF-produced multiplane images with the pretrained model\footnote{\url{https://github.com/Fyusion/LLFF}}.
\item \textbf{NeuralVolumes (NV)}\cite{Lombardi19tog}: 
One prior-art volumetric video rendering method using a warped canonical model. We follow the same setting as the original paper.
\item \textbf{NeRF-T}:
a temporal NeRF baseline as described in Eq.~\ref{eq:Nerf-T}.
\item \textbf{DyNeRF$^\dagger$}:
An ablation setting of DyNeRF without our proposed hierarchical training and importance sampling.

\end{itemize}
Due to page limit, we provide more ablation analysis of our importance sampling strategies and latent code dimension in Supp. Mat.

\qheading{\textbf{Metrics}.}
We evaluate the rendering quality on test view and the following quantitative metrics:
(1) Peak signal-to-noise ratio (PSNR);
(2) Mean square error (MSE);
(3) Structural dissimilarity index measure (DSSIM) \cite{upchurch2016dssim,sara2019dssim}
(4) Perceptual quality measure LPIPS \cite{zhang2018unreasonable};
(5) Perceived error difference FLIP \cite{andersson2020flip};
(6) Just-Objectionable-Difference (JOD) \cite{Mantiuk2021tog}.
Higher PSNR and scores indicate better reconstruction quality and higher JOD represents less visual difference compared to the reference video.
For all other metrics, lower numbers indicate better quality.

For any video of length shorter than 60 frames, we evaluate the model frame-by-frame on the complete video. 
Considering the significant amount required for high resolution rendering, we evaluate the model every 10 frames to calculate the frame-by-frame metrics reported for any video of length equal or longer than 300 frames in Tab.~\ref{tab:ablations_methods_wo_is}. 
For video metric JOD which requires a stack of continuous video frames, we evaluate the model on the whole sequence reported in Tab.~\ref{tab:video_eval_methods}.
We verified on 2 video sequences with a frame length of 300 that the PSNR differs by at most $0.02$ comparing evaluating them every $10$th frame vs.~on all frames.
We evaluate all the models at $1$K resolution, and report the average of the result from every evaluated frame.

\qheading{\textbf{Implementation Details.}}
We implement our approach in PyTorch.
We use the same MLP architecture as in NeRF \cite{Mildenhall20eccv} except that we use $512$ activations for the first 8 MLP layers instead of $256$.
We employ $1024$-dimensional latent codes.
In the hierarchical training we first only train on keyframes that are $K=30$ frames apart.
We employ the Adam optimizer \cite{Kingma2015Adam} with parameters $\beta_1 = 0.9$ and $\beta_2 = 0.999$.
In the keyframe training stage, we set a learning rate of $5\mathrm{e}{-4}$ and train for $300K$ iterations.
We include the details on the important sampling scheme in the Supp. Mat.
We set the latent code learning rate to be $10\times$ higher than for the other network parameters.
The per-frame latent codes are initialized from $\mathcal{N}(0, \frac{0.01}{\sqrt{D}})$, where $D=1024$. 
The total training takes about a week with 8 NVIDIA V100 GPUs and a total batch size of 24576 rays.

%% file: figures/fig_results.tex
\begin{figure}[t]
\centering

\begin{subfigure}[t]{\linewidth}

    \includegraphics[width=0.25\linewidth]{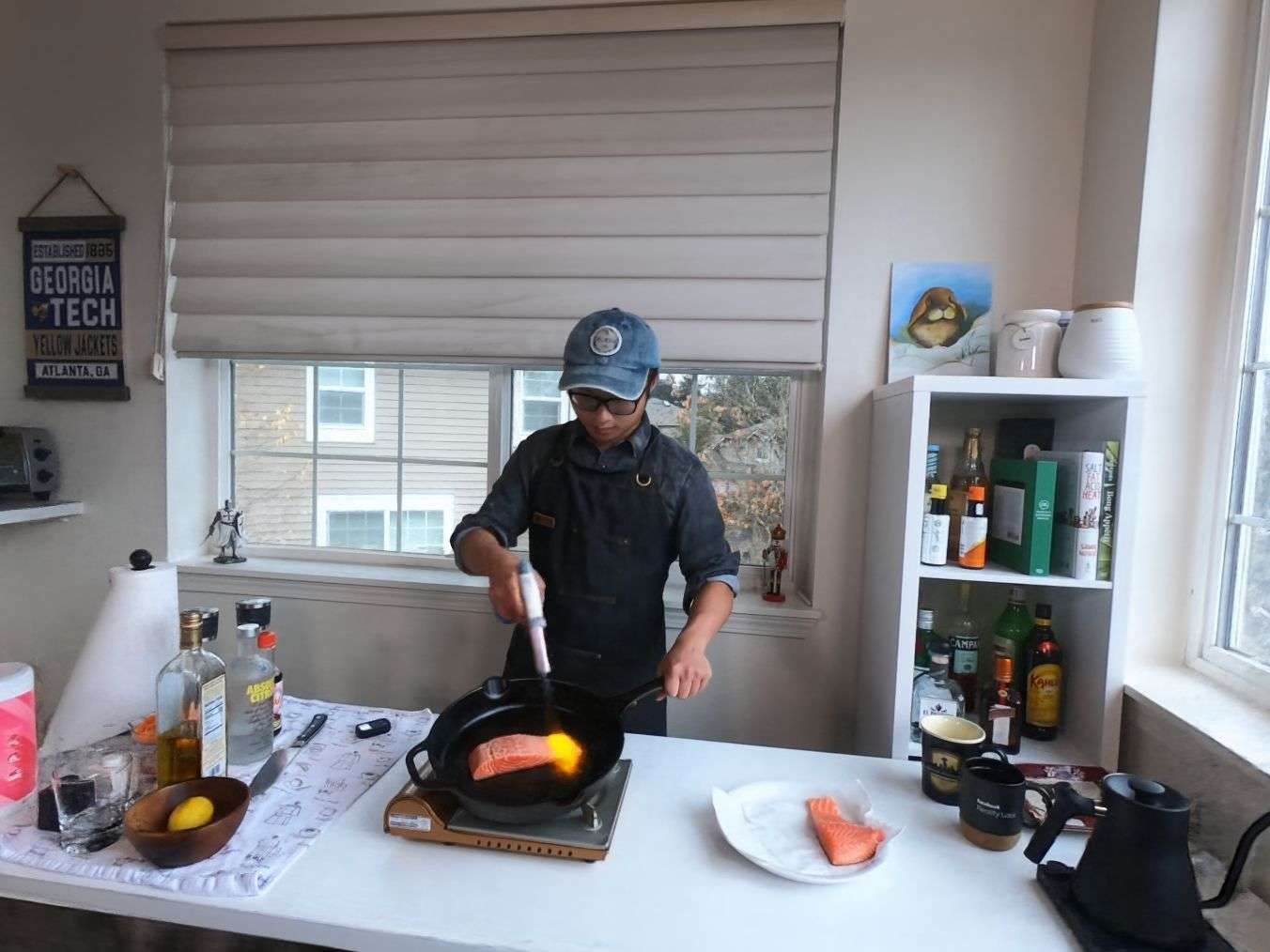}~
    \includegraphics[width=0.25\linewidth]{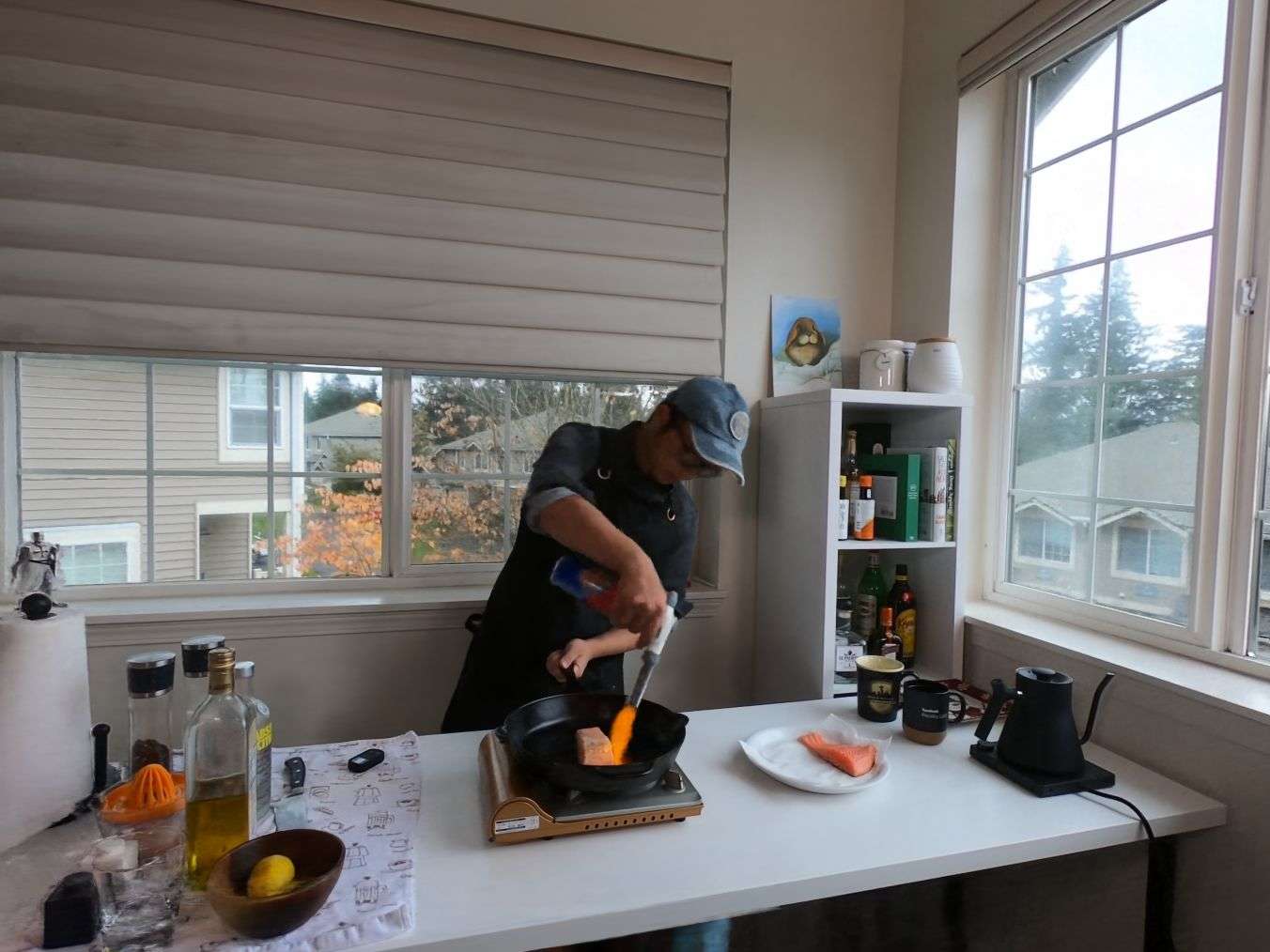}~
    \includegraphics[width=0.25\linewidth]{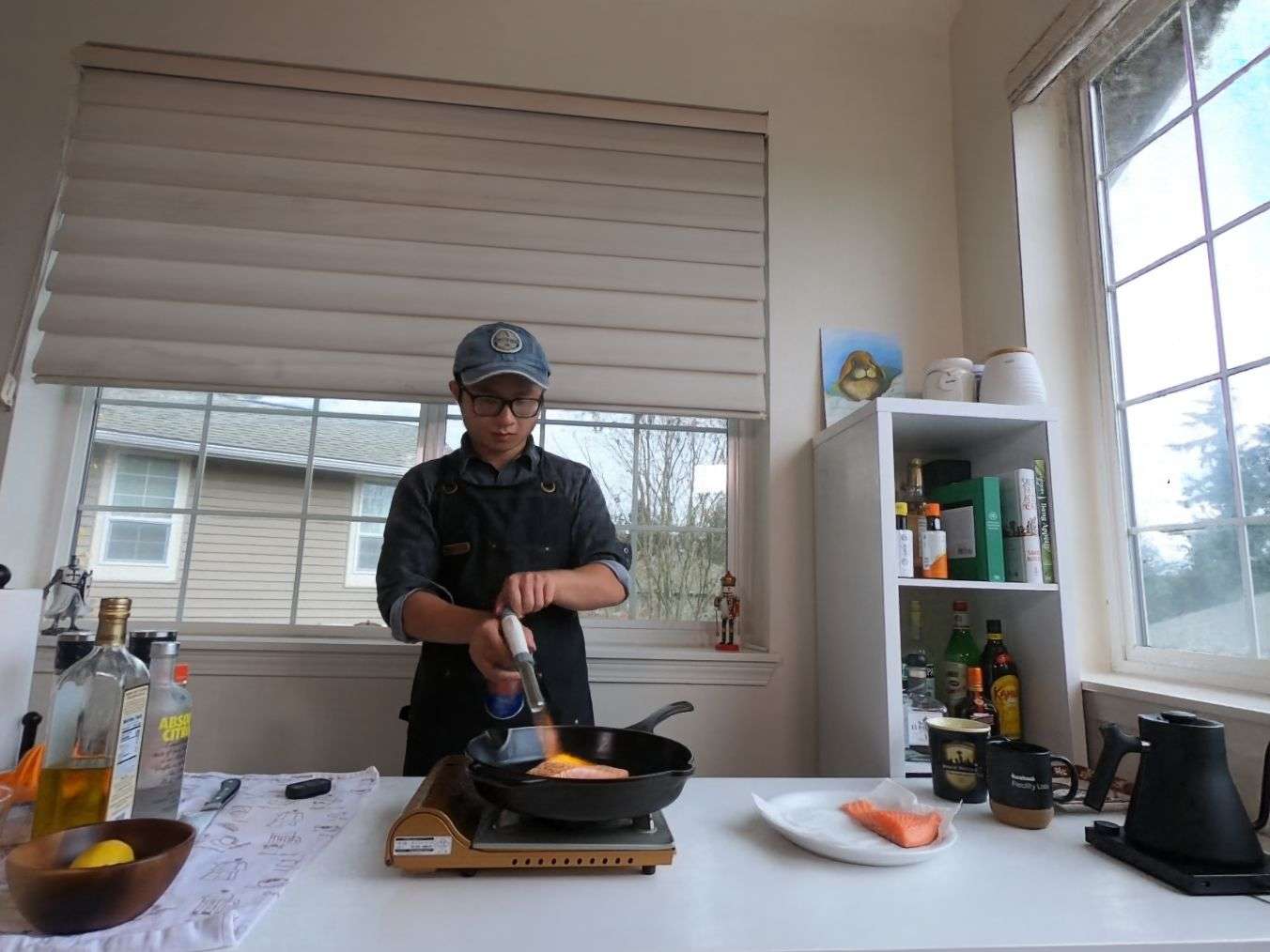}~
    \includegraphics[width=0.25\linewidth]{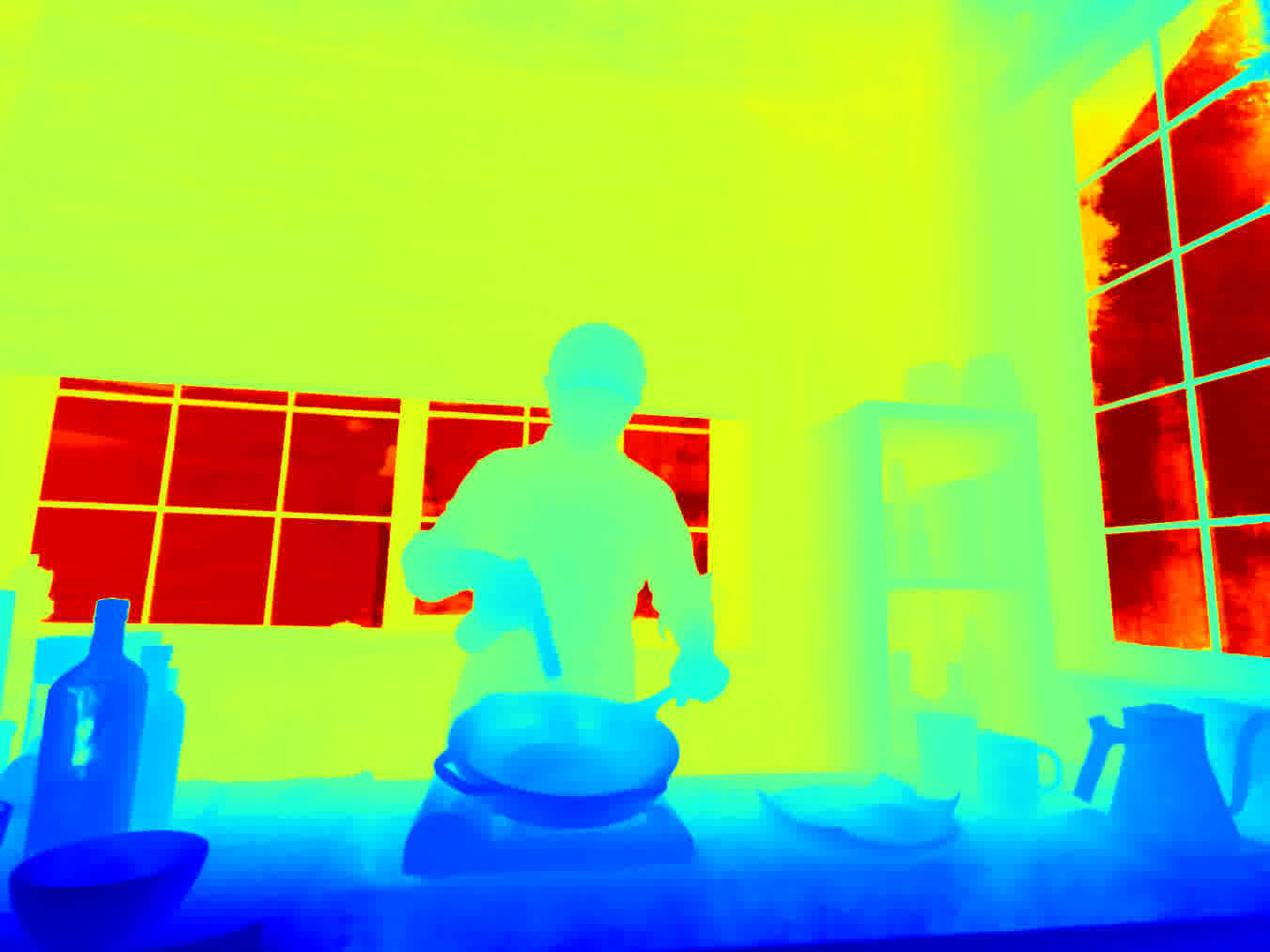}~ \\
    \includegraphics[width=0.25\linewidth]{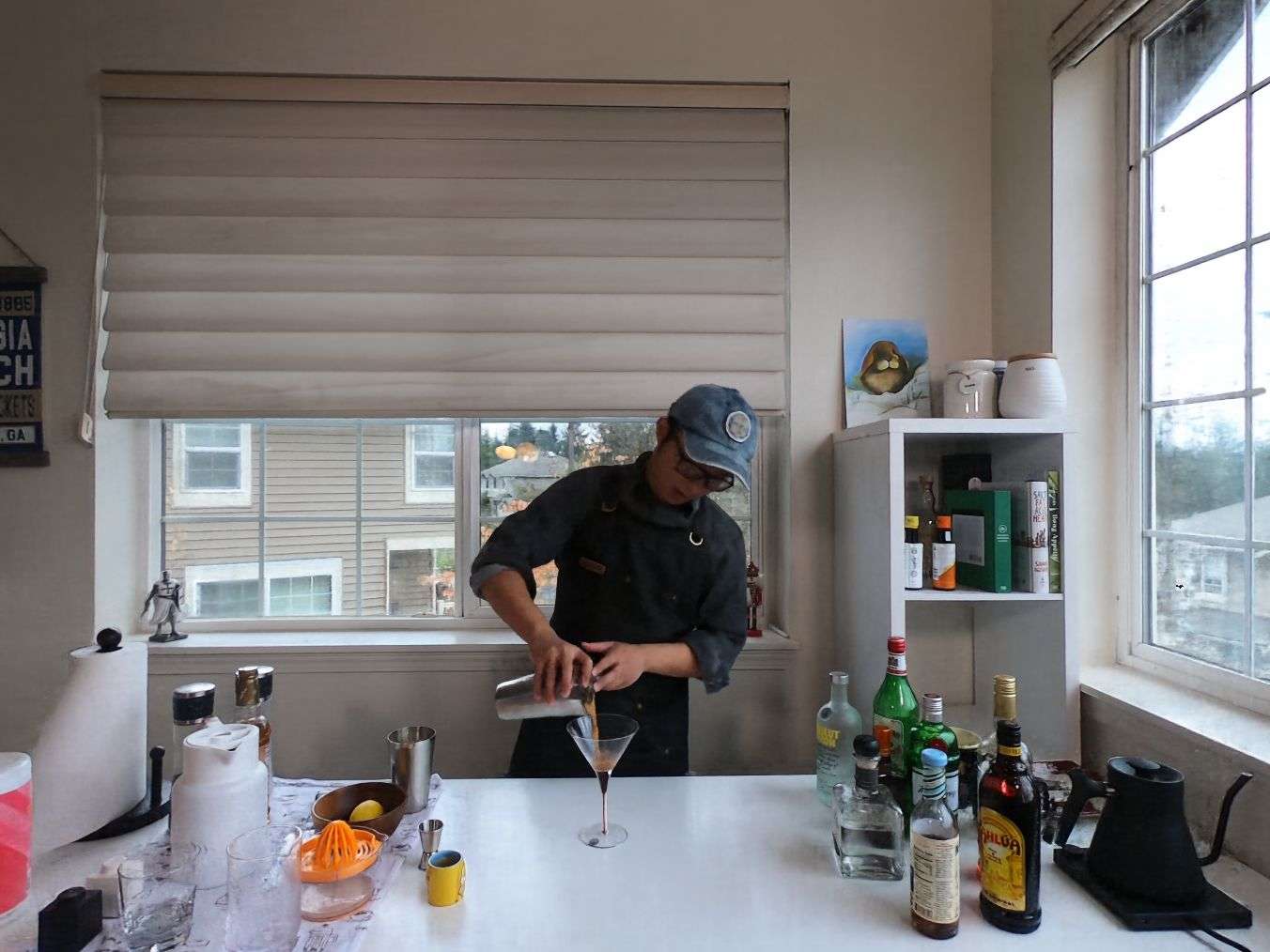}~
    \includegraphics[width=0.25\linewidth]{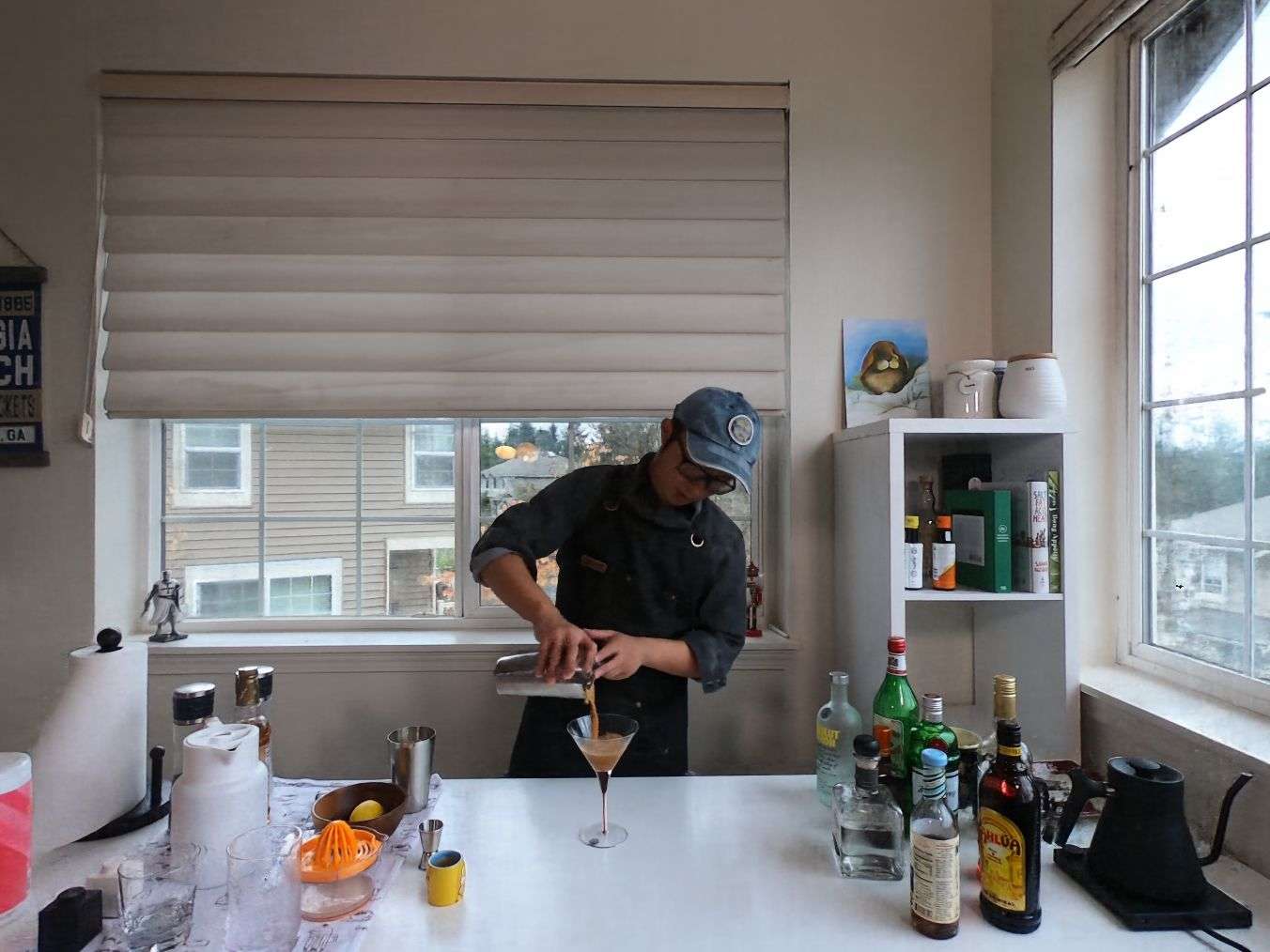}~
    \includegraphics[width=0.25\linewidth]{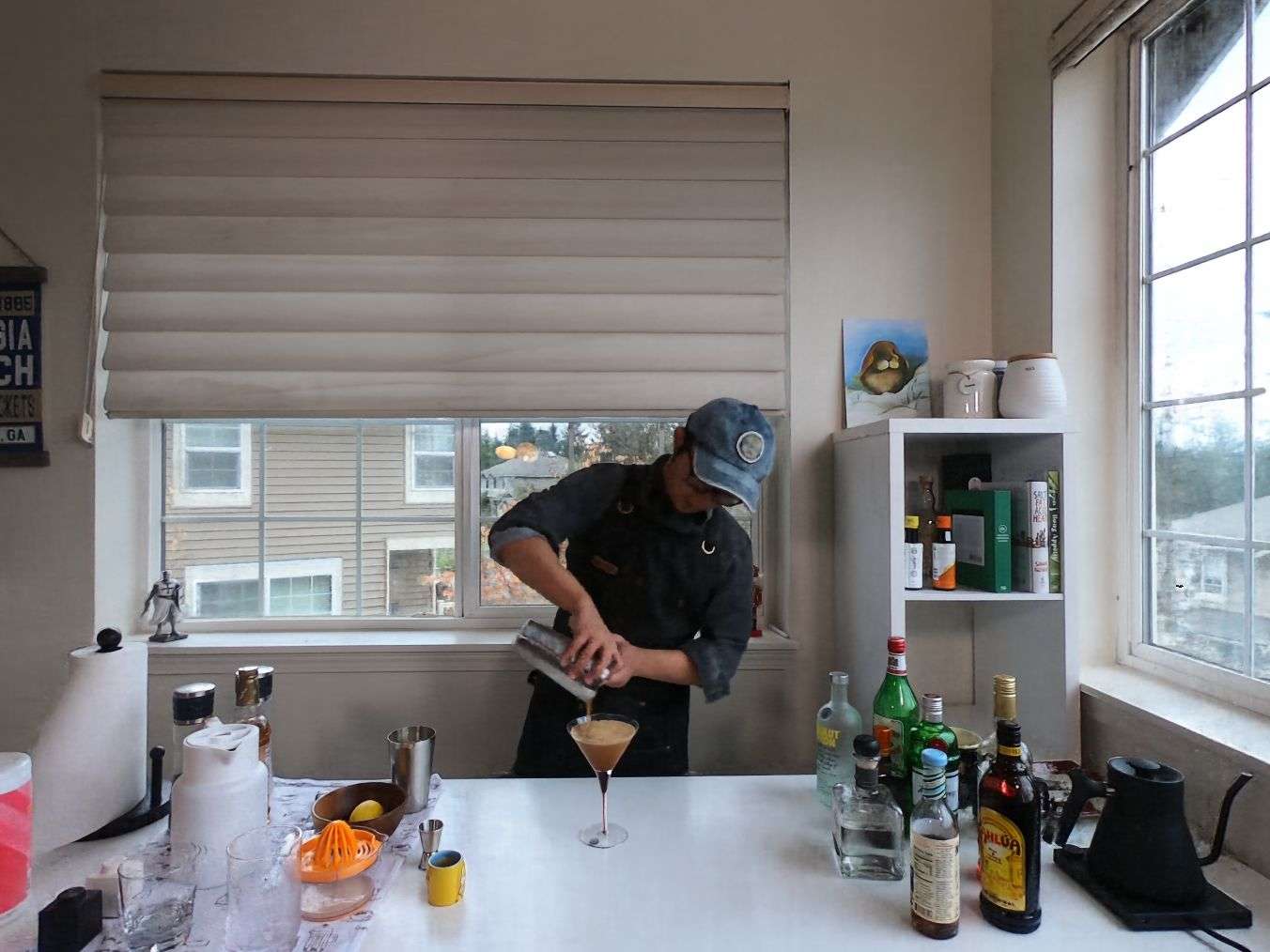}~
    \includegraphics[width=0.25\linewidth]{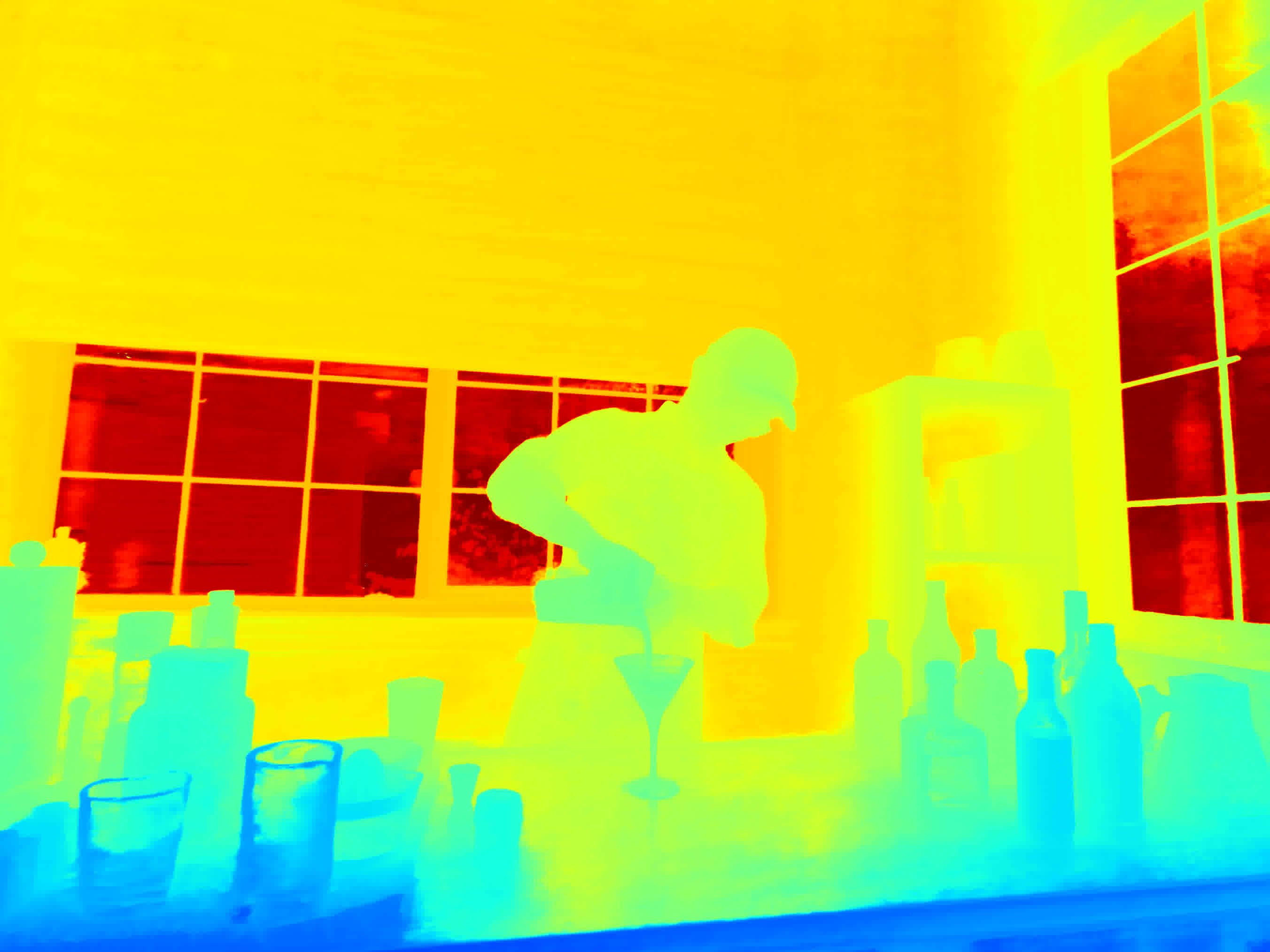}~ \\
    \includegraphics[width=0.25\linewidth]{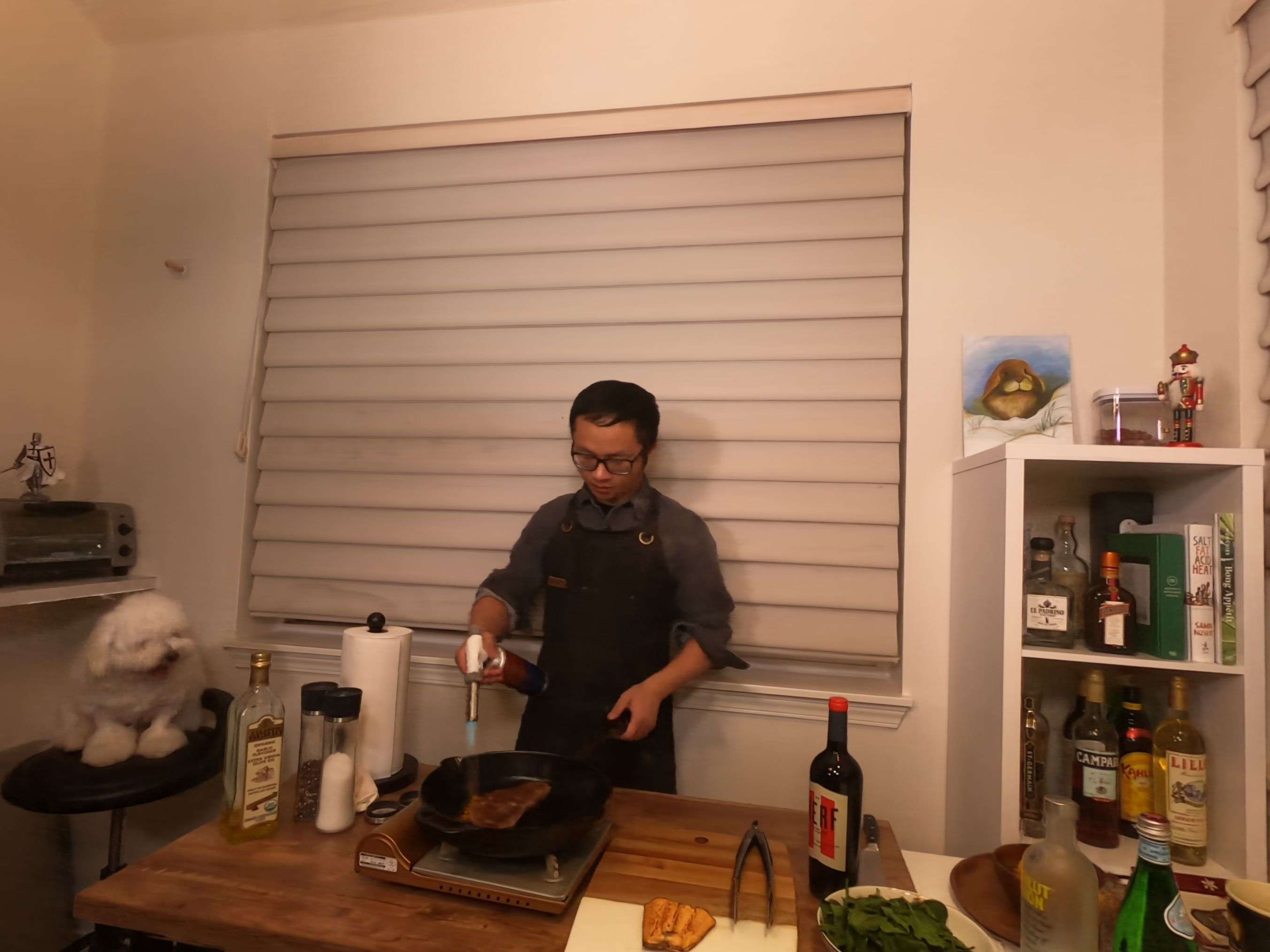}~
    \includegraphics[width=0.25\linewidth]{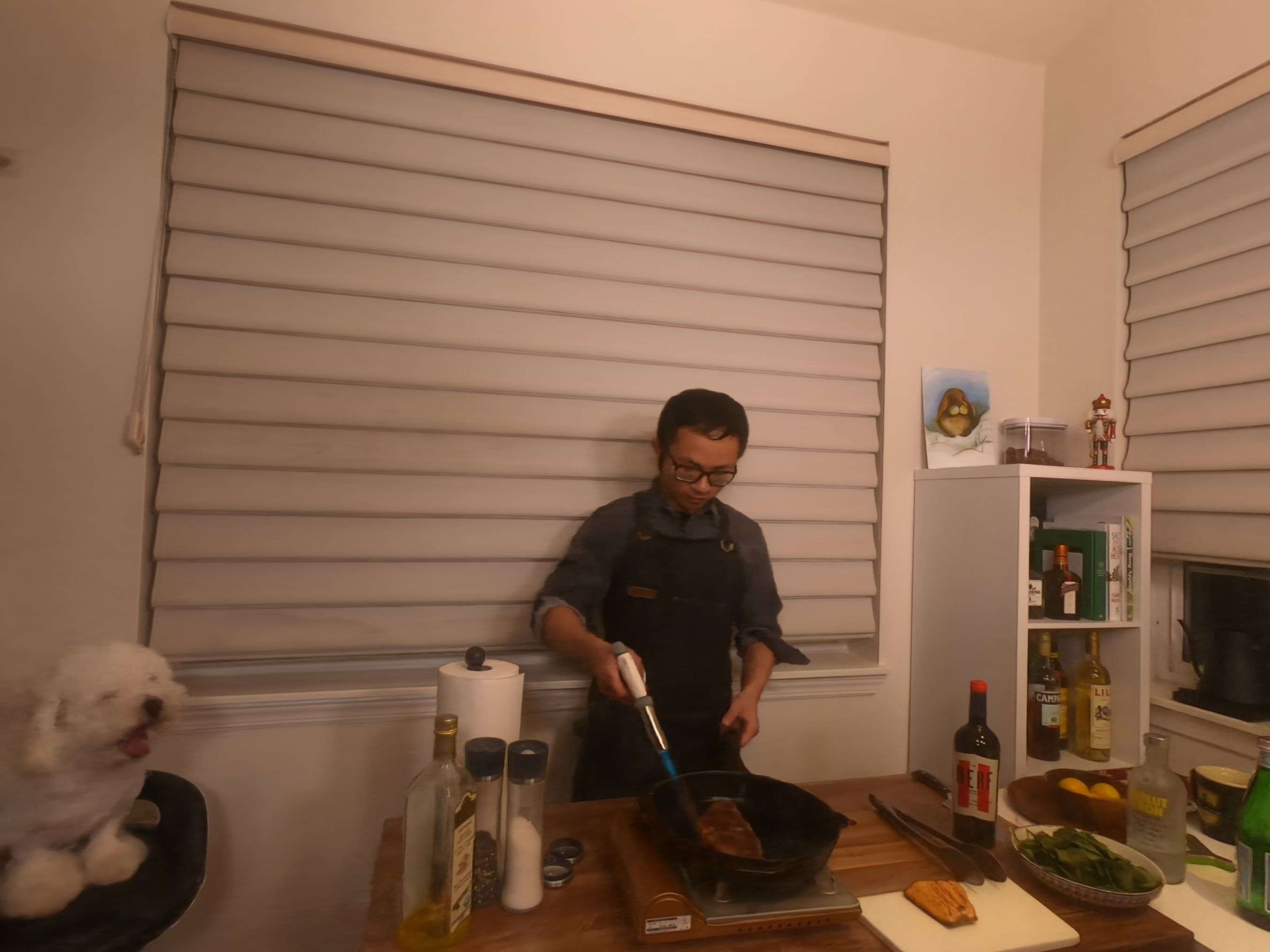}~
    \includegraphics[width=0.25\linewidth]{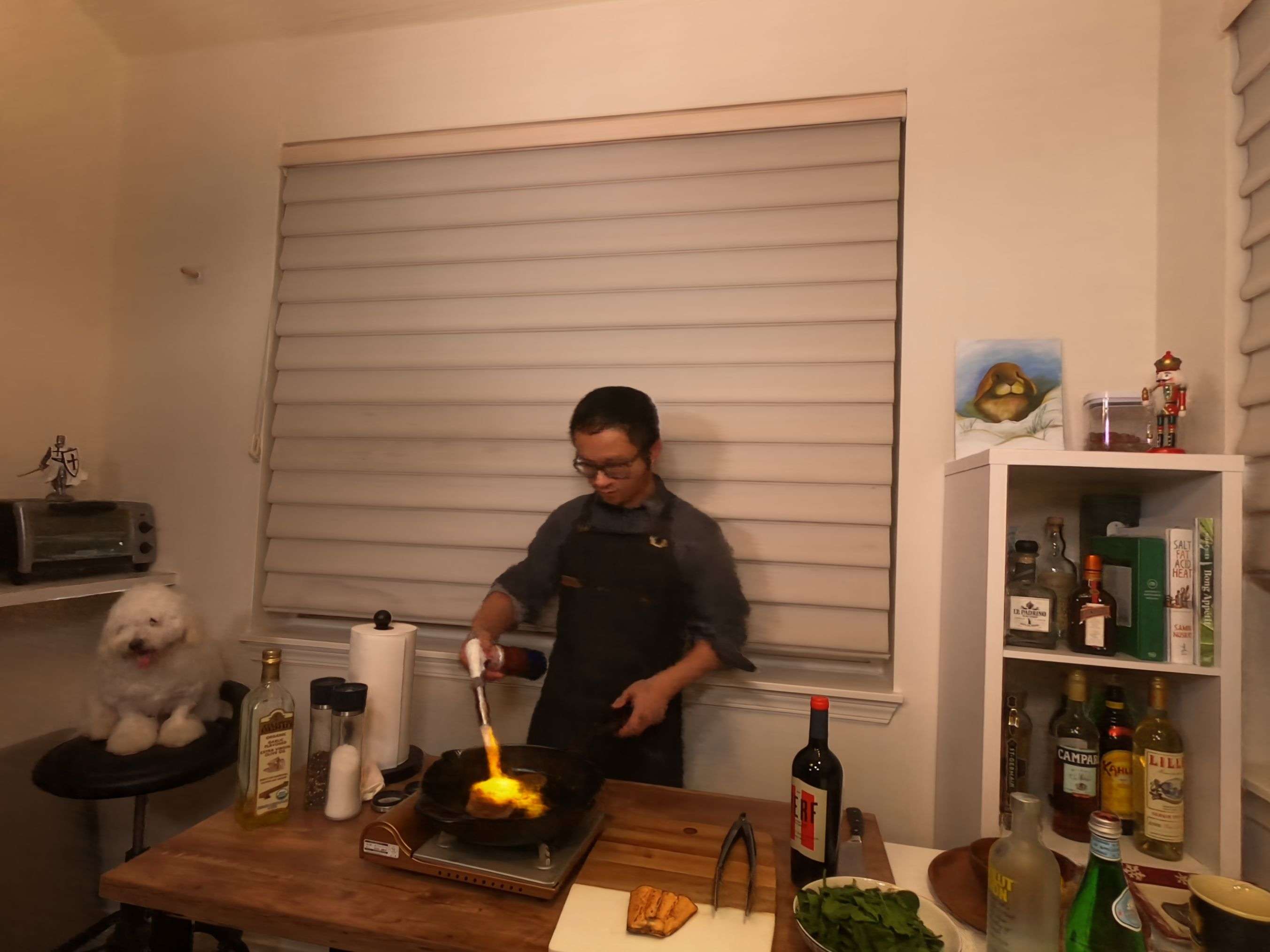}~ 
    \includegraphics[width=0.25\linewidth]{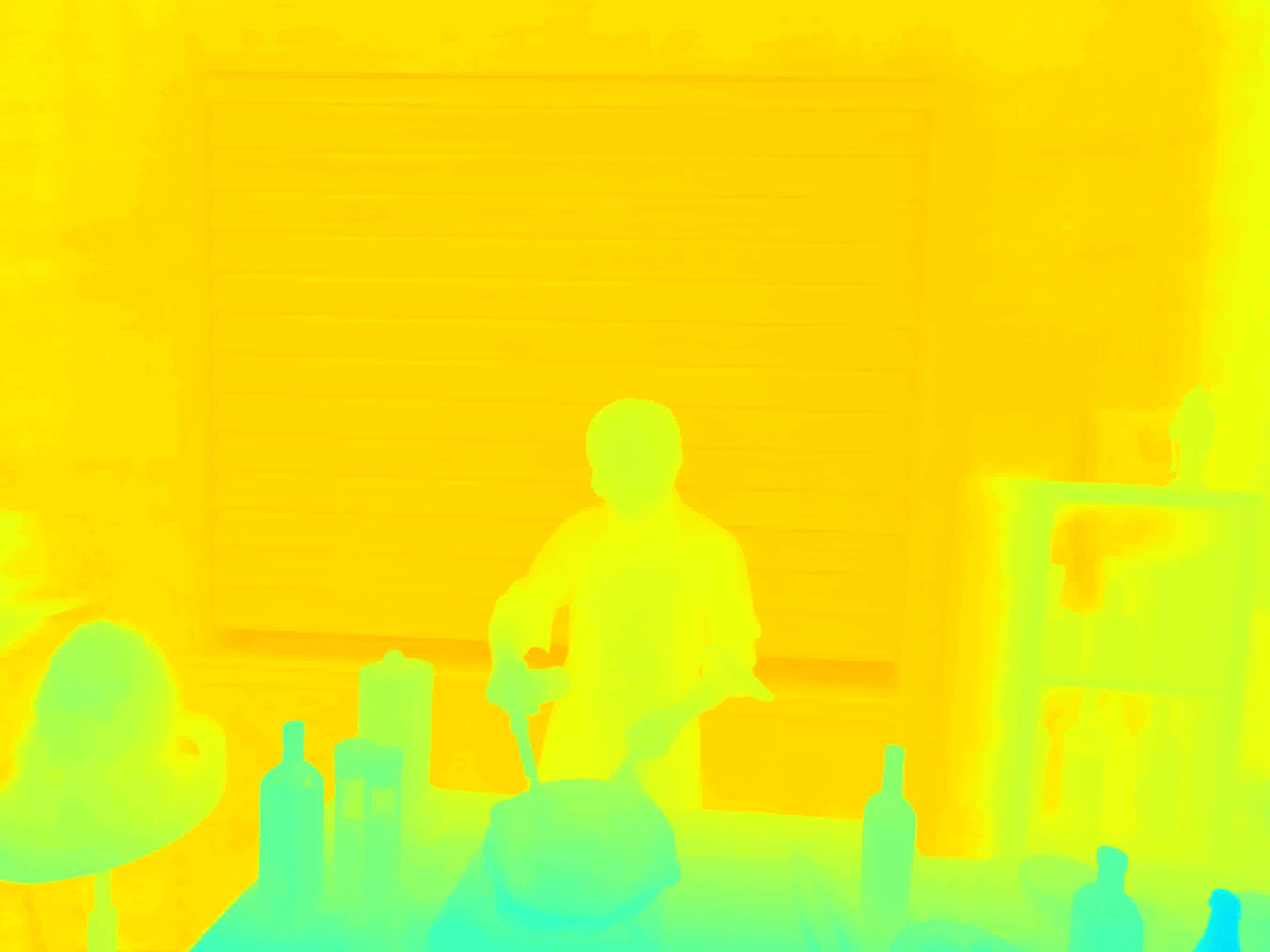}~ \\
    \includegraphics[width=0.25\linewidth]{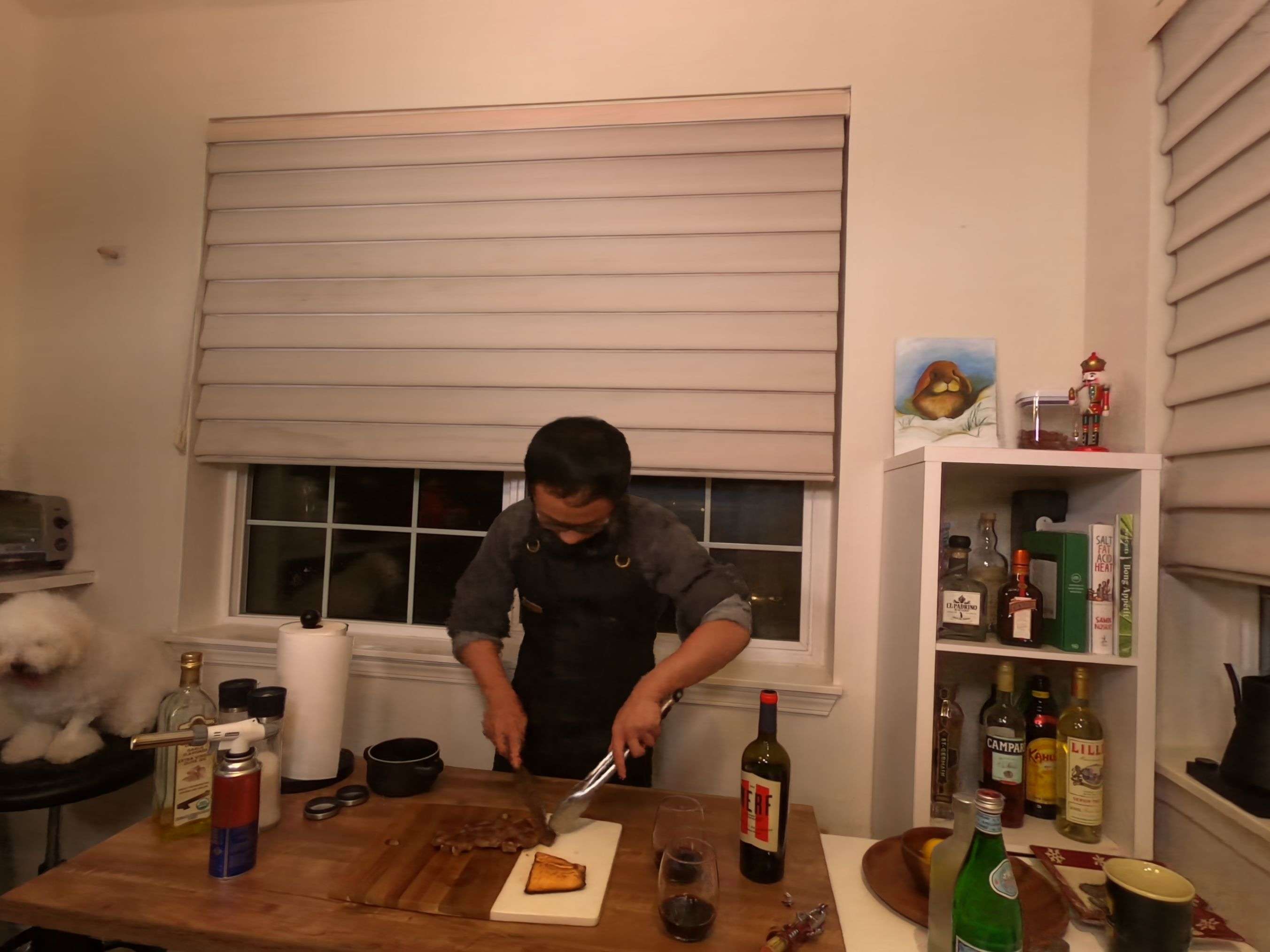}~
    \includegraphics[width=0.25\linewidth]{images/results_reduced/cut_steak/000003.jpg}~
    \includegraphics[width=0.25\linewidth]{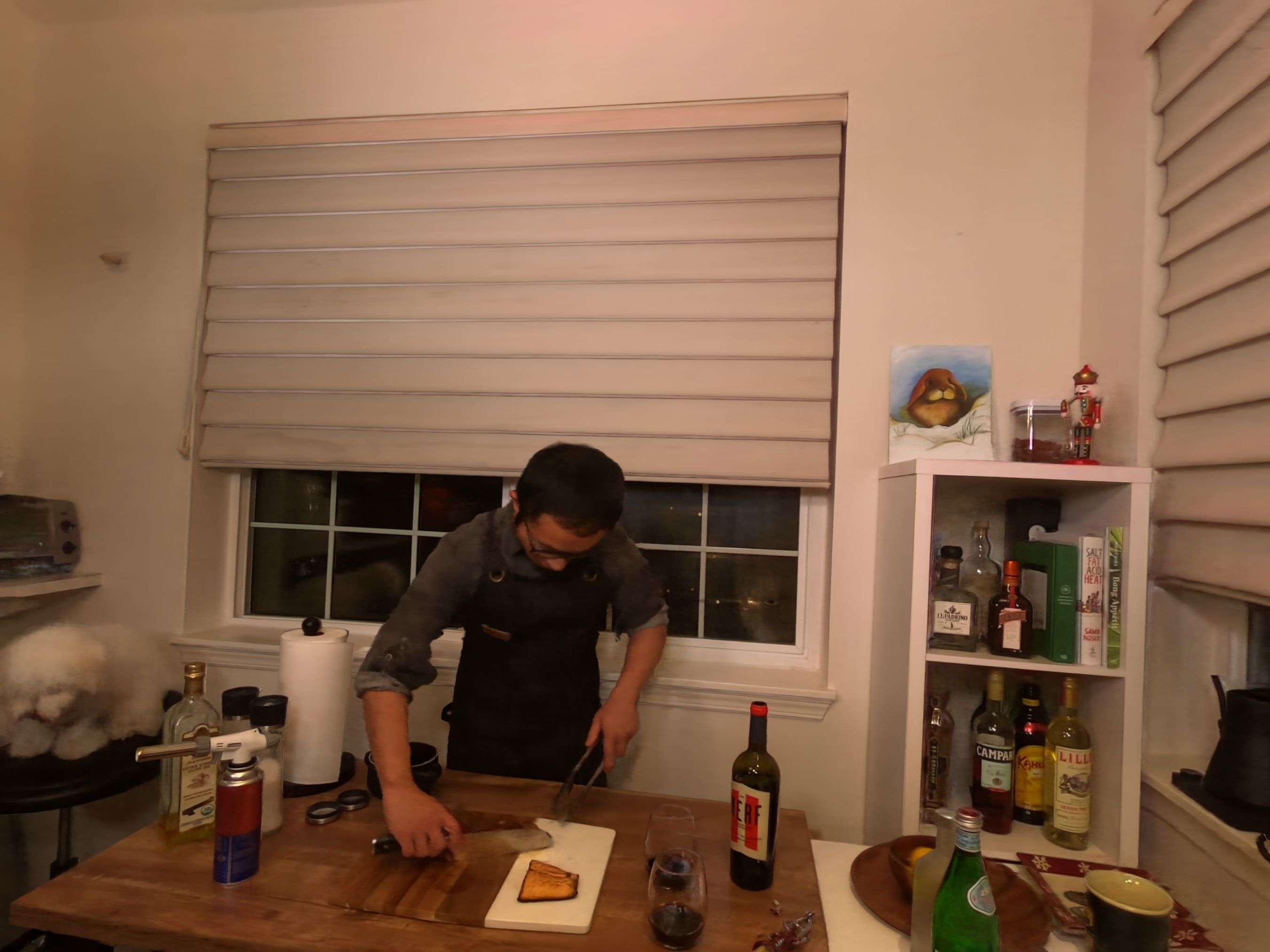}~
    \includegraphics[width=0.25\linewidth]{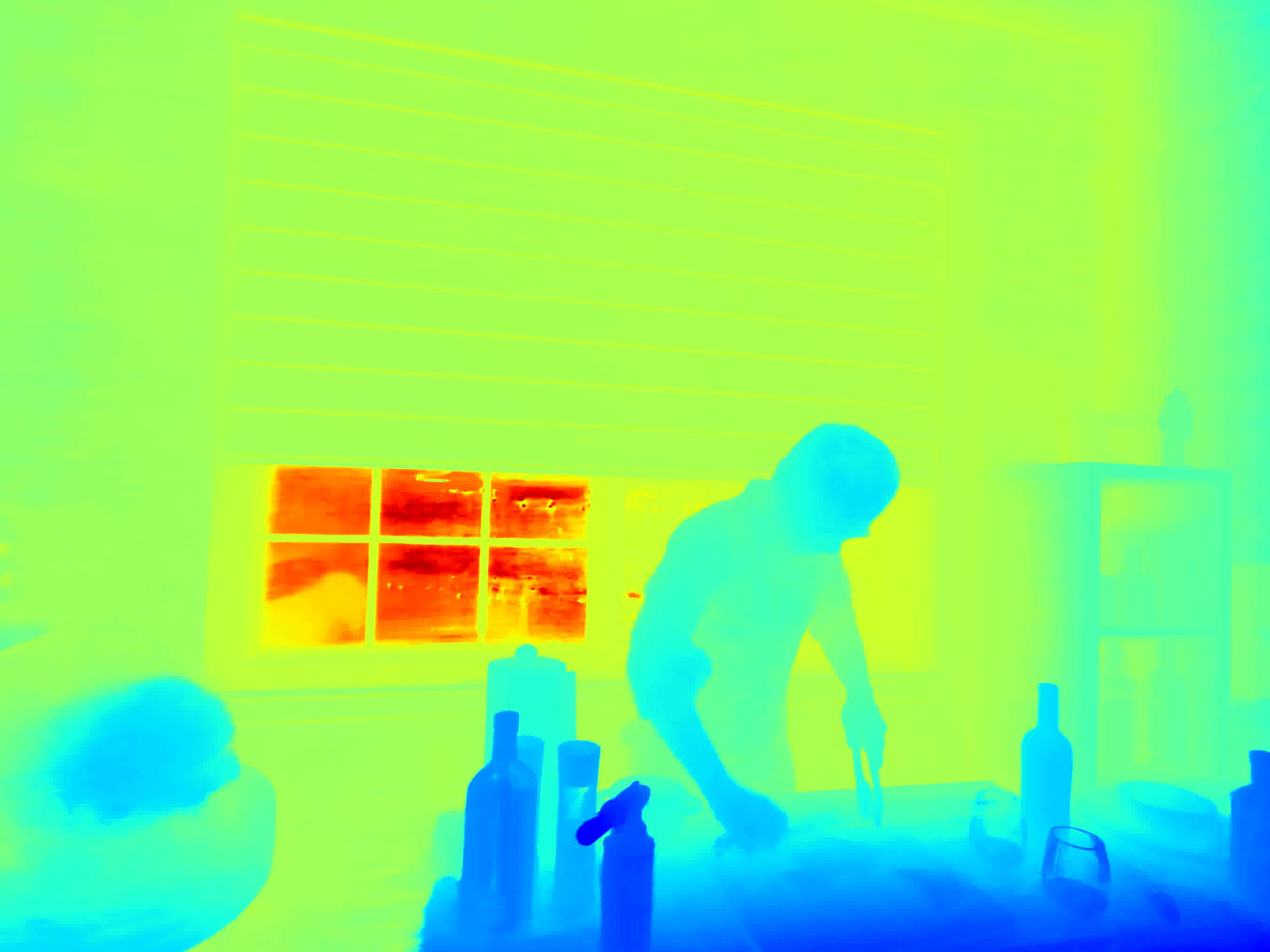}~\\
    \includegraphics[width=0.25\linewidth]{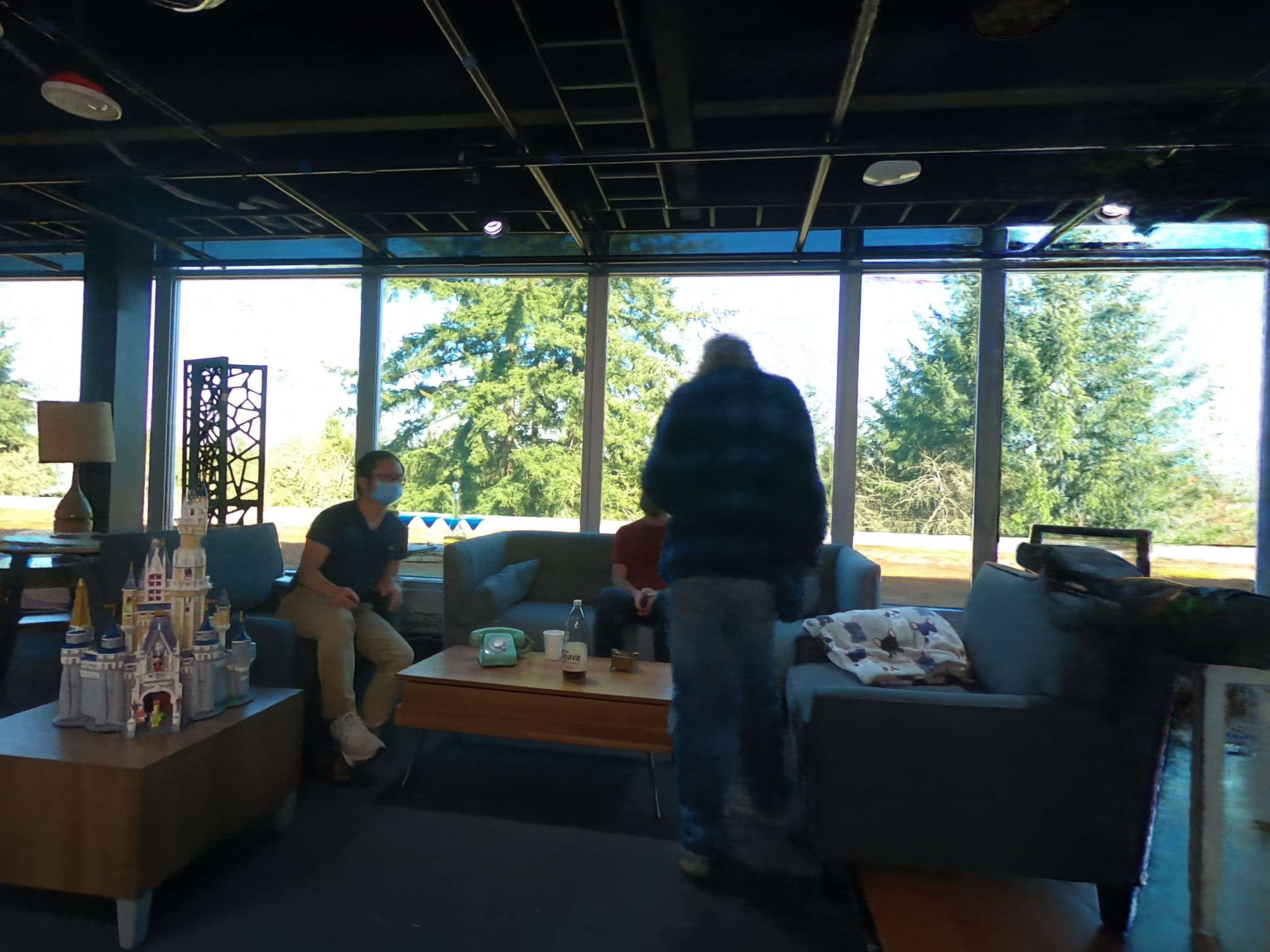}~
    \includegraphics[width=0.25\linewidth]{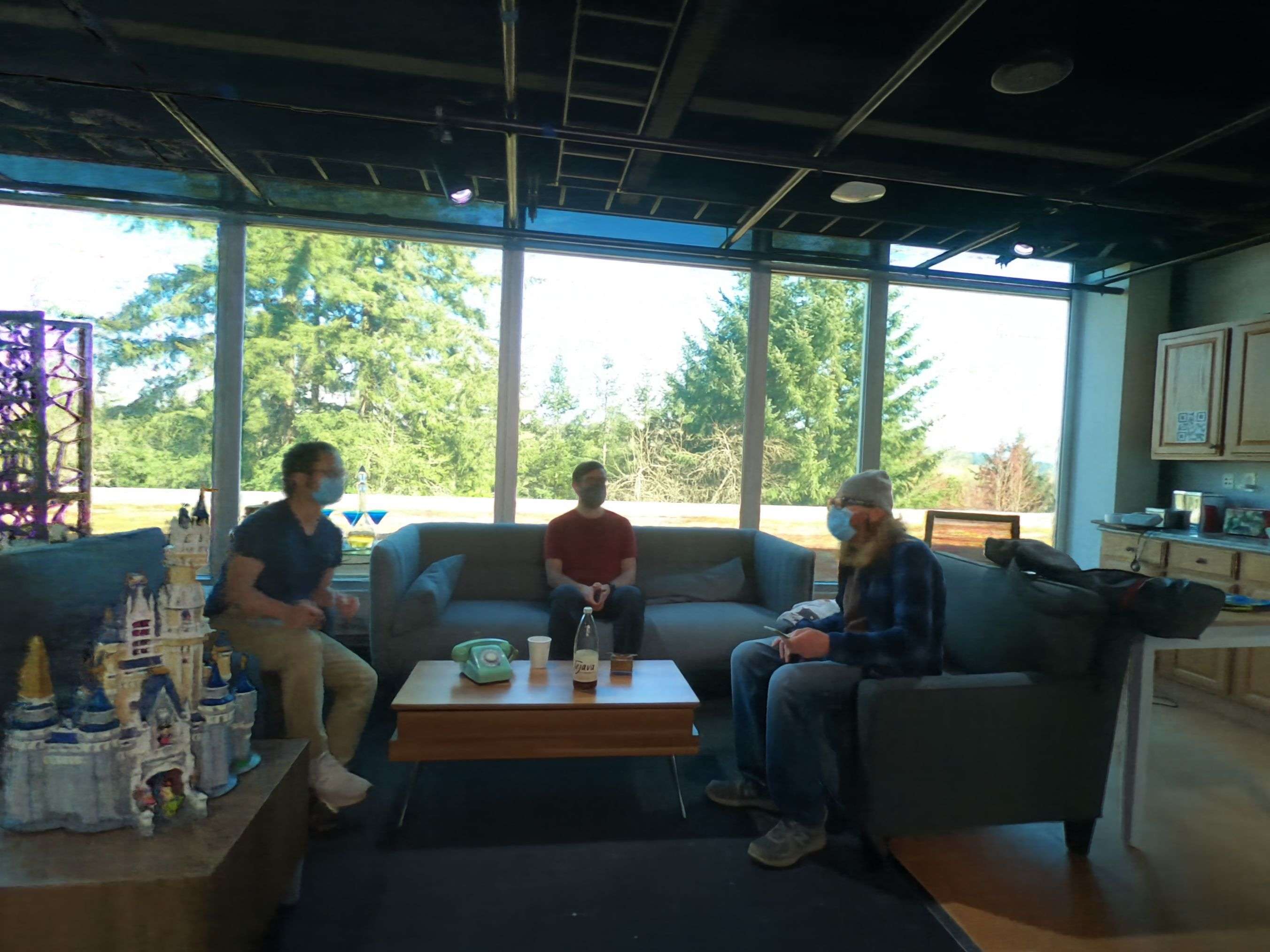}~
    \includegraphics[width=0.25\linewidth]{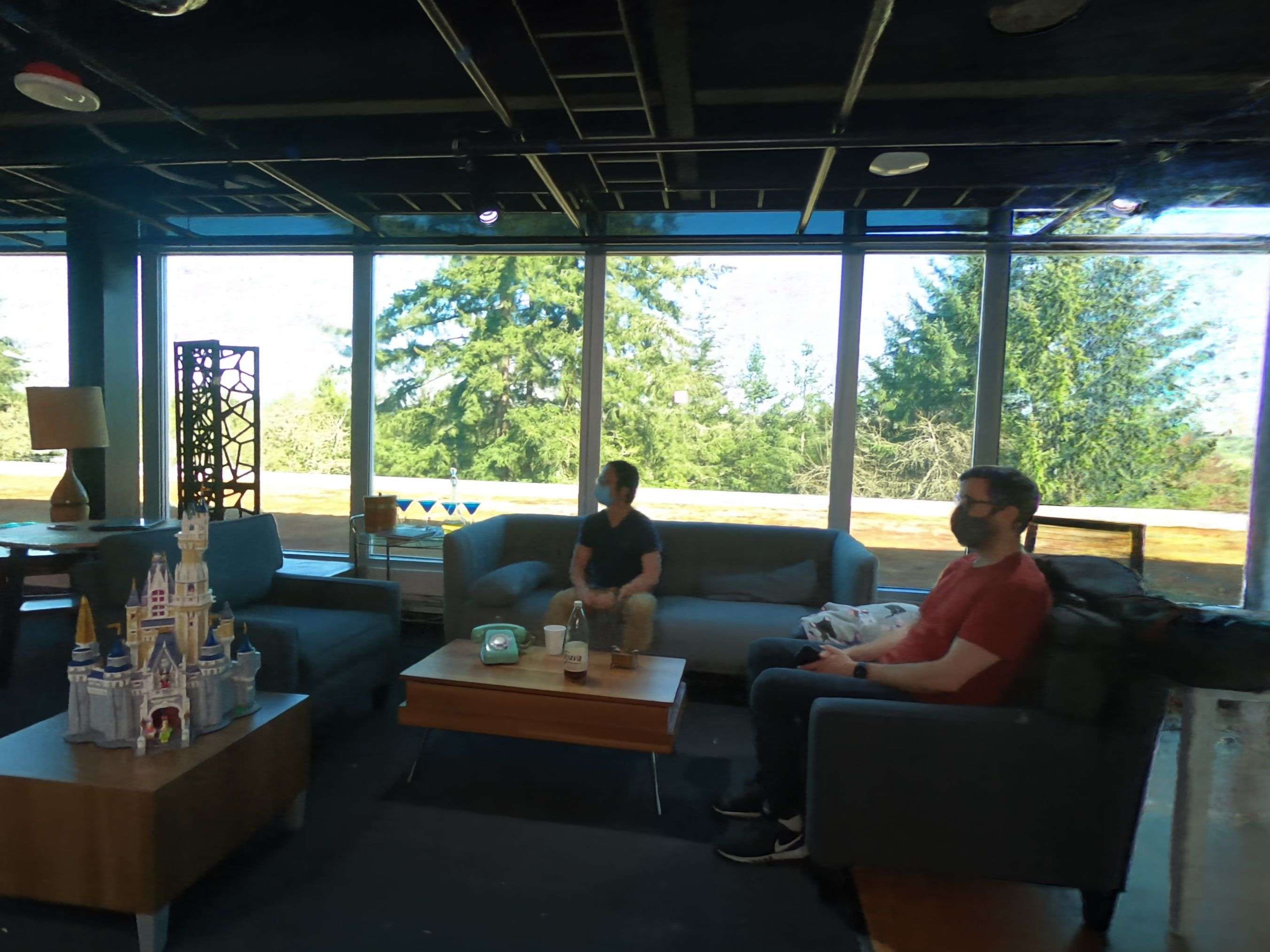}~
    \includegraphics[width=0.25\linewidth]{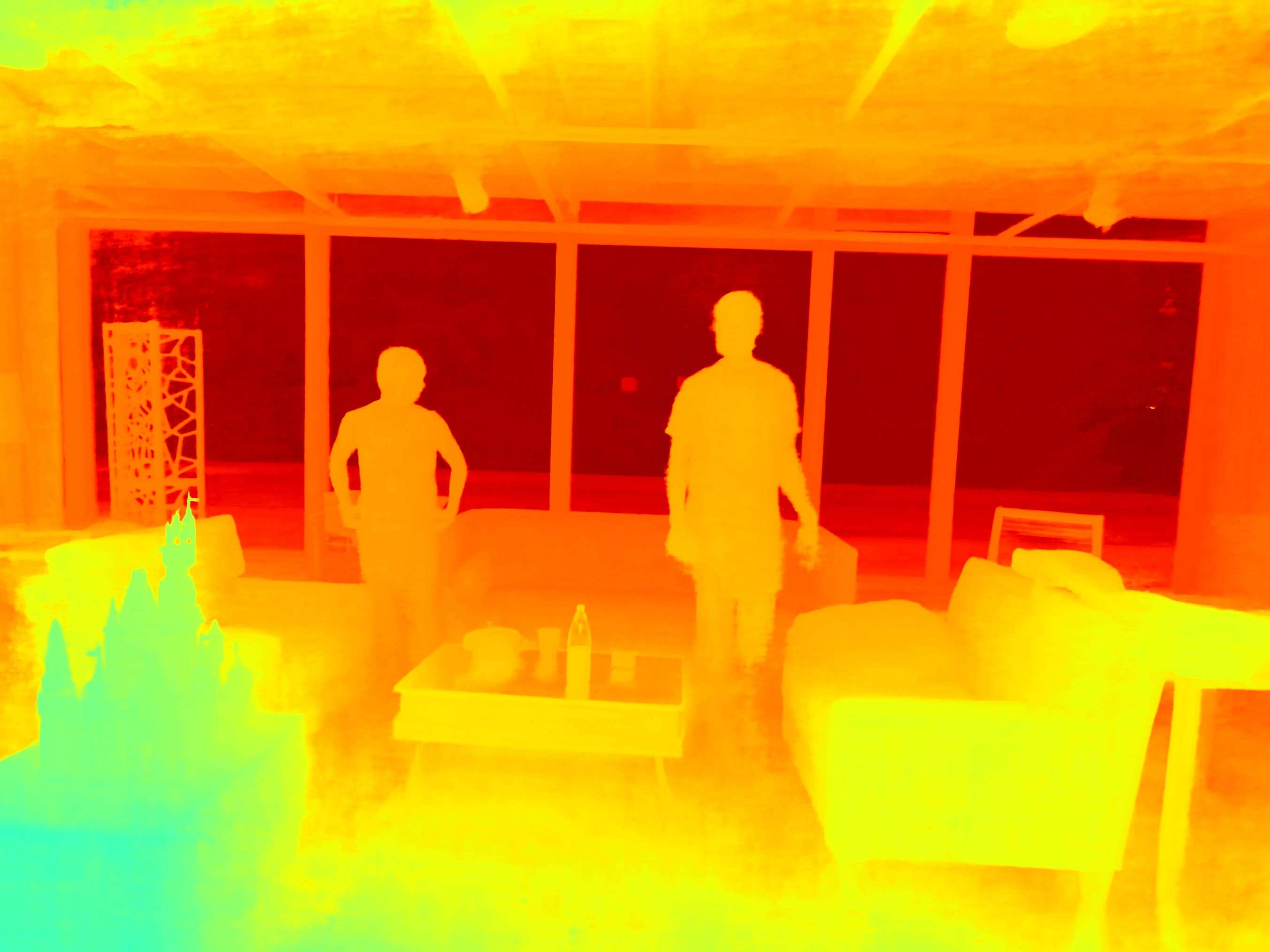}~\\
    \includegraphics[width=0.25\linewidth]{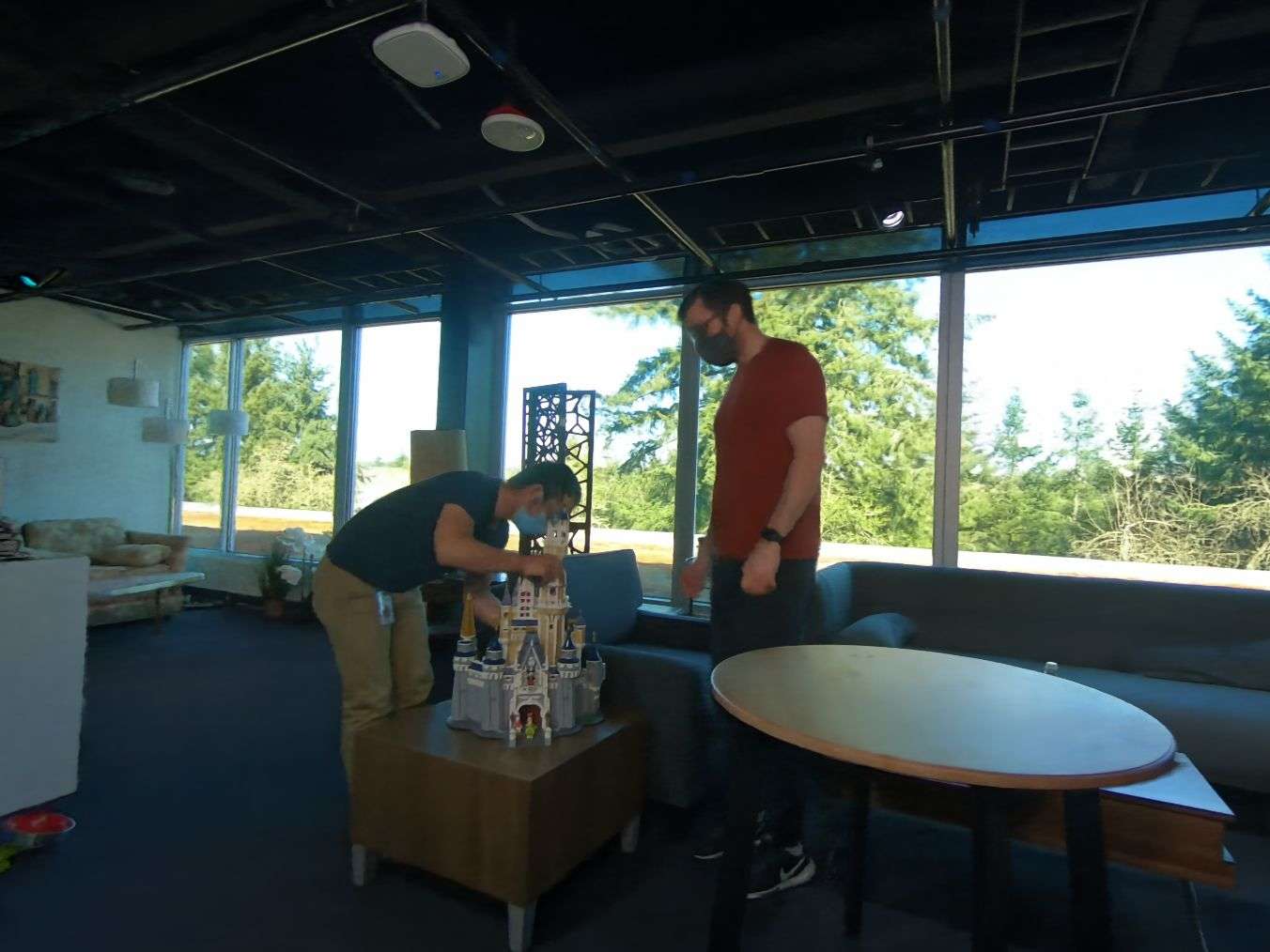}~
    \includegraphics[width=0.25\linewidth]{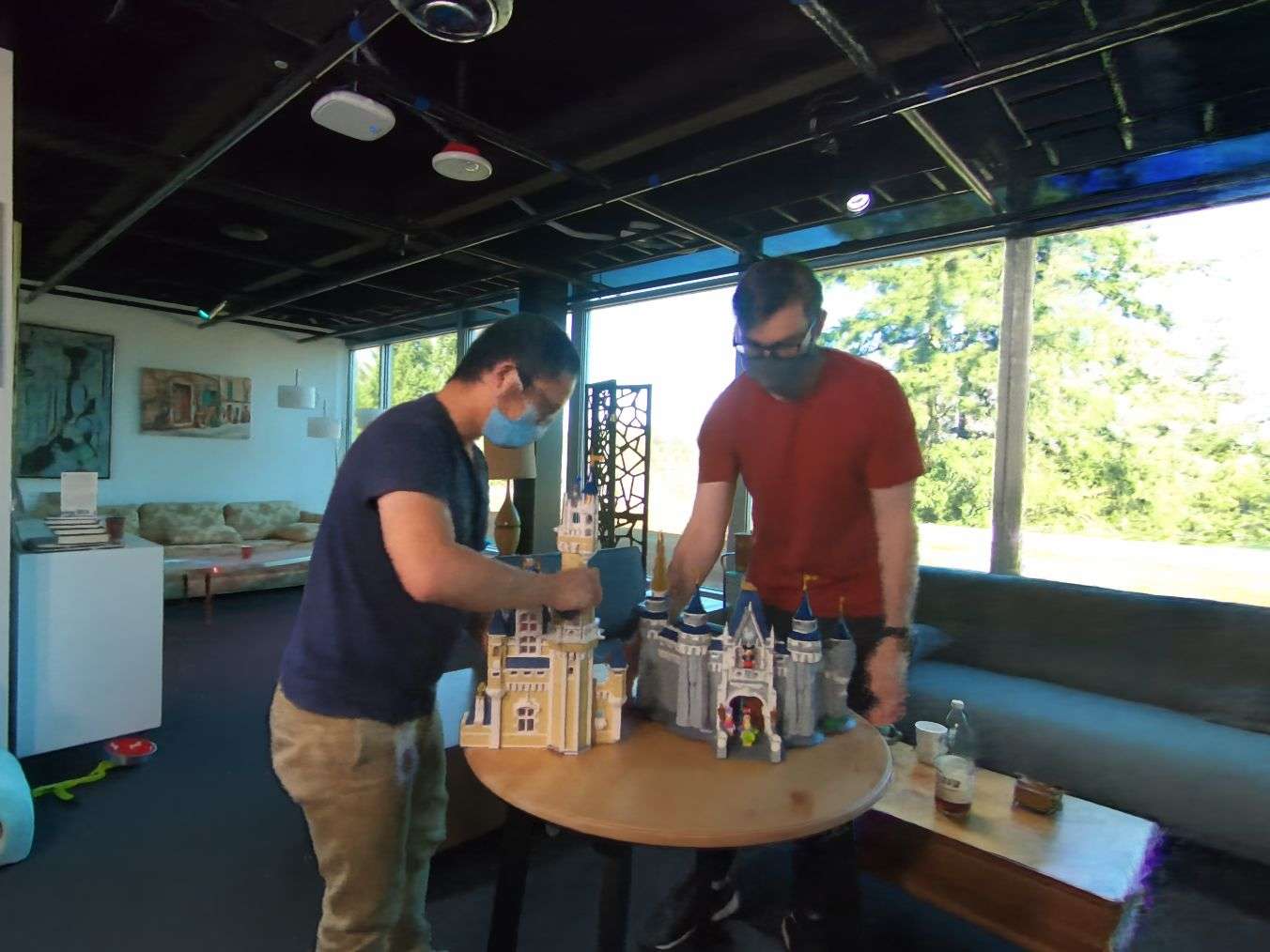}~
    \includegraphics[width=0.25\linewidth]{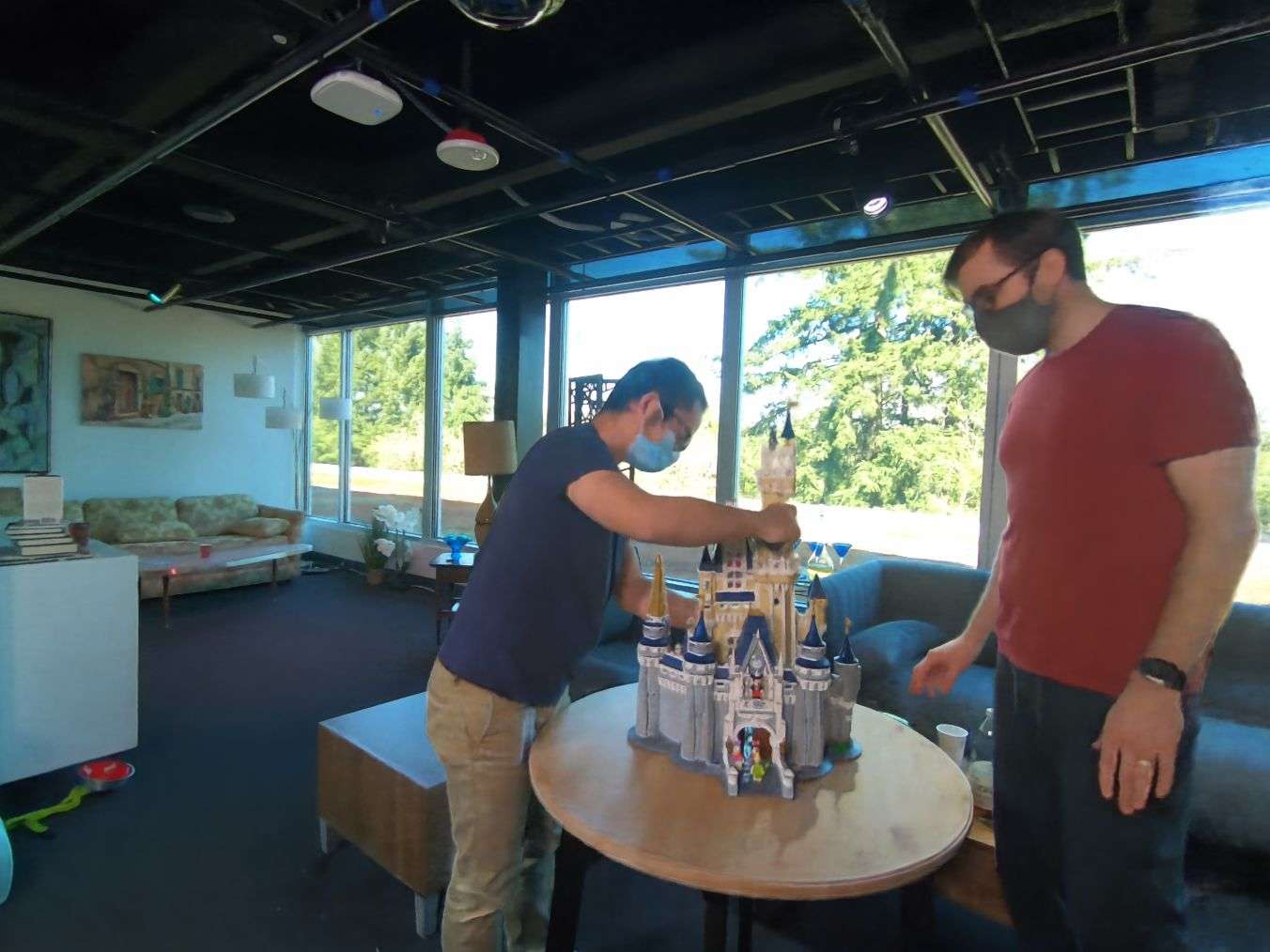}~
    \includegraphics[width=0.25\linewidth]{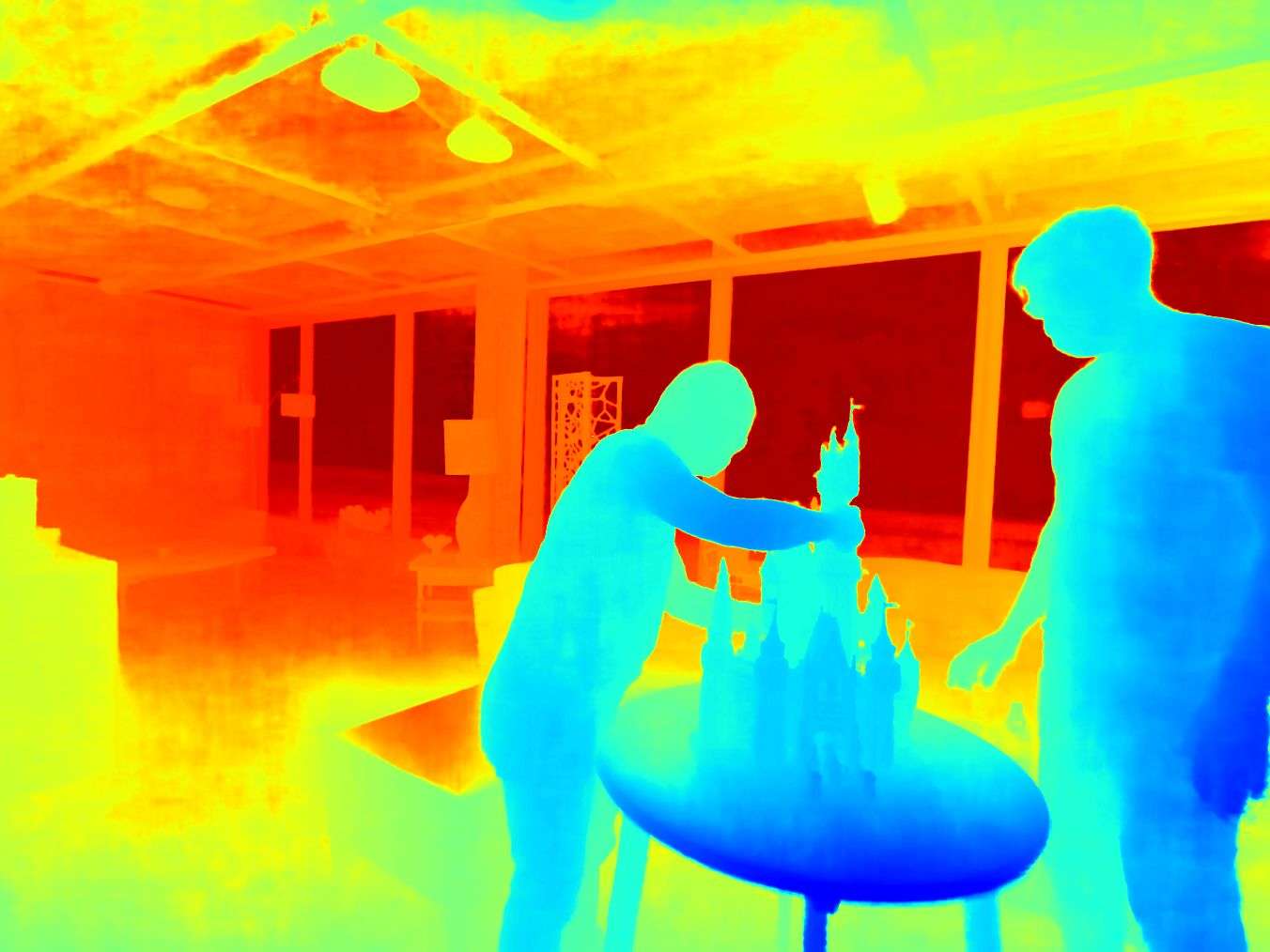}~\\

\end{subfigure}\\
\vspace{-0.5cm}
\caption{
\textbf{High-quality novel view videos} synthesized by our approach for dynamic real-world scenes. We visualize normalized depth in color space on the last column in the each row.
Our representation is compact, yet expressive and even handles complex specular reflections and translucency.
}
\label{fig:results}
\end{figure}

%% file: sections/experiments_results.tex
\input{figures/fig_comparisons}

\subsection{Results}

We demonstrate our novel view rendering results on different sequences in Fig.~\ref{fig:teaser} and Fig.~\ref{fig:results}.
Our method can represent a \SI{30}{FPS} multi-view video of up to $10$ seconds in length with at high quality. %
Our reconstructed model can enable near photorealistic continuous novel-view rendering at 1K resolution.
In the \textit{Supp. Video}, we render special visual effects such as slow motion by interpolating sub-frame latent codes between two discrete time-dependent latent codes and the ``bullet time'' effect with view-dependent effect by querying any latent code at any continuous time within the video.
Rendering with interpolated latent codes resulted in a smooth and plausible representation of dynamics between the two neighboring input frames.
Please refer to our supplementary video for the 3D video visualizations.

\input{figures/tab_ablations_all_wo_IS}  %

\input{figures/tab_video_eval}

\input{figures/fig_ablation_method_variants_wo_IS}   %

\qheading{Quantitative Comparison to the Baselines.}
Tab.~\ref{tab:ablations_methods_wo_is} shows the quantitative comparison of our methods to the baselines using an average of single frame metrics and Tab.~\ref{tab:video_eval_methods} shows the comparison to baselines using a perceptual video metric.
We train all the neural radiance field based baselines and our method the same number of iterations for fair comparison.
Compared to the existing methods, MVS, NeuralVolumes and LLFF, our method is able capture and render significant more photo-realistic images, in all the quantitative measures. 
Compared to the time-variant NeRF baseline NeRF-T and our basic DyNeRF model without our proposed training strategy (DyNeRF$^\dagger$), our DyNeRF model variants trained with our proposed training strategy perform significantly better in all metrics.

\qheading{Qualitative Comparison to the Baselines.}
We highlight visual comparisons of our methods to the baselines in Fig.~\ref{fig:baseline_comparisons} and Fig.~\ref{fig:ablation_method_variants}.
The visual results of the rendered images and FLIP error maps highlight the advantages of our approach in terms of photorealism that are not well quantified using the metrics.
In Fig.~\ref{fig:baseline_comparisons} we compare to the existing methods.
MVS with texturing suffers from incomplete reconstruction, especially for occlusion boundaries, such as image boundaries and the window regions.
The baked-in textures also cannot capture specular and transparent effects properly, e.g., the window glasses. 
LLFF~\cite{mildenhall2019local} produces blurred images with ghosting artifacts and less consistent novel view across time, especially for objects at occlusion boundaries and greater distances to the foreground, e.g., trees through the windows behind the actor. 
The results from Neural Volumes~\cite{Lombardi19tog} contain cloudy artifacts and suffer from inconsistent colors and brightness (which can be better observed in the supplemental video).
In contrast, our method achieves clear images, unobstructed by ``cloud artifacts'' and produces the best results compared to the existing methods.
In particular, the details of the actor (e.g., hat, hands) and important details (e.g., flame torch, which consists of a highly reflective surface as well as the volumetric flame appearance) are faithfully captured by our method.
Furthermore, MVS and LLFF and NeuralVolume cannot model scenes as compact and continuous spatio-temporal representation as our DyNeRF representation.
In Fig.~\ref{fig:ablation_method_variants}, we compare various settings of the dynamic neural radiance fields.
NeRF-T can only capture a blurry motion representation, which loses all appearance details in the moving regions and cannot capture view-dependent effects.
Though DyNeRF$^\dagger$ has a similar quantitative performance as NeRF-T, it has significantly improved visual quality in the moving regions compared to NeRF-T, but still struggles to recover the sharp appearance details.
DyNeRF with our proposed training strategy can recover sharp details in the moving regions, including the torch gun and the flames.

\qheading{Comparisons on Training Time.}
Our proposed method is computationally more efficient compared to alternative solutions.
Training a NeRF model frame-by-frame is the only baseline that can achieve the same photorealism as DyNeRF.
However, we find that training a single frame NeRF model to achieve the same photorealism requires about $50$ GPU hours, which in total requires $15$K GPU hours for a \SI{30}{FPS} video of $10$ seconds length.
Our method only requires $1.3$K GPU hours for the same video, which reduces the required compute by one order of magnitude.

\input{figures/fig_immersive_videos}

\qheading{Results on Immersive Video Datasets \cite{broxton2020siggraph}.} 
We further demonstrates our DyNeRF model can create reasonably well 3D immersive video using non-forward-facing and spherically distorted multi-view videos with the same parameter setting and same training time.
Fig.~\ref{fig:immersive_video_results} shows a few novel views rendered from our trained models. We include the video results in the supplementary video.
DyNeRF is able to generate an immersive coverage of the whole dynamic space with a compact model.
Compared to the frame-by-frame multi-spherical images (MSI) representation used in \cite{broxton2020siggraph}, DyNeRF represents the video as one spatial temporal model which is more compact in size (28MB for a 5s 30 FPS video) and can better represent the view-dependent effects in the scene. 
Given the same amount of training time, we also observe there are some challenges, particularly the blurriness in the fast moving regions given the same compute budget and as above.
We estimate one epoch of training time will take 4 weeks while we only trained all models using 1/4 of all pixels for a week.
It requires longer training time to gain sharpness, which remains as a challenge to our current method in computation.

%% file: figures/fig_comparisons.tex
\begin{figure*}[t]
\centering

\includegraphics[keepaspectratio, width=1\textwidth]{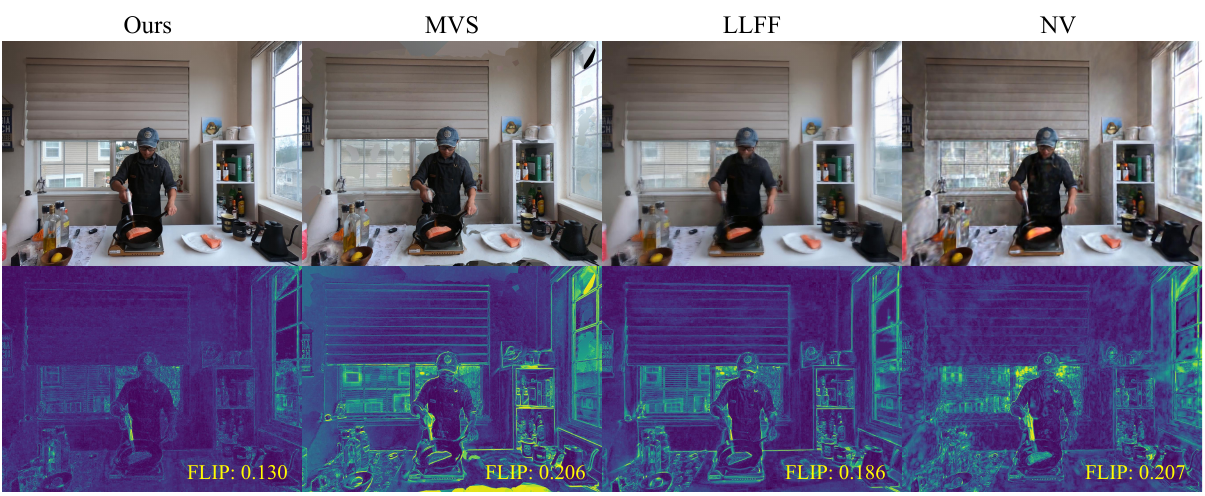}

\caption{
\textbf{Comparison of our final model to existing methods}, including Multi-view Stereo (MVS), local light field fusion (LLFF)\cite{mildenhall2019local} and NeuralVolume (NV) \cite{Lombardi19tog}. The first row shows novel view rendering on a test view. The second row visualizes the FLIP compared to the ground truth image. Compared to alternative methods, our method can achieve best visual quality. %
}
\label{fig:baseline_comparisons}
\end{figure*}

%% file: figures/tab_ablations_all_wo_IS.tex
\begin{table}[t]
\caption{
\textbf{Quantitative comparison} of our proposed method to baselines of existing methods and radiance field baselines trained at 200K iterations on a 10-second sequence. %
}
\resizebox{\linewidth}{!}{
\small
\centering
\begin{tabular}{@{}lccccc@{}}
\toprule
  Method & PSNR $\uparrow$ & MSE $\downarrow$ & DSSIM $\downarrow$ & LPIPS $\downarrow$ & FLIP $\downarrow$ \\ %
 \midrule
MVS & 19.1213 & 0.01226 & 0.1116 & 0.2599 & 0.2542 \\
NeuralVolumes & 22.7975 & 0.00525 & 0.0618 & 0.2951  & 0.2049 \\ %
LLFF  & 23.2388 & 0.00475 & 0.0762 & 0.2346 & 0.1867 \\
NeRF-T & 28.4487 & 0.00144 & 0.0228 & 0.1000 & 0.1415 \\ %
DyNeRF$^\dagger$ & 28.4994 & 0.00143 & 0.0231 & 0.0985 & 0.1455 \\ %
DyNeRF & \textbf{29.5808} & \textbf{0.00110} & \textbf{0.0197} & \textbf{0.0832} & \textbf{0.1347} \\
\bottomrule
\end{tabular}
}
\label{tab:ablations_methods_wo_is}
\end{table}

%% file: figures/tab_video_eval.tex
\begin{table}[t]
\caption{\textbf{Quantitative comparison} of our proposed method to baselines using perceptual video quality metric Just-Objectionable-Difference (JOD) \cite{Mantiuk2021tog}. Higher number (maximum 10) indicates less noticeable visual difference to the ground truth.}
\small
\centering
\begin{tabular}{@{}lcccc@{}}
\toprule
  Method & NeuralVolumes & LLFF & NeRF-T & DyNeRF \\
 \midrule
JOD $\uparrow$ & 6.50 & 6.48 & 7.73 & \textbf{8.07} \\
\bottomrule
\end{tabular}
\label{tab:video_eval_methods}
\end{table}

%% file: figures/fig_ablation_method_variants_wo_IS.tex
\begin{figure}[ht]

\centering

\includegraphics[width=1\linewidth]{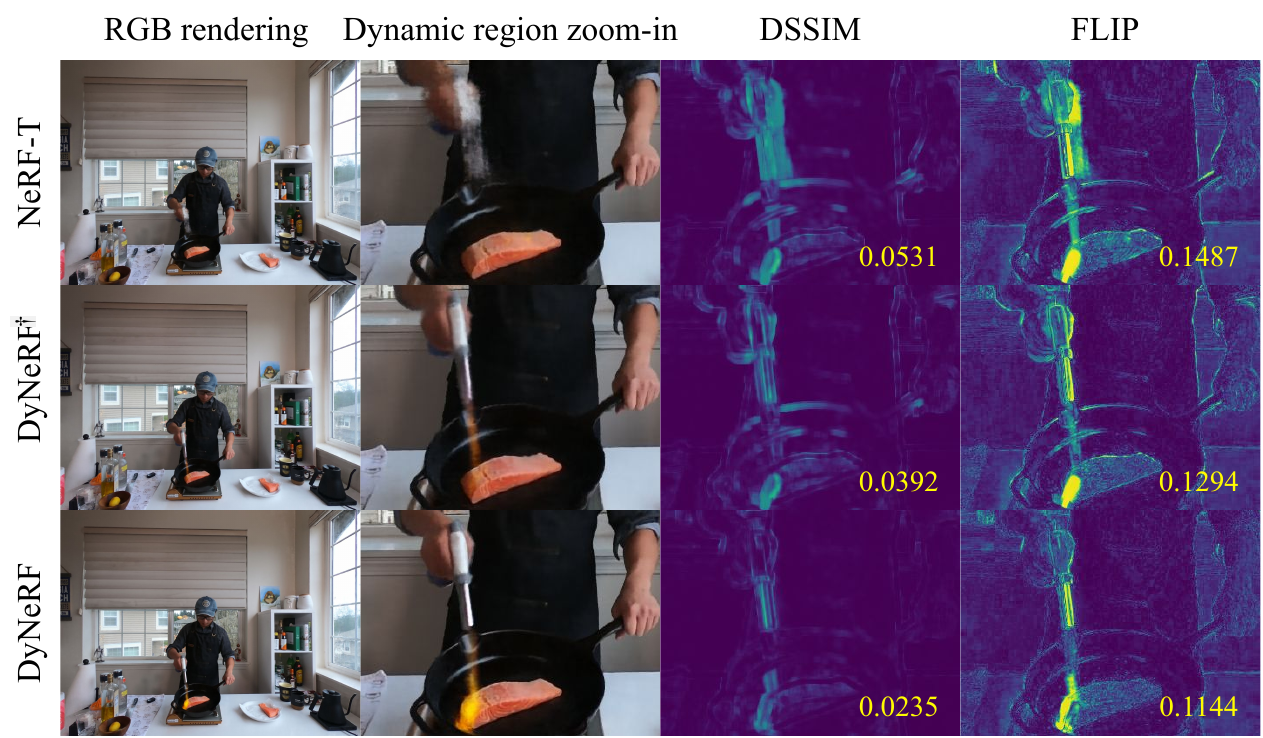}

\caption{
\textbf{Qualitative comparisons} of DyNeRF variants on one image of the sequence whose averages are reported in Tab.~\ref{tab:ablations_methods_wo_is}.
From left to right we show the rendering by each method, then zoom onto the moving flame gun, then visualize DSSIM and FLIP for this region using the \emph{viridis} colormap  (dark blue is 0, yellow is 1, lower is better).
}
\label{fig:ablation_method_variants}

\end{figure}

%% file: figures/fig_immersive_videos.tex
\begin{figure}[t]
\centering

\begin{subfigure}[t]{\linewidth}\centering
    \includegraphics[width=0.24\linewidth]{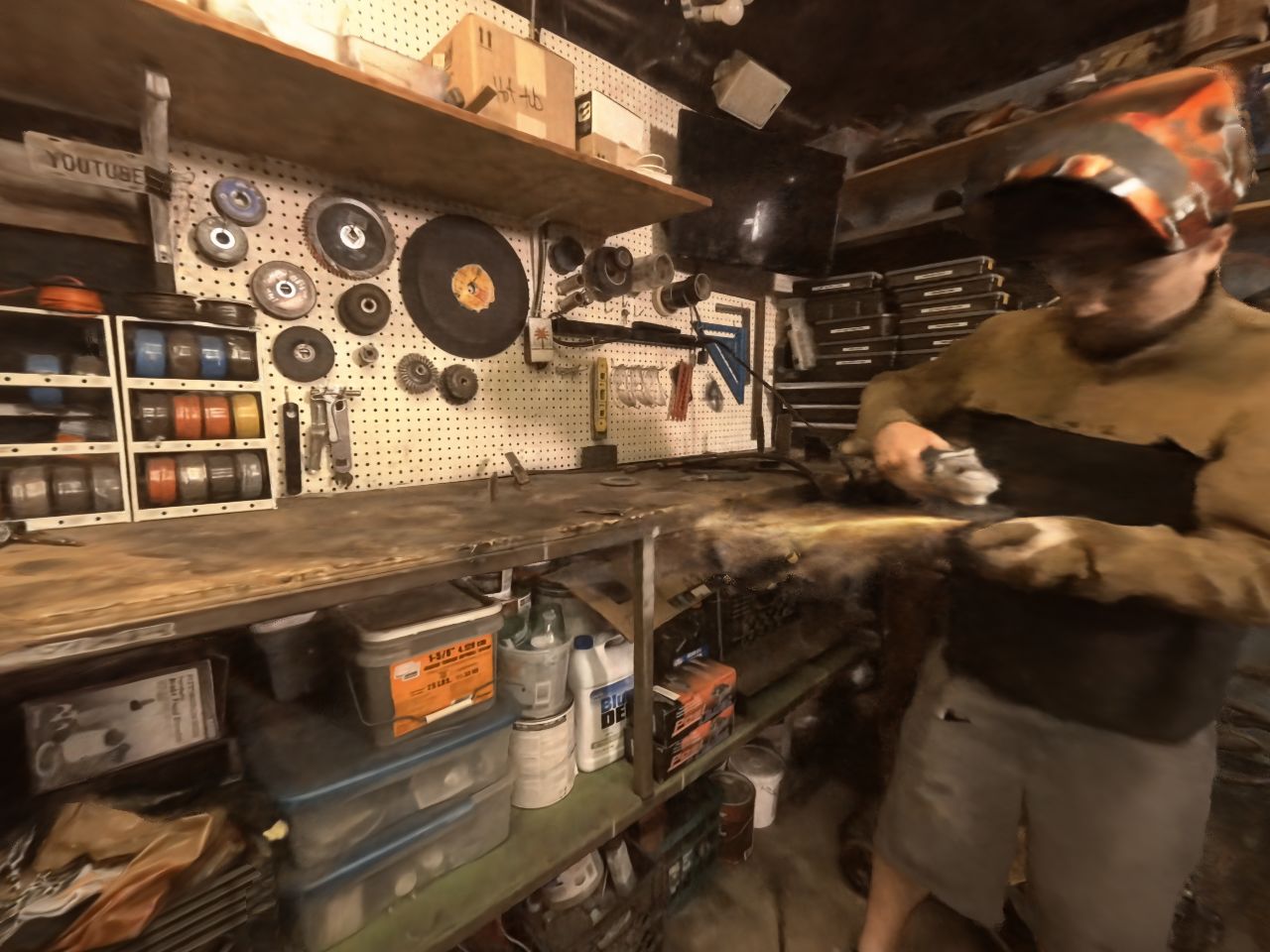}~
    \includegraphics[width=0.24\linewidth]{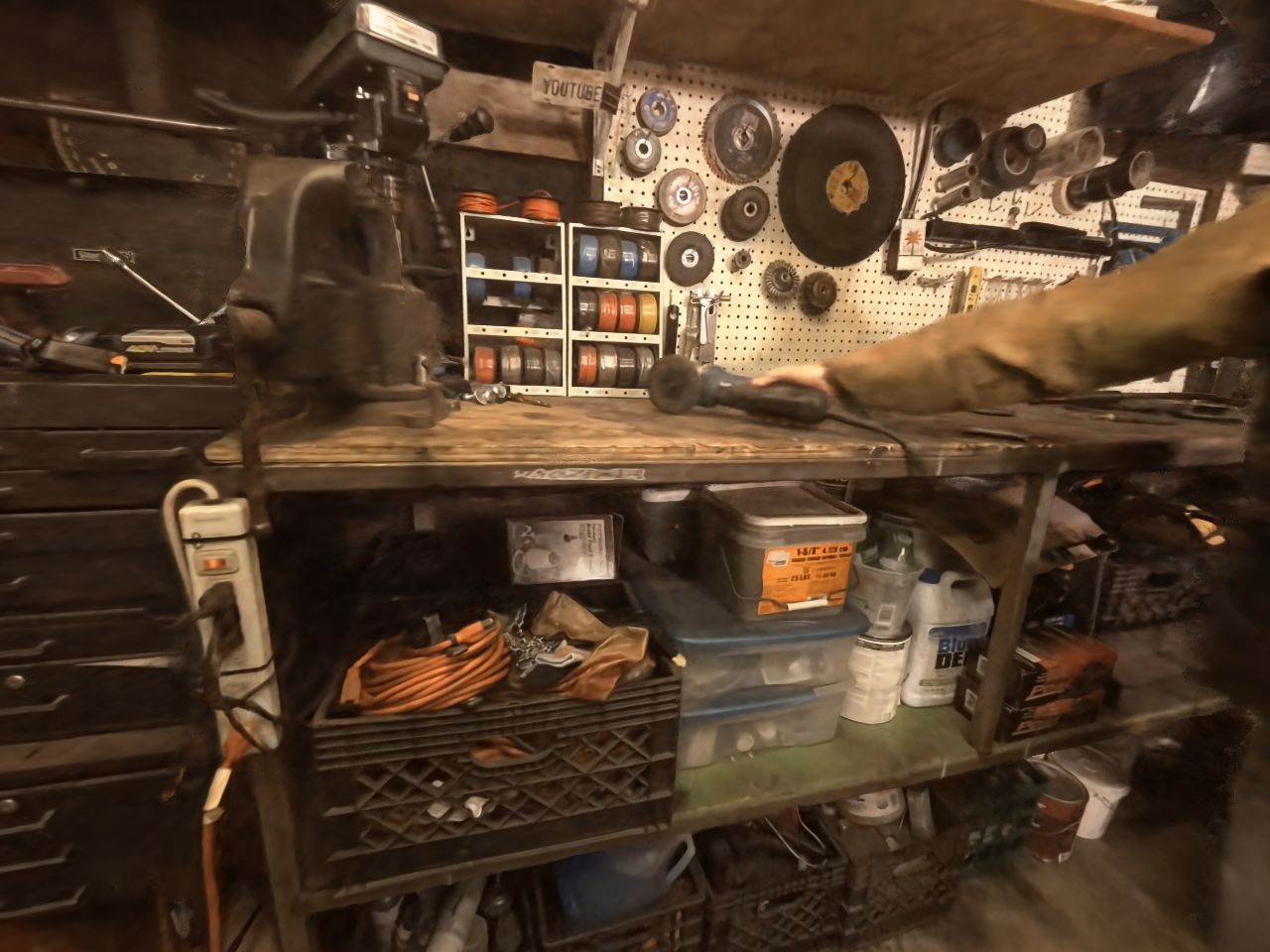}~ 
    \includegraphics[width=0.24\linewidth]{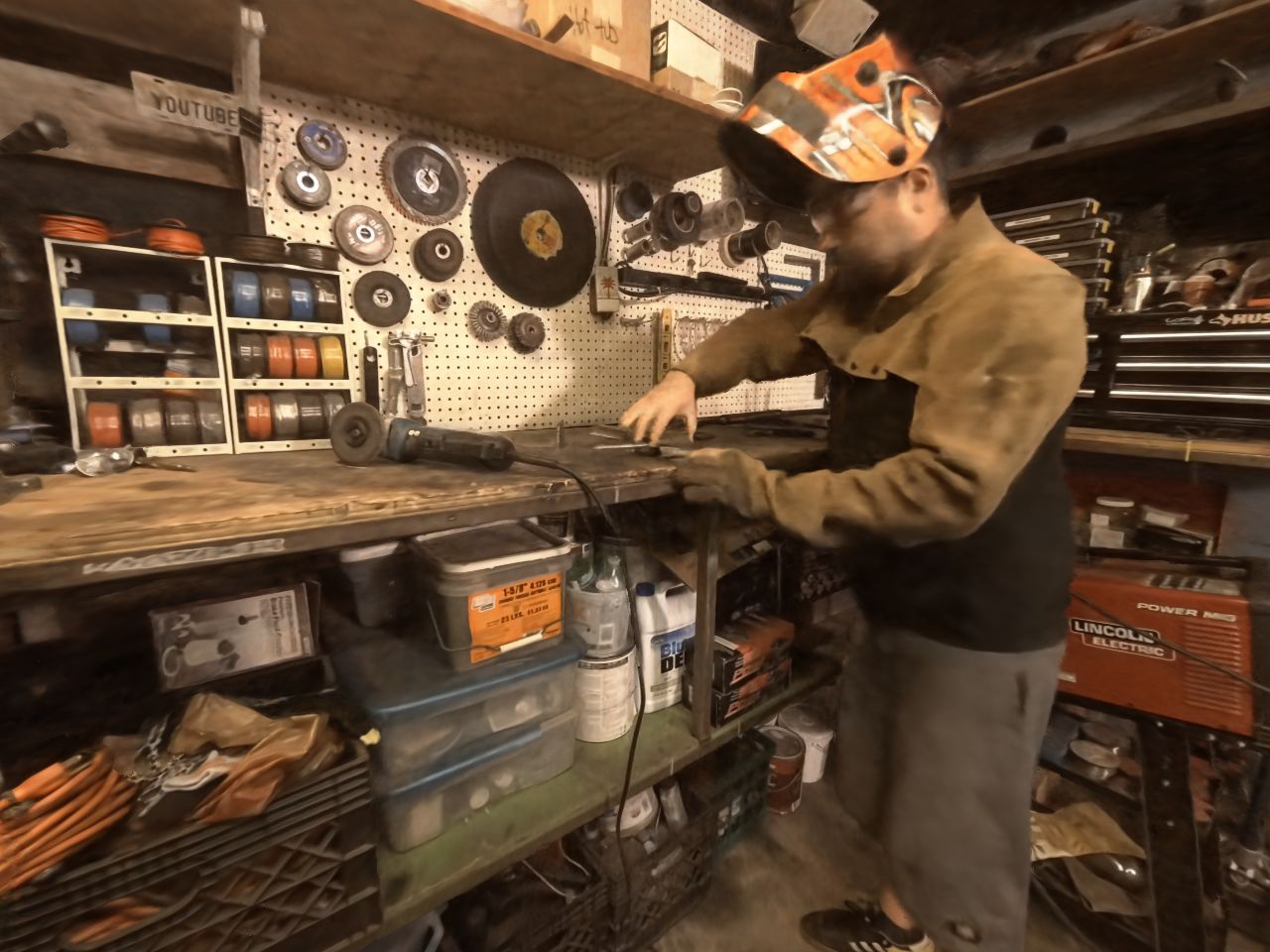}~
    \includegraphics[width=0.24\linewidth]{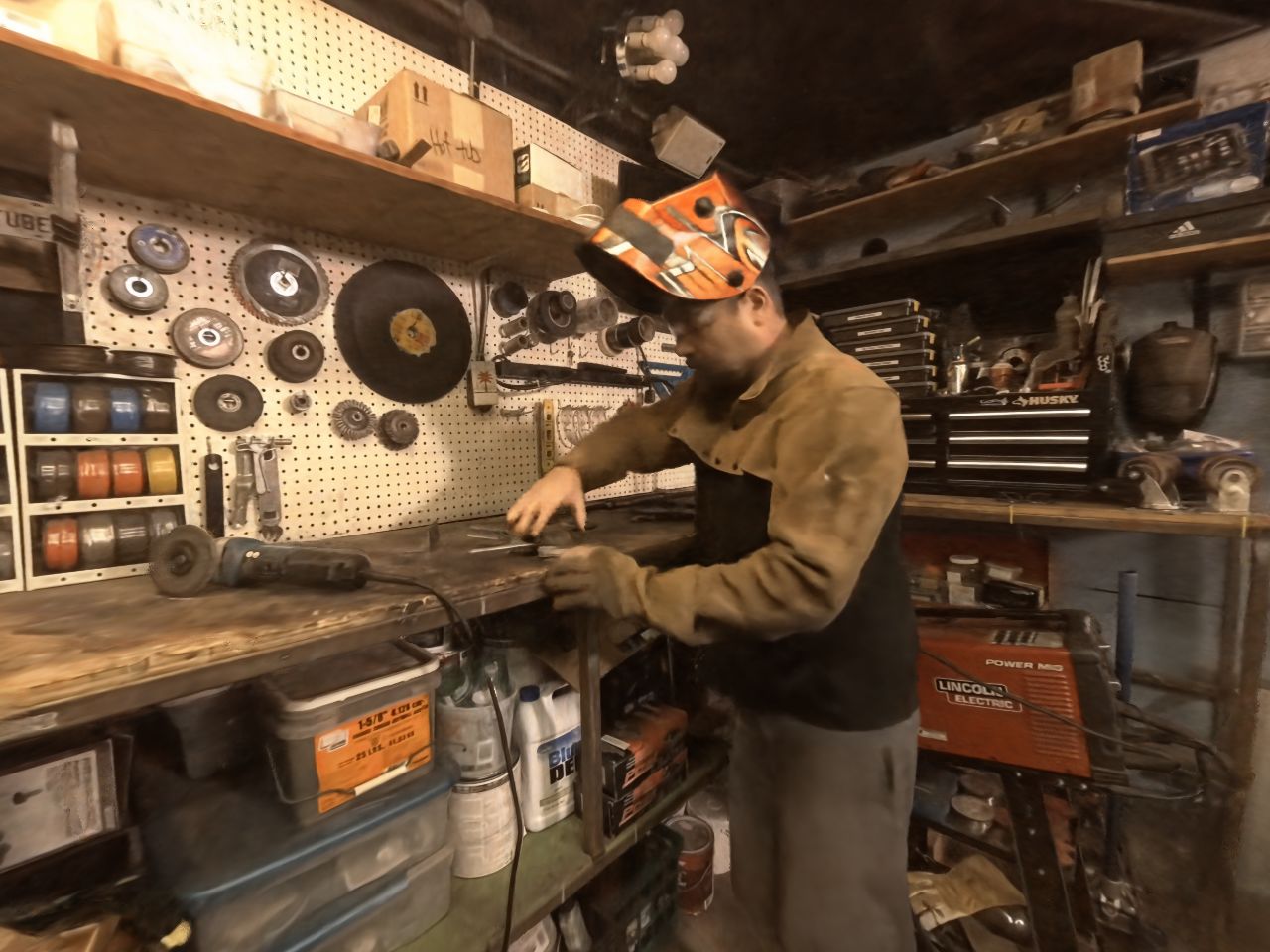}~ 
    \\
    \includegraphics[width=0.24\linewidth]{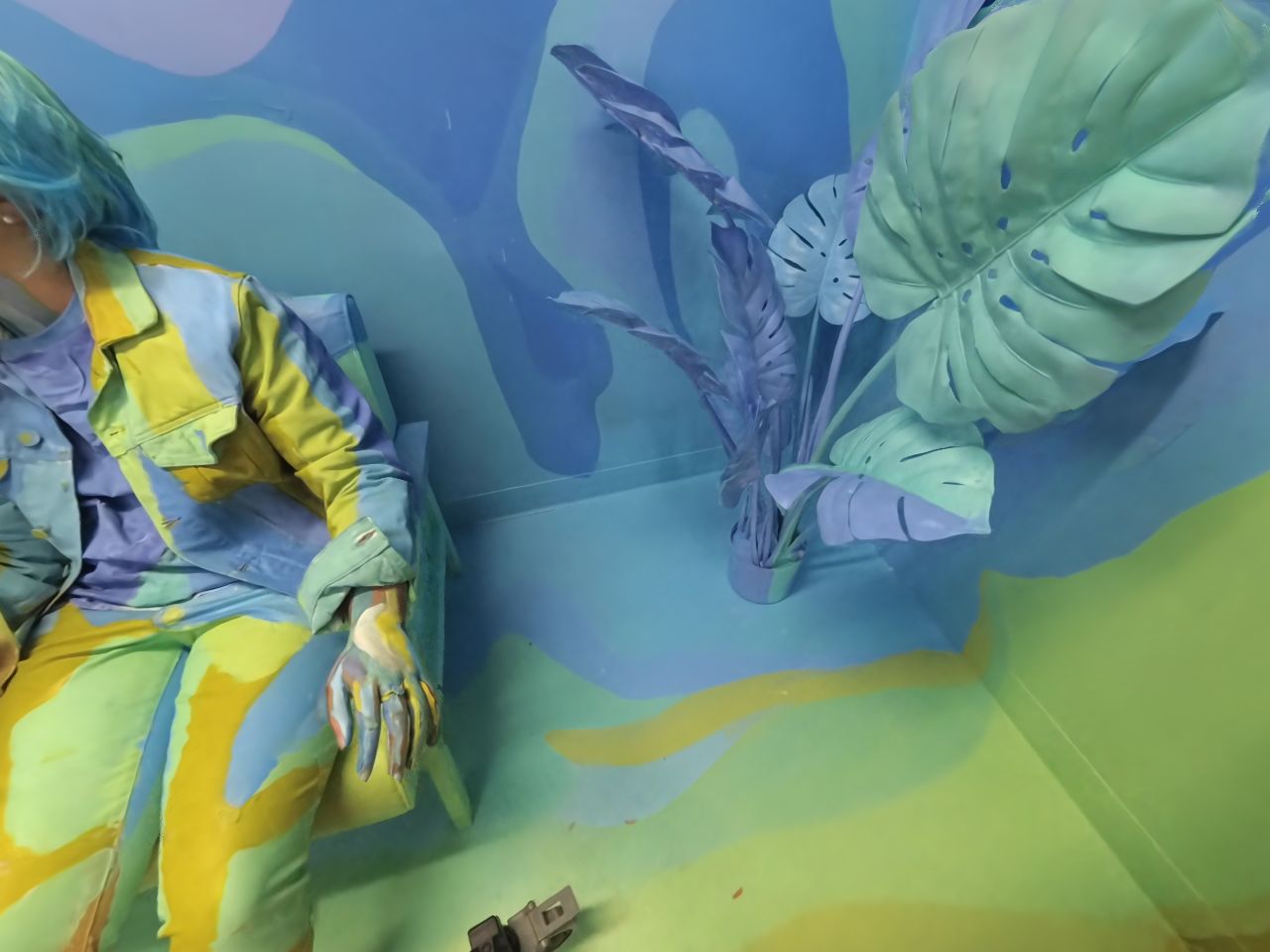}~
    \includegraphics[width=0.24\linewidth]{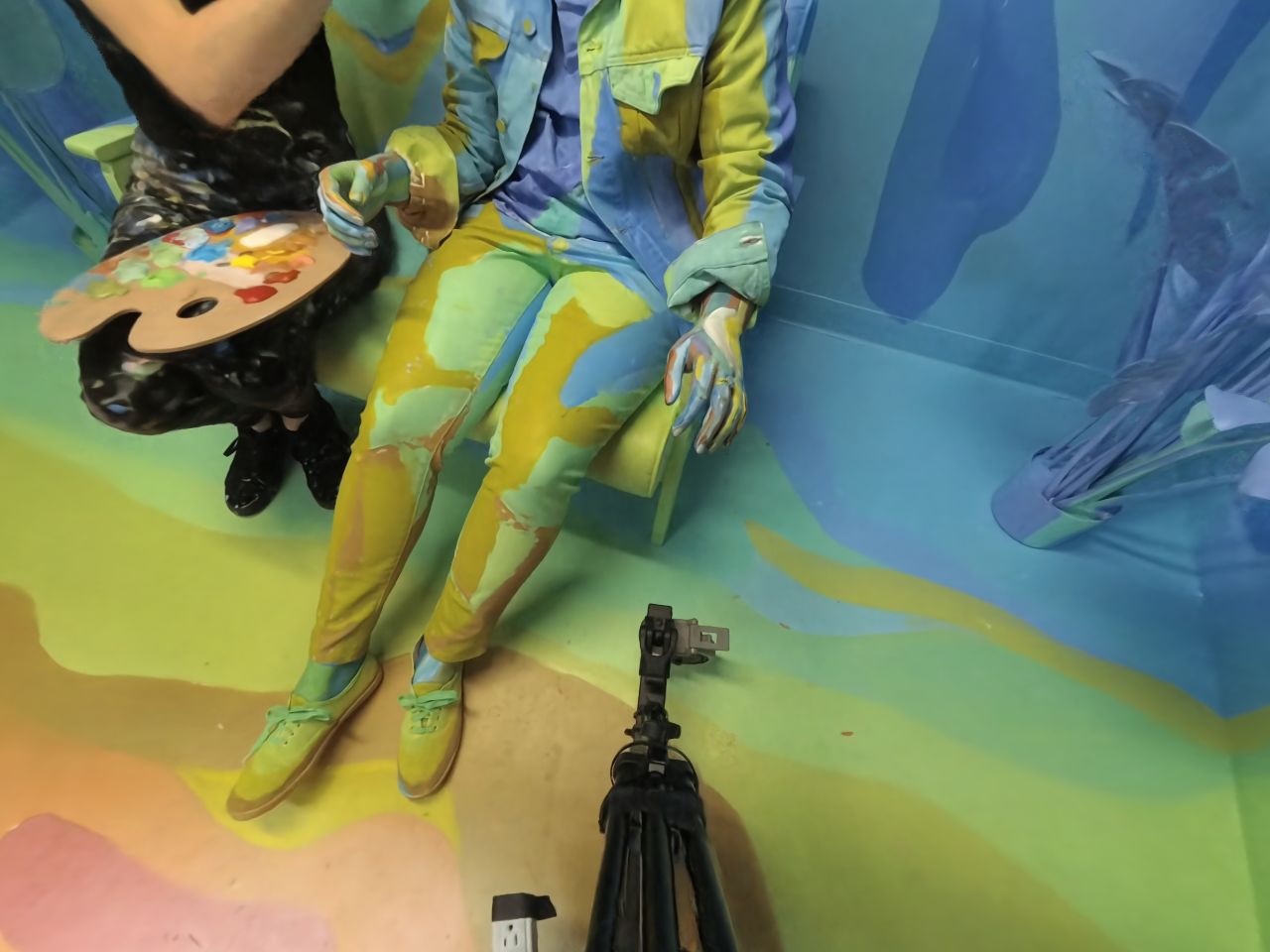}~
    \includegraphics[width=0.24\linewidth]{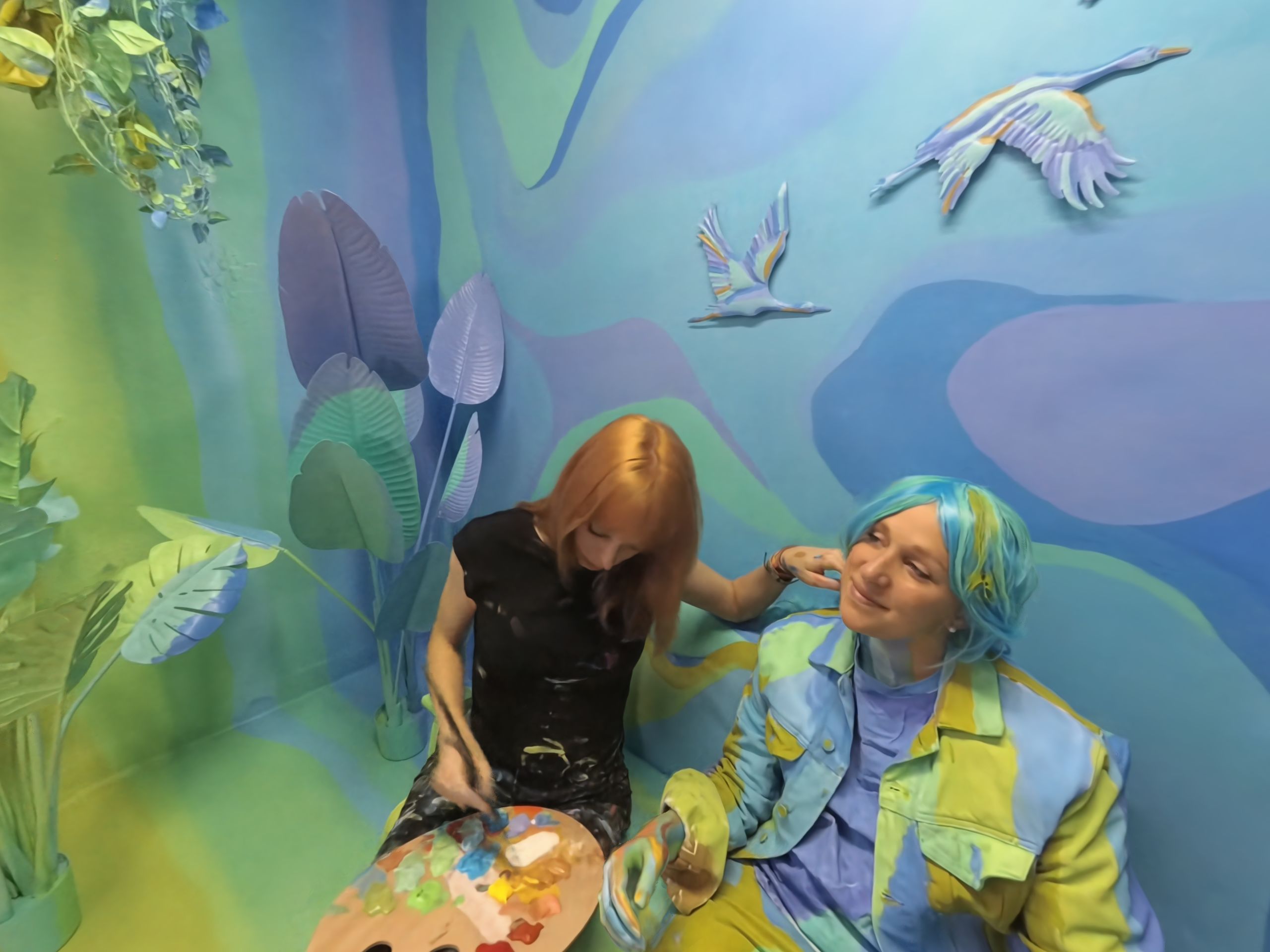}~
    \includegraphics[width=0.24\linewidth]{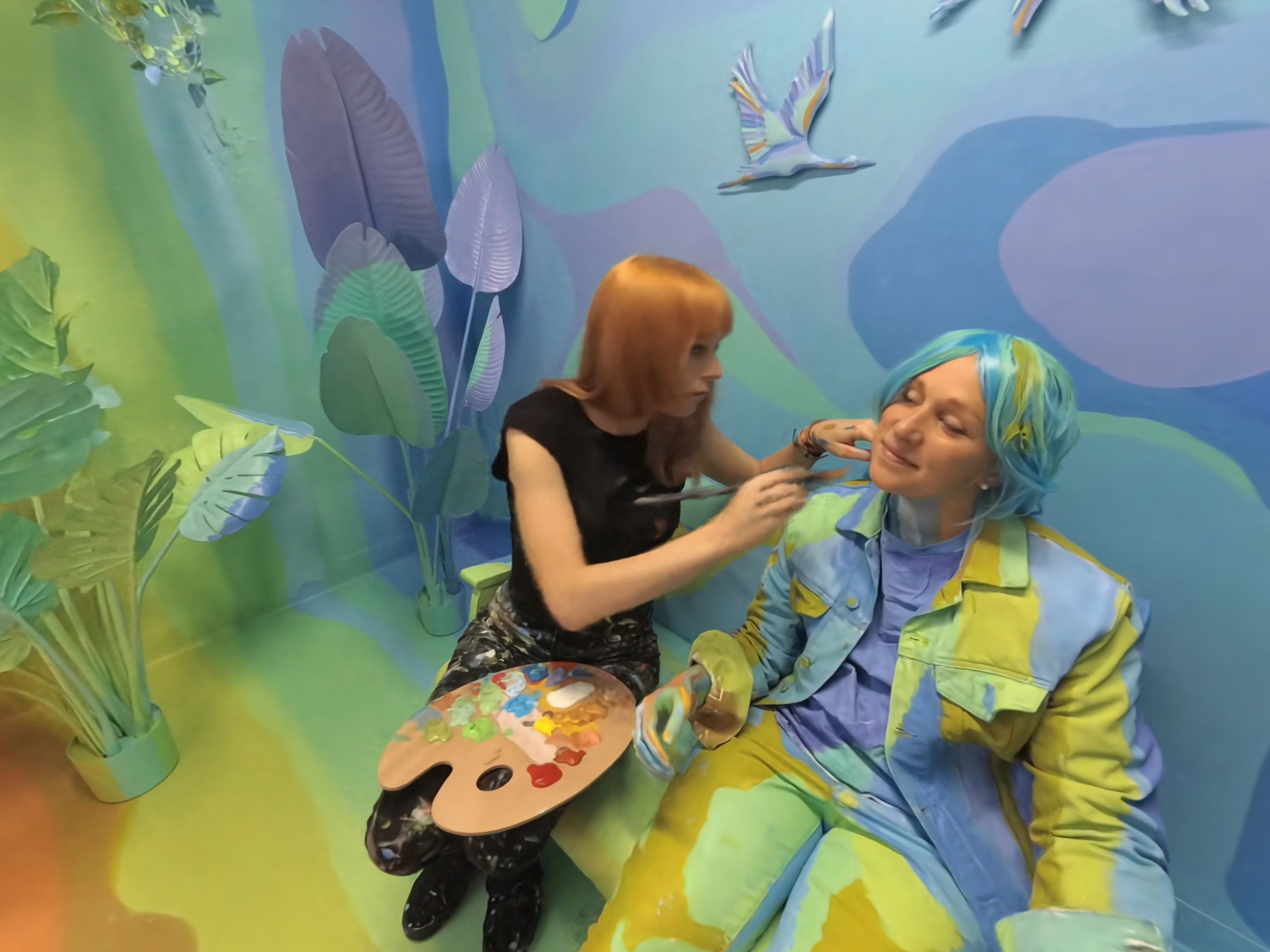}~ \\

\end{subfigure}\\
\caption{
\textbf{Snapshots of novel view rendered videos} on immersive video datasets \cite{broxton2020siggraph}. 
}
\label{fig:immersive_video_results}
\end{figure}

%% file: sections/limitations.tex
\qheading{Limitations.}
There are a few challenging scenarios that our method is currently facing.
(1)~Highly dynamic scenes with large and fast motions are challenging to model and learn, which might lead to blur in the moving regions. As shown in Fig. \ref{fig:limitation}, we observe it is particularly difficult to tackle fast motion in a complex environment, e.g. outdoors with forest structure behind. 
An adaptive sampling strategy during the hierarchical training that places more keyframes during the challenging parts of the sequence or more explicit motion modeling could help to further improve results.
(2)~While we already achieve a significant improvement in terms of training speed compared to the baseline approaches, training still takes a lot of time and compute resources.
Finding ways to further decrease training time and to speed up rendering at test time are required.
(3)~Viewpoint extrapolation beyond the bounds of the training views is challenging and might lead to artifacts in the rendered imagery.
We hope that, in the future, we can learn strong scene priors that will be able to fill in the missing information.
(4)~We discussed the importance sampling strategy and its effectiveness based on the assumption of videos observed from static cameras. 
We leave the study of this strategy on videos from moving cameras as future work.
We believe these current limitations are good directions to explore in follow-up work and that our approach is a stepping stone in this direction.

\input{figures/fig_limitation}

%% file: figures/fig_limitation.tex
\begin{figure}[t]

\centering

\includegraphics[width=0.24\linewidth]{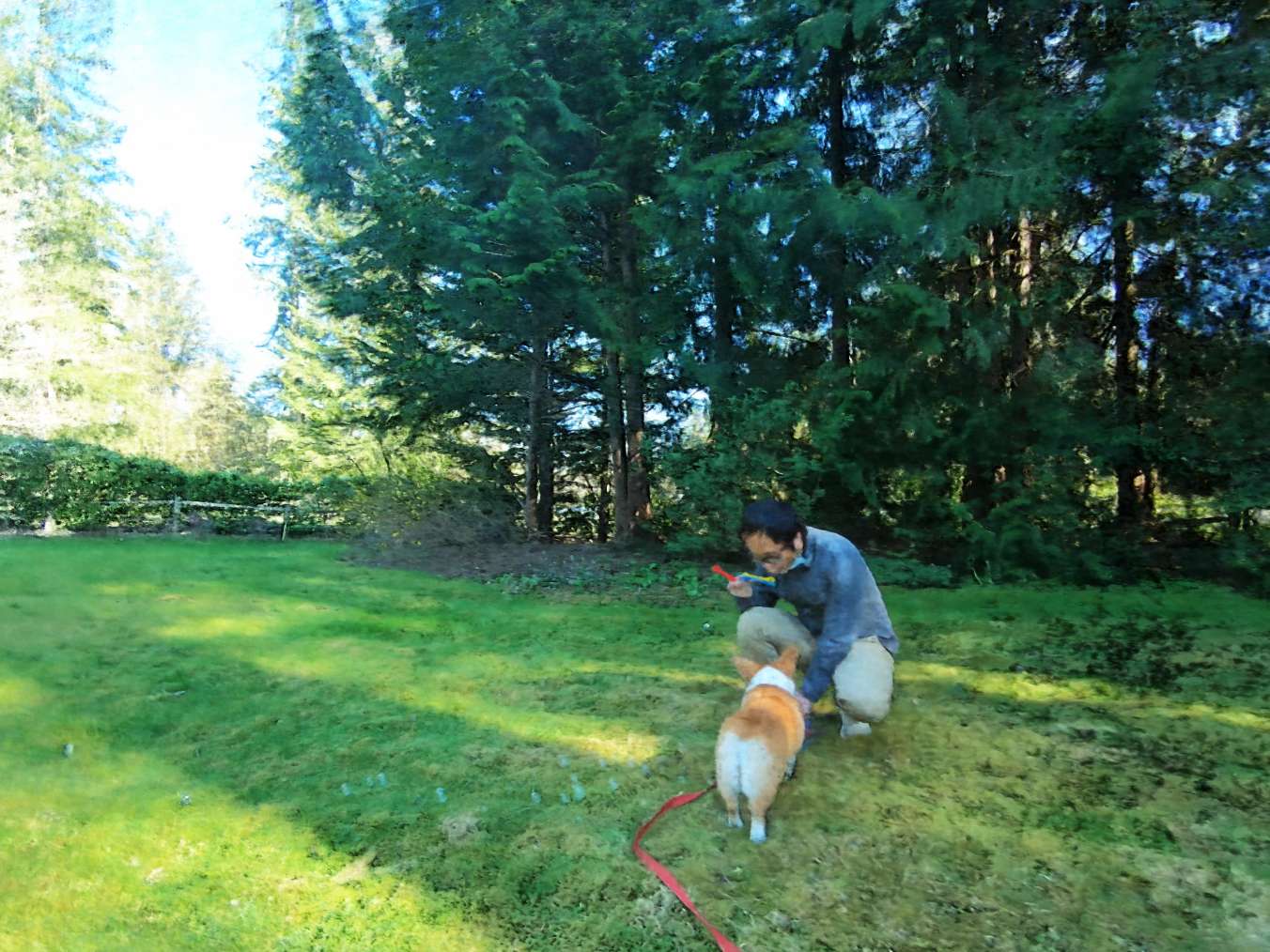}~
\includegraphics[width=0.24\linewidth]{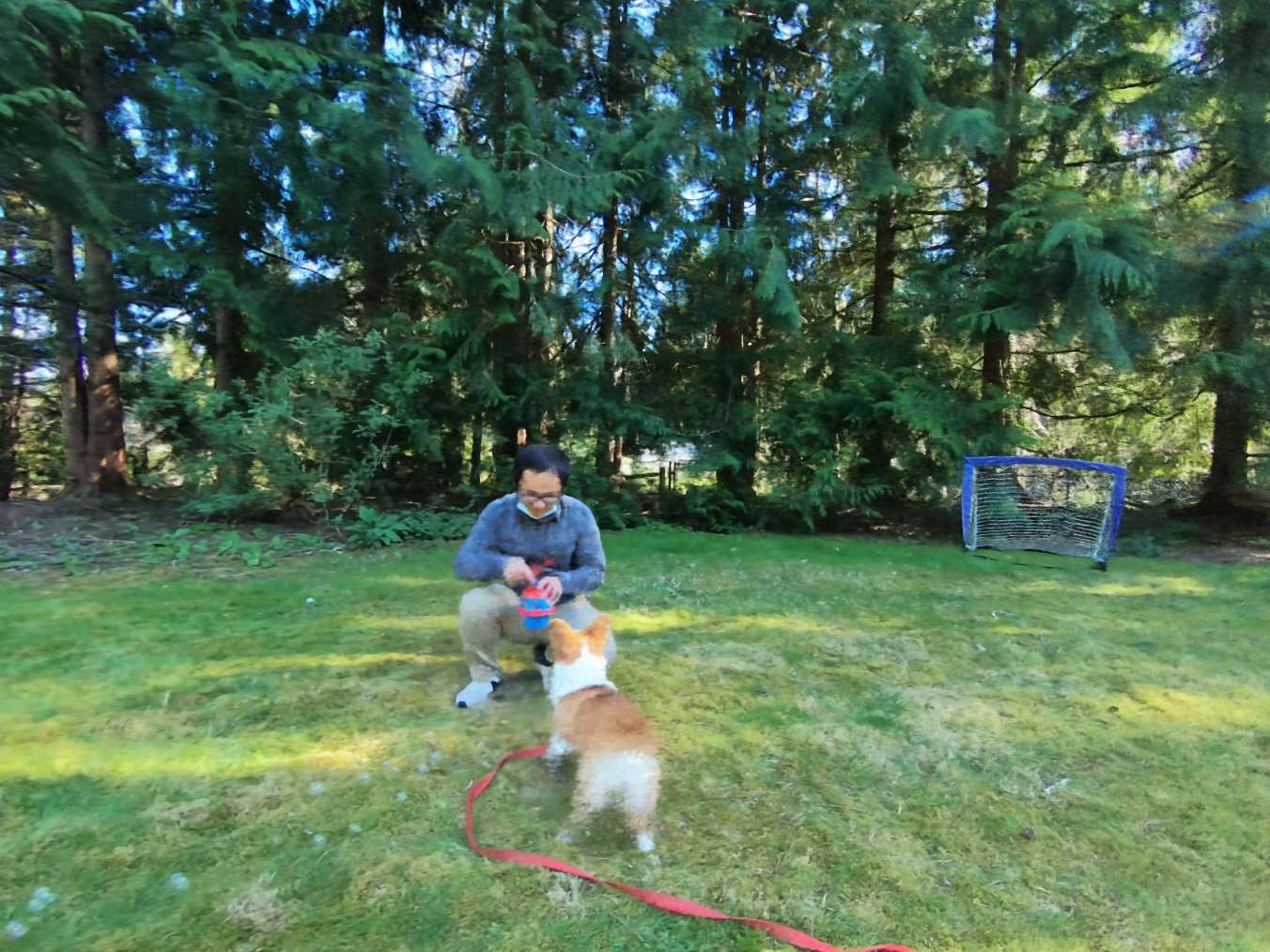}~
\includegraphics[width=0.24\linewidth]{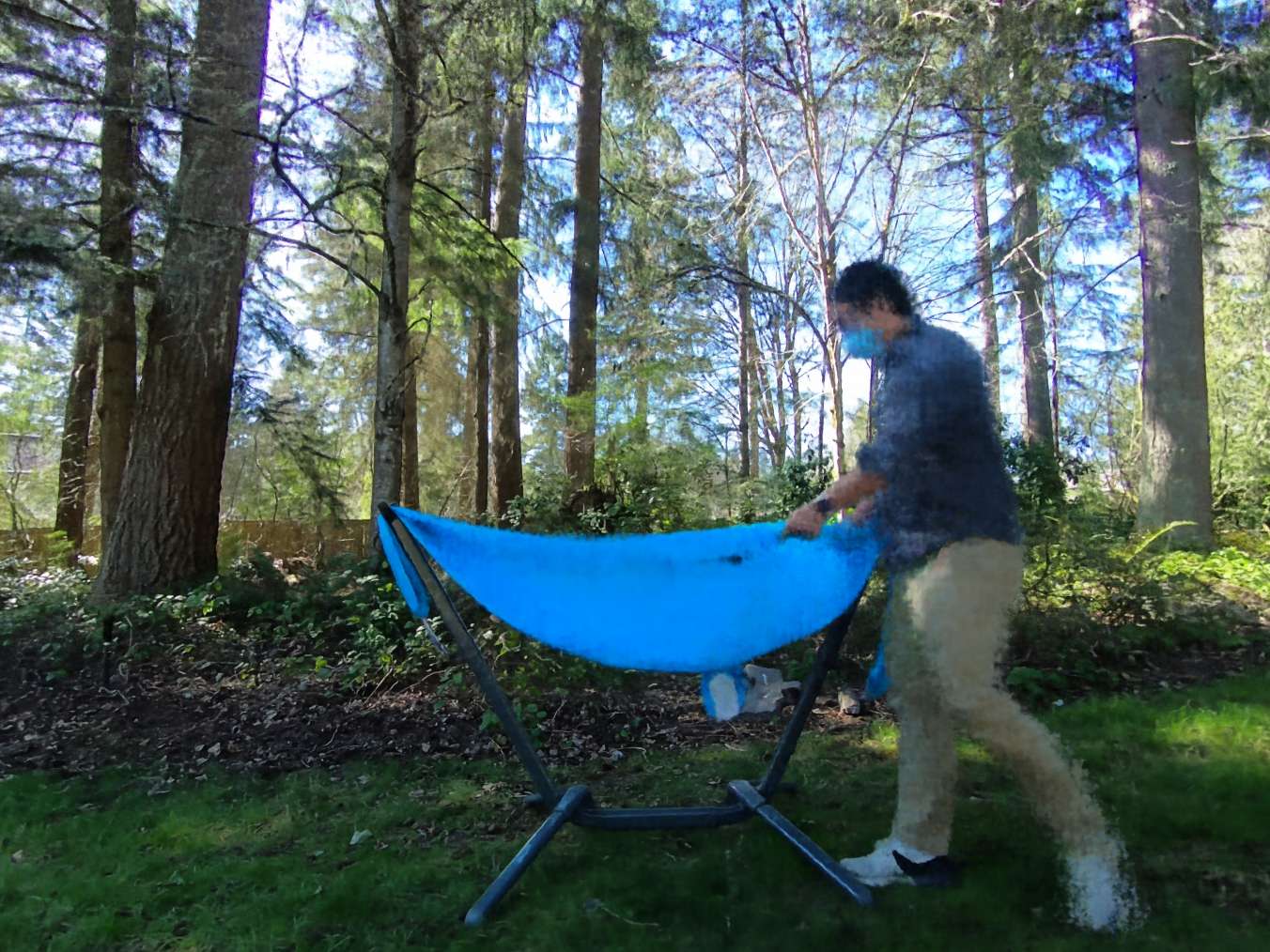}~
\includegraphics[width=0.24\linewidth]{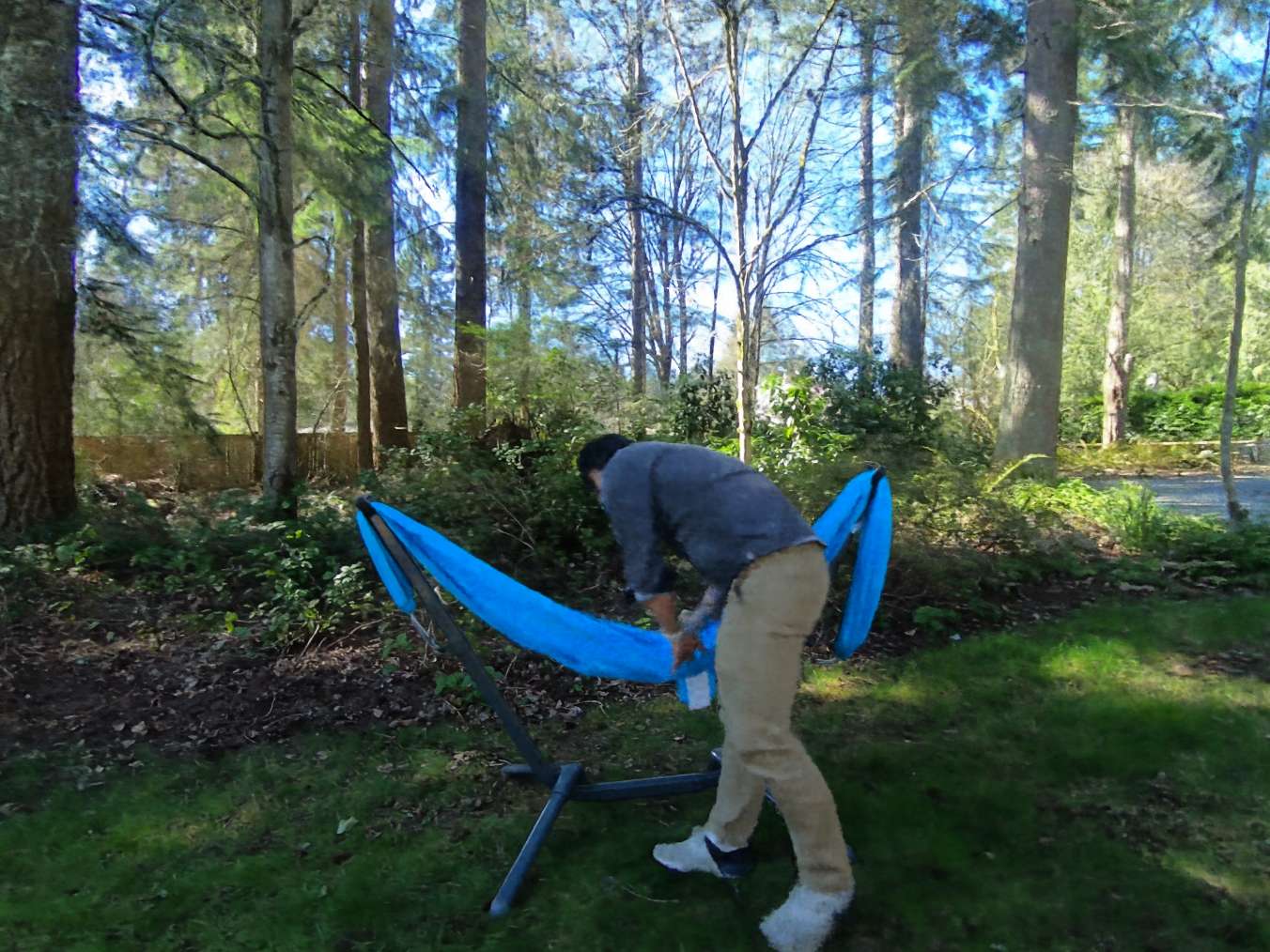}~

\caption{
A few examples of failed outdoor reconstruction using DyNeRF.
}
\label{fig:limitation}

\end{figure}

%% file: sections/conclusion.tex
\section{Conclusion}
We have proposed a novel neural 3D video synthesis approach that is able to represent real-world multi-view video recordings of dynamic scenes in a compact, yet expressive representation.
As we have demonstrated, our approach is able to represent a 10 second long multi-view recording by 18 cameras in under 28MB. 
Our model-free representation enables both high-quality view synthesis as well as motion interpolation.
At the core of our approach is an efficient algorithm to learn dynamic latent-conditioned neural radiance fields that significantly boosts training speed, leads to fast convergence, and enables high quality results.
We see our approach as a first step forward in efficiently training dynamic neural radiance fields and hope that it will inspire follow-up work in the exciting and emerging field of neural scene representations.

%% file: appendix.tex
\section{Supplemental Video}

We strongly recommend the reader to watch our \textit{supplemental video}, hosted at the project website: \url{https://neural-3d-video.github.io/}, to better judge the photorealism of our approach at high resolution, which cannot be represented well by the metrics.
The \textit{supplemental video} includes:
\begin{itemize}[nosep,align=parleft,leftmargin=*]
    \itemsep0em 
    \item 3D video synthesis results on various dynamic scenes including challenging dynamic topology change fast motion, view-dependent effects such as specularity and transparency, varying illuminations and shadows, and volumetric effects such as steam and fire;
    \item A short presentation on the method (the DyNeRF representation and the efficient training method);
    \item Video comparisons to baseline methods: NeRF-T, DyNeRF-noIS, LLFF~\cite{mildenhall2019local}, NeuralVolume~\cite{Lombardi19tog};
    \item Visualization of the estimated geometry (rendered as depth maps);
    \item Slow-motion and bullet-time effects by our DyNeRF;
    \item More results on more challenging indoor scenes;
    \item Results on immersive video datasets \cite{broxton2020siggraph};
    \item Demonstration of interactive playback of our 3D videos in commodity VR headset \textit{Quest 2} using layered meshes distilled from our pretrained DyNeRF model;
    \item Limitation of our results on more challenging outdoor scenes.
\end{itemize}

\section{Datasets}

\paragraph{Details on the Capture Setup.}

\input{figures/fig_captured_data}

\input{figures/fig_camera_rig}

We build a mobile multi-view capture system using 21 GoPro Black Hero 7 cameras, as shown in Fig.~\ref{fig:camera_rig}.
For all results discussed in this paper, we capture videos using the linear camera mode at a resolution of $2028 \times 2704$ (2.7K) and frame rate of \SI{30}{FPS}. 
The multi-view inputs are synchronized by a timecode system, and the camera intrinsic and extrinsic parameters are obtained by COLMAP \cite{schoenberger2016sfm} and are kept the same throughout the capture.

Our collected data can provide sufficient synchronized camera views for high quality 4D reconstruction of challenging dynamic objects and view-dependent effects in a natural daily indoor environment, which did not exist in public 4D datasets.
Our captured data demonstrates a variety of challenges for video synthesis, including objects of high specularity, translucency and transparency.
It also contains scene changes and motions with changing topology (poured liquid), self-cast moving shadows, volumetric effects (fire flame), and an entangled moving object with strong view-dependent effects (the torch gun and the pan), various lighting conditions (daytime, night, spotlight from the side), multiple people moving around in open living room space with outdoor scenes seen through transparent windows with relatively dark indoor illumination.
We visualize one snapshot of the sequence in Fig.~\ref{fig:captured_data_viz}.
Unless otherwise stated, we use keyframes that are 30 frames apart.
In total, we trained our methods on a 60 second video sequence (\emph{flame salmon}) in 6 chunks with each 10 seconds in length, five other 10 seconds cooking videos captured at different time with different motion and lighting, and one 25 seconds video in indoor videos in 5 chunks.
We also trained a few additional videos of outdoor scenes in chunks of 5 seconds with denser keyframes, which are 10 frames apart.
In the end, we employ a subset of 18 camera views for training, and 1 view for quantitative evaluation for all datasets except one sequence observing multiple people moving, which only uses 14 cameras views for training.
We calculate a continuous interpolated spiral trajectory based on the training camera views, which we employ for qualitative novel view evaluation.

We found that the GoPro linear FOV mode sufficiently well compensates for fisheye effects, thus we employ a pinhole camera model for all our experiments.
For all training, we hold out the top center camera for testing, and use the rest of the cameras for training.
For each captured multi-view sequence, we removed a particular camera if the time synchronization did not work.
We also notice there are some inconsistent appearance in the some video streams caused by different lighting sources observed from different view angles, which we excluded in training.

\paragraph{\textbf{Additional Immersive Videos from \cite{broxton2020siggraph}}.} 
We also demonstrate our method using the multi-view captured videos from \cite{broxton2020siggraph} which have been made publicly available recently. 
Due to the time constraints, we train DyNeRF models individually on a few 5s video clips from ``Welder'', ``Flames'', and ``Alexa Meade Face Paint'' to validate our algorithm. 
There are a few differences in their capture setup which pose different opportunities and challenges to our method. 
First, different from our captured linear camera videos which are front-facing, their videos are captured on a half spherical inside-out rig with heavy distortion in each view.
Second, their rig is composed of 46 cameras in each scene, which contains more than two times more numbers of cameras in training. 
Successfully training on this scene using Dynerf requires us to compress a larger dynamic view space and utilizing all training video pixels more efficiently.
During training, we sample the rays directly from the raw resolutions of the distorted multi-view videos and render the novel view video using a pinhole camera.
We demonstrate our algorithm can work on this type of data to create an immersive 3D video experience without any change in the representation.

\section{Importance Sampling Schemes}

\paragraph{\textbf{Sampling Based on Global Median Maps (DyNeRF-ISG)}.}
For each ground truth video, we first calculate the global median value of each ray for all time stamps $\overline{\mathbf{C}}(\mathbf{r}) = \underset{t\in \mathcal{T}}{\mathrm{median}}\;\mathbf{C}^{(t)}(\mathbf{r})$
and cache the global median image. 
During training, we compare each frame to the global median image and compute the residual. 
We choose a robust norm of the residuals to balance the contrast of weight.
The norm measures the transformed values by a non-linear transfer function $\psi(\cdot)$ that is parameterized by $\gamma$ 
to adjust the sensitivity at various ranges of variance:
\begin{equation}
\mathbf{W}^{(t)}(\mathbf{r}) =
\frac{1}{3}
\left\|
\psi \left(
\mathbf{C}^{(t)}(\mathbf{r})
-
\overline{\mathbf{C}}(\mathbf{r});
\gamma
\right)
\right\|_{1} \enspace .
\end{equation}
Here, $\psi(x; \gamma) = \frac{x^2}{x^2 + \gamma^2}$ is the Geman-McLure robust function \cite{geman1985bayesian} applied element-wise.
Intuitively, a larger $\gamma$ will lead to a high probability to sample the time-variant region, and $\gamma$ approaching zero will approximate uniform sampling.
$\overline{\mathbf{C}}(\mathbf{r})$ is a representative image across time, which can also take other forms such as a mean image. We empirically validated that using a median image is more effective to handle high frequency signal of moving regions across time, which helps us to approach sharp results faster during training.

\paragraph{\textbf{Sampling Based on Temporal Difference (DyNeRF-IST)}.}
An alternative strategy, DyNeRF-IST, calculates the residuals by considering two nearby frames in time $t_i$ and $t_j$. 
In each training iteration we load two frames within a 25-frame distance, $|t_i-t_j|\leq 25$. 
In this strategy, we focus on sampling the pixels with largest temporal difference. 
We calculate the residuals between the two frames, averaged over the 3 color channels
\begin{equation}
\mathbf{W}^{(t_i)}(\mathbf{r}) = \min \Big(
\frac{1}{3} \left\| \, \mathbf{C}^{(t_i)}(\mathbf{r}) - \mathbf{C}^{(t_j)}(\mathbf{r}) \right\|_{1}, 
\alpha
\Big)\enspace.
\end{equation}
To ensure that we do not sample pixels whose values changed due to spurious artifacts, we clamp $\mathbf{W}^{(t_i)}(\mathbf{r})$ with a lower-bound $\alpha$, which is a hyper-parameter. 
Intuitively, a small value of $\alpha$ would favor highly dynamic regions, while a large value would assign similar importance to all rays.

\paragraph{\textbf{Combined Method (DyNeRF-IS$^\star\!$}).}
We empirically observed that training DyNeRF-ISG with a high learning rate leads to very quick recovery of dynamic detail, but results in some jitter across time.
On the other hand, training DyNeRF-IST with a low learning rate produces a smooth temporal sequence which is still somewhat blurry.
Thus, we combine the benefits of both methods in our final strategy, DyNeRF-IS$^\star$, which first obtains sharp details via DyNeRF-ISG and then smoothens the temporal motion via DyNeRF-IST.

\input{figures/tab_ablations_all}  %

\paragraph{\textbf{Training Details with the Important Sampling Schemes}.}

We apply global median map importance sampling (DyNeRF-ISG) in both the keyframe training and full video training stage, and subsequently refine with temporal derivative importance sampling only for the full video.
For faster computation in DyNeRF-ISG we calculate temporal median maps and pixel weights for each view at $\frac{1}{4}$th of the resolution, and then upsample the median image map to the input resolution.
For $\gamma$ in the Geman-McClure robust norm, we set $1\mathrm{e}{-3}$ during keyframe training, and $2\mathrm{e}{-2}$ in the full video training stage.
Empirically, this samples the background more densely in the keyframe training stage than for the following full video training. 
We also found out that using importance sampling has a larger impact in the full video training, as keyframes are highly different. 
We set $\alpha = 0.1$ in DyNeRF-IST. 
In the full video training stage we first train for $250K$ iterations of DyNeRF-ISG with learning rate $1\mathrm{e}{-4}$ and then for another $100K$ iterations of DyNeRF-IST with learning rate $1\mathrm{e}{-5}$.

\section{More Results}

\paragraph{\textbf{Details on Baseline Methods}.}

\begin{itemize}[nosep,align=parleft,leftmargin=*]

\item \textbf{Multi-View Stereo (MVS)}:
We reconstruct the textured 3D meshes using commercial photogrammetry software RealityCapture\footnote{\url{https://www.capturingreality.com/}} and render the novel view with from the textured 3D meshes frame-by-frame. This baseline demonstrates the challenges using traditional geometry based approaches.
\item \textbf{Local Light Field Fusion (LLFF)} \cite{mildenhall2019local}:
LLFF is one of the state-of-the-art Multiplane Images based methods tailored to front-facing scenes. We apply the pre-trained network in LLFF to produce the multiplane images and render the novel views using default parameters. To work with videos in our datasets, we produce the novel view frame-by-frame by query the inputs at each corresponding time.
\item \textbf{Neural Volumes (NV)} \cite{Lombardi19tog}:
NV is one of the state-of-the-art learning based volumetric methods can generate novel view videos. We use the same training videos and apply the default parameters to train the network. We set the bounding volume according to the geometry of the scene. We use $128^3$ voxel grid for the RGB$\alpha$ volume and $32^3$ for the warping grid. It renders a novel view image via ray marching a warped voxel grid at each timestamps.
\item \textbf{NeRF-T}:
Refers to the version in Eq.~1. in the main paper, which is a straight-forward temporal extension of NeRF. We implement it following the details in \cite{Mildenhall20eccv}, with only one difference in the input.
The input concatenates the original positionally-encoded location, view direction, and time.
We choose the positionally-encoded bandwidth for the time variable to be $4$ and we do not find that increasing the bandwidth further improves results.
\item \textbf{DyNeRF$^\dagger$}:
We compare to DyNeRF without our proposed hierarchical training strategy and without importance sampling, i.e. this strategy uses per-frame latent codes that are trained jointly from scratch.
\item \textbf{DyNeRF with varying hyper-parameters}:
We vary the dimension of the employed latent codes ($8$, $64$, $256$, $1024$, $8192$). We also apply ablation studies on different versions of DyNeRF with important sampling methods: \textbf{DyNeRF-ISG}, \textbf{DyNeRF-IST}, and \textbf{DyNeRF-IS$^\star\!$}.

\end{itemize}

\paragraph{\textbf{Quantitative Comparison to the Baselines}.}
Tab.~\ref{tab:ablations_methods} shows the quantitative comparison of our methods to the baselines using an average of single frame metrics.
We train all the neural radiance field based baselines and our method the same number of iterations for fair comparison.
Compared to the existing methods, MVS, NeuralVolumes and LLFF, our method is able capture and render significant more photo-realistic images, in all the quantitative measures. 
Compared to the time-variant NeRF baseline NeRF-T and our basic DyNeRF model without our proposed training strategy (DyNeRF$^\dagger$), our DyNeRF model variants trained with our proposed training strategy perform significantly better in all metrics.
DyNeRF-ISG and DyNeRF-IST can both achieve high quantitative performance, with DyNeRF-IST slightly more favorable in terms of the metrics. 
Our complete strategy DyNeRF-IS$^\star$ requires more iterations and is added to the table only for completeness.

\input{figures/tab_model_size}

\input{figures/tab_ablations_all_separate}

\paragraph{\textbf{The Impact of Importance Sampling}.}
In Fig.~\ref{fig:importance_sampling_effect} we evaluate the effect of our importance sampling strategies, DyNeRF-ISG, DyNeRF-IST and DyNeRF-IS$^\star$, against a baseline DyNeRF-noIS that also employs a hierarchical training strategy with latent codes initialized from trained keyframes, but instead of selecting rays based on importance, selects them at random like in standard NeRF~\cite{Mildenhall20eccv}. 
The figure shows zoomed-in crops of the dynamic region for better visibility. 
We clearly see that all the importance sampling strategies manage to recover the moving flame gun better than DyNeRF-noIS in two times less iterations. 
At 100k iterations DyNeRF-ISG and DyNeRF-IST look similar, though they converge differently with DyNeRF-IST being blurrier in early iterations and DyNeRF-ISG managing to recover moving details slightly faster. 
The visualizations of the final results upon convergence in Fig.~\ref{fig:importance_sampling_effect} demonstrate the superior photorealism that DyNeRF-IS$^\star$ achieves, as DyNeRF-noIS remains much blurrier in comparison.
We notice that without importance sampling, the system cannot reach an acceptable visual quality within extended training time, indicating the necessity of the importance sampling scheme.
In Fig.~\ref{fig:ablation_method_variants_full}, we compare various settings of the dynamic neural radiance fields.
NeRF-T can only capture a blurry motion representation, which loses all appearance details in the moving regions and cannot capture view-dependent effects.
Though DyNeRF$^\dagger$ has a similar quantitative performance as NeRF-T, it has significantly improved visual quality in the moving regions compared to NeRF-T, but still struggles to recover the sharp appearance details.
DyNeRF with our proposed training strategy, DyNeRF-ISG, DyNeRF-IST and DyNeRF-IS$^\star$, can recover sharp details in the moving regions, including the torch gun and the flames.

\input{figures/fig_importance_sampling_compare}

\input{figures/fig_ablation_method_variants}

\paragraph{\textbf{Comparisons on of Model Compression}.}
Our model is compact in terms of model size.
In Tab.~\ref{tab:model_size}, we compare our model DyNeRF to the alternatives in terms of storage size.
Compared to the raw videos stored in different images, e.g., PNG or JPEG, our representation is more than two orders of magnitude smaller.
Compared to a highly compact 2D video codec (HEVC), which is used as the default video codec for the GoPro camera, our model is still 50 times smaller.
It is worth noting that these compressed 2D representations do not provide a 6D continuous representation as we do.
Though NeRF is a compact model for a single static frame, representing the whole captured video without dropping frames requires a stack of frame-by-frame reconstructed NeRF networks, which is more than 30 times larger in size compared to our single DyNeRF model.
Compared to the convolutional model used in NeuralVolume, DyNeRF is more compact in size and can represent the dynamic scene with better quality.

\paragraph{\textbf{Impact of Latent Embedding Size on DyNeRF}.}
We run an ablation on latent code length on $60$ continuous frames and present the results in Tab.~\ref{tab:ablations_parameters}.
In this experiment, we do not include keyframe training or importance sampling.
We ran the experiments until $300$K iterations, which is when most models are starting to converge in rendering qualities.
Note that with a code length of 8,192 we cannot fit the same number of samples in the GPU memory as in the other cases, so we report a score from a later iteration when roughly the same number of samples have been used.
We use $4\times$ $16$GB GPUs and network width 256 for the experiments with this short sequence.
From the metrics we clearly conclude that a code of length 8 is insufficient to represent the dynamic scene well.
Moreover, we have visually observed that results with such a short code are typically blurry.
With increasing latent code size, the performance also increases respectively, which however saturates at a dimension of 1024.
A latent code size of 8192 has longer training time per iteration.
Taking the capacity and speed jointly into consideration, we choose $1024$ as our default latent code size for all the sequences in this paper and the supplementary video.

\paragraph{\textbf{Additional Discussions on the Latent Codes}.}
Besides all the above findings, we also observe some failure cases to manipulate the latent codes. Extrapolating the latent codes in time cannot directly create high quality extrapolated views.  
We have extensively investigated latent code optimization with various combinations of parameter learnability for latent codes (keyframe / remaining frames) and the network.
With frozen keyframe latent codes, we observe blurrier results than the all-learnable case. 
Therefore learning both latent codes (keyframe / remaining frames) and the network is necessary for producing sharp and high-quality renderings.

\paragraph{\textbf{View-dependent Effects in Dark Indoor Scenes}.} DyNeRF can represent view-dependent effects as well as motion in one continuous representation. When input cameras streams have slightly different appearance differences in observation, we find DyNeRF will model this difference as part of the view-dependent effects when generating novel views. We can observe this artifact in all of our dark indoor scenes where there are more obvious color inconsistency from wide-angle input video streams. Incorporating more careful color calibration and learning color calibration may address this problem, which we leave for future work.

\paragraph{\textbf{3D Video Editing via Manipulating the Latent Codes}.}
DyNeRF represents a continuous spatial-temporal dynamic scene which supports rendering any view within the interpolation boundary of space and time.
We can create a latent code at a sub-frame time via interpolation and render a ``slow motion'' 3D video with any given FPS rate. 
DyNeRF can enable smooth interpolation from 30fps to 60fps or even 150fps.
Furthermore our method can render ``bullet-time'' effect by by freezing the latent code at any arbitrary time and manipulating camera views in space.
We include the video effects of ``slow motion`` and ``bullet-time`` from arbitrary time in our supplementary videos.

\paragraph{\textbf{Rendering Time}.}
The rendering time of our method is on par with NeRF due to the structural similarity of the approaches.
Our current, not fully optimized version achieves a rendering time of 45 seconds for one 1080p frame using two V-100 GPUs with 16 GB memory.

\balance

%% file: figures/fig_captured_data.tex
\begin{figure*}[t]
\centering
\begin{subfigure}{\linewidth}
\includegraphics[width=\linewidth]{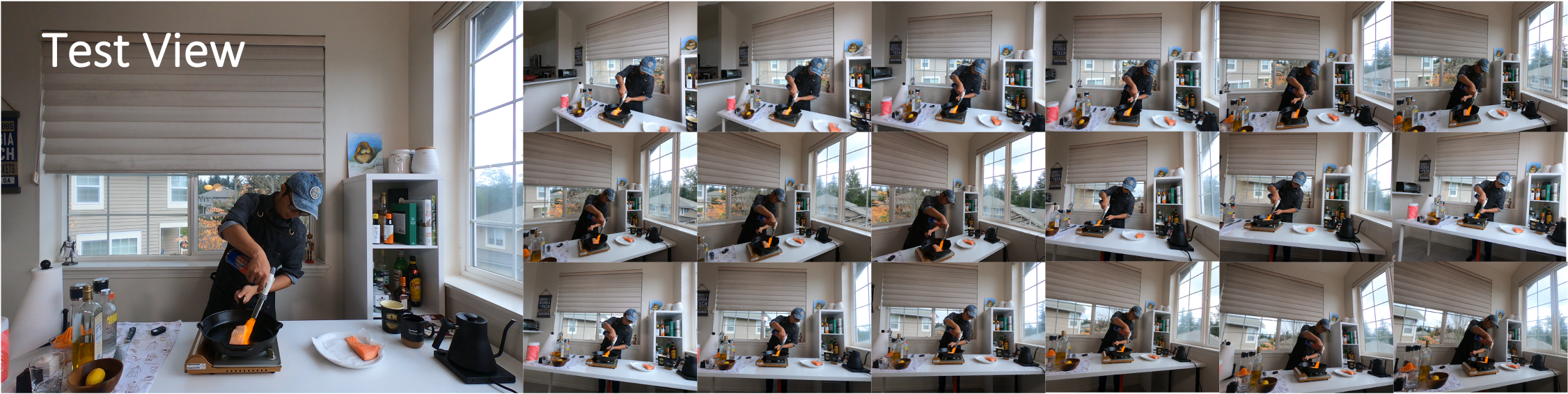}
\end{subfigure}\\
\begin{subfigure}{\linewidth}
    \includegraphics[width=0.19\linewidth]{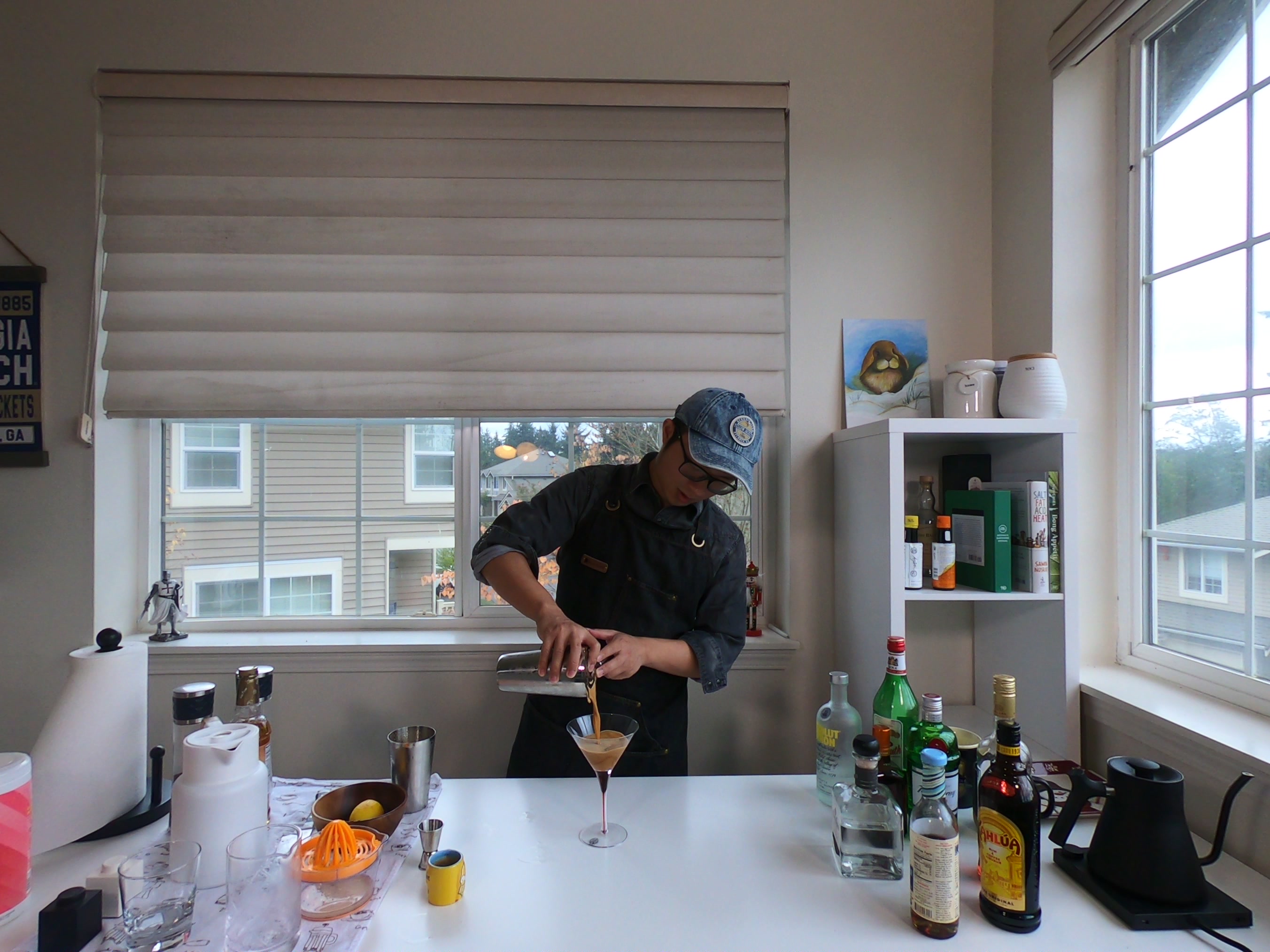}~
    \includegraphics[width=0.19\linewidth]{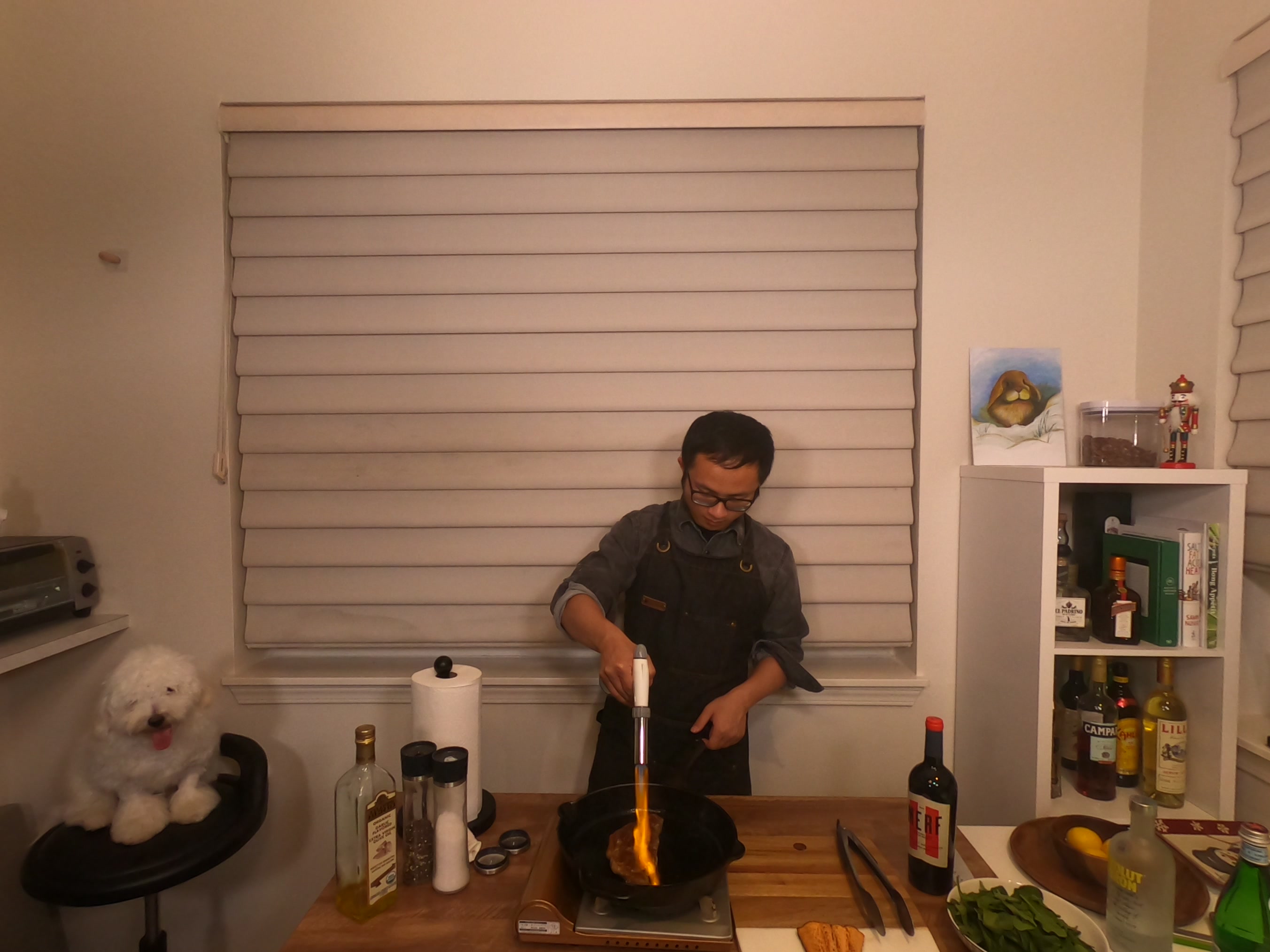}~
    \includegraphics[width=0.19\linewidth]{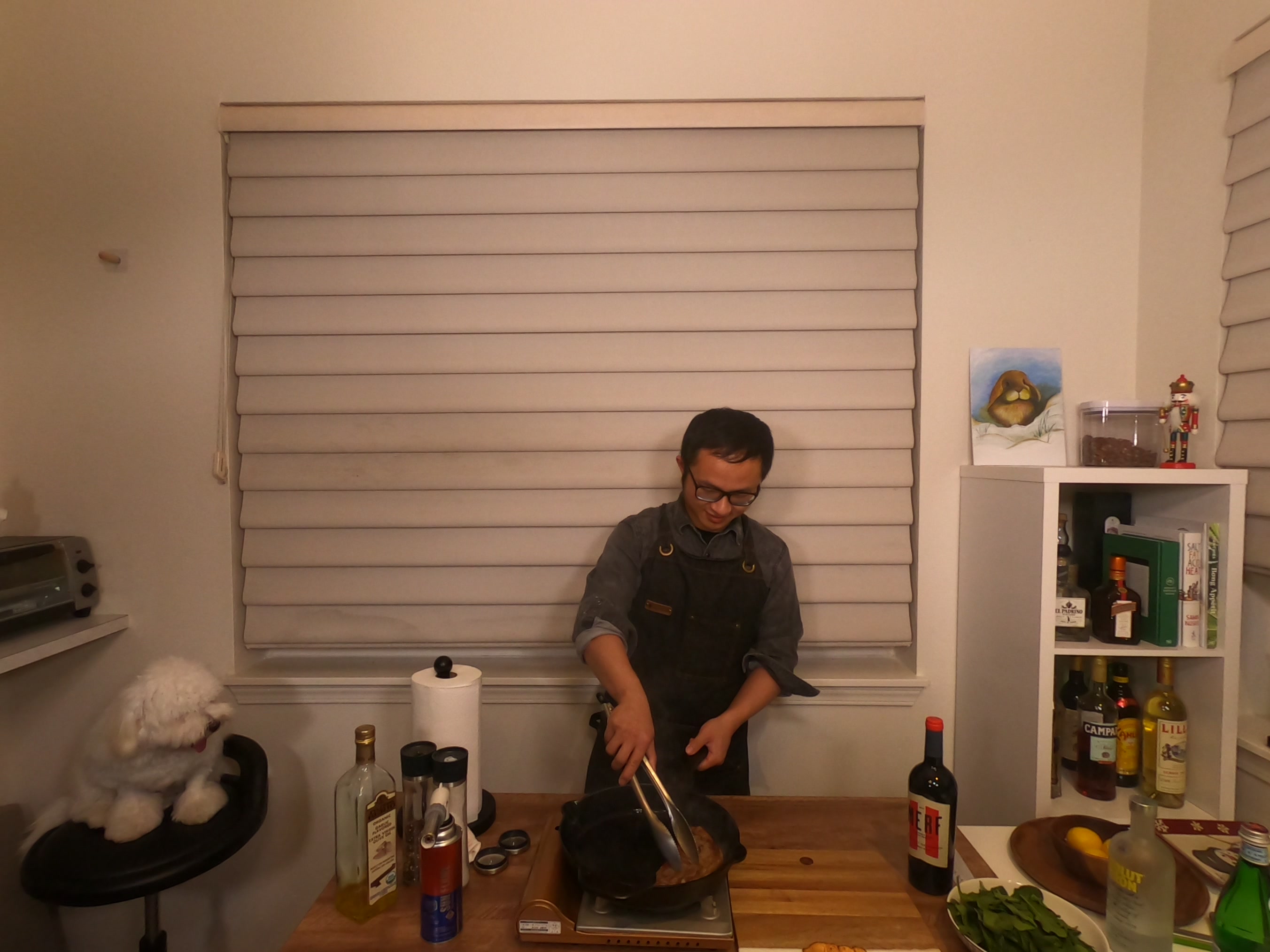}~
    \includegraphics[width=0.19\linewidth]{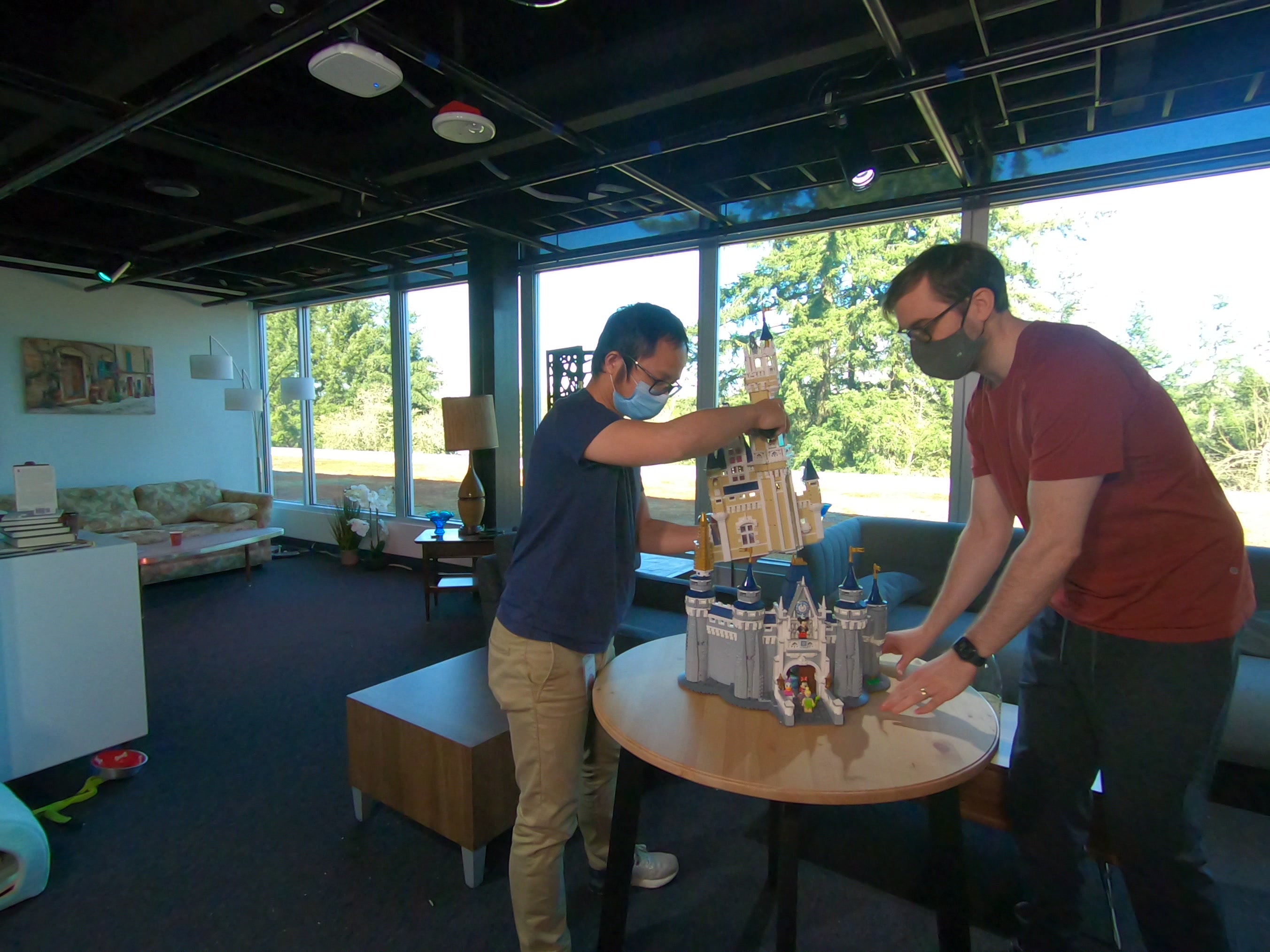}~
    \includegraphics[width=0.19\linewidth]{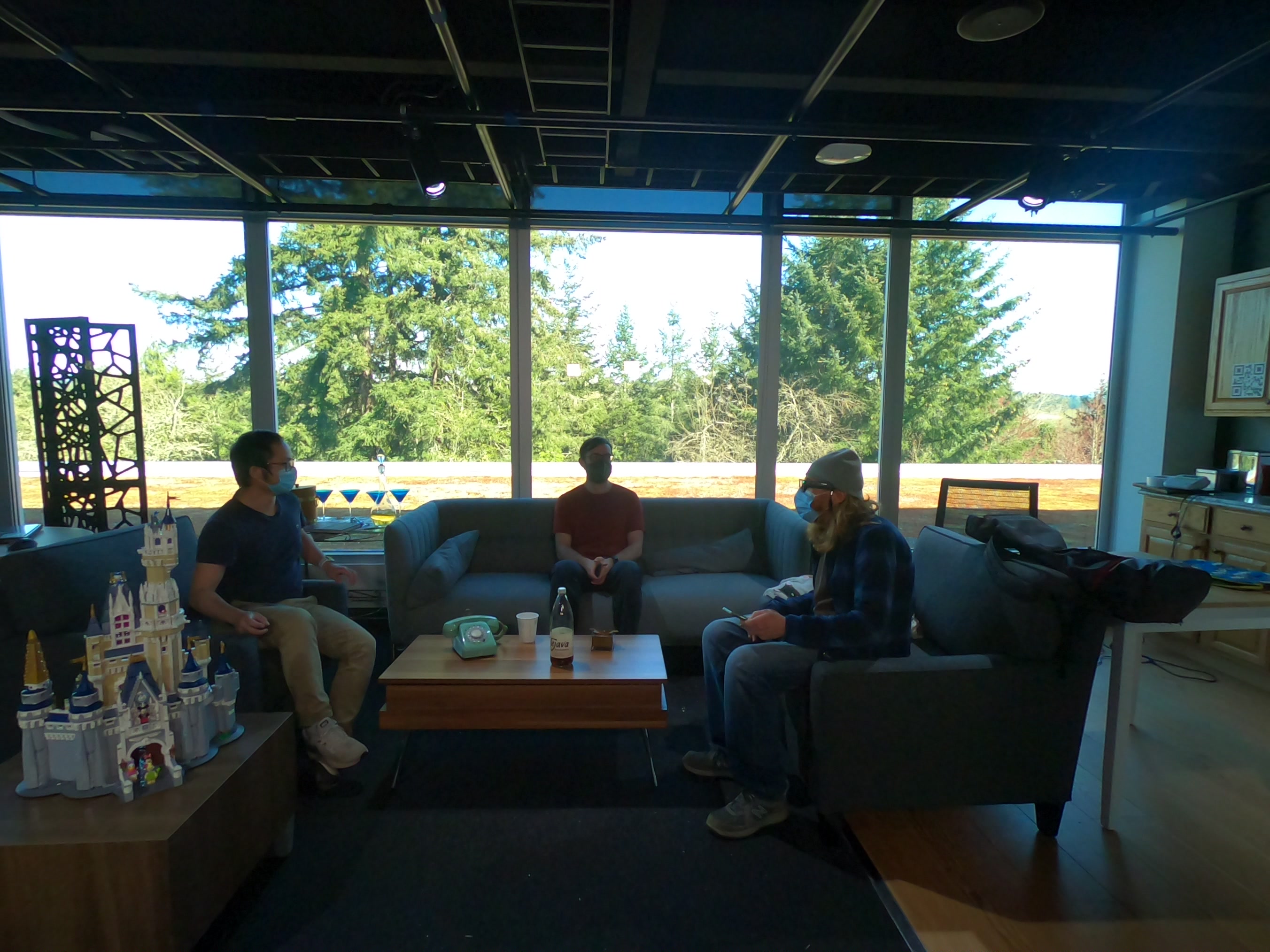}~
\end{subfigure}\\
\vspace{-5pt}
\caption{
\textbf{Frames from our captured multi-view video} \emph{flame salmon} sequence (top). We use 18 camera views for training (downsized on the right), and held out the upper row center view of the rig as novel view for quantitative evaluation.
We captured sequences at different physical locations, time, and under varying illumination conditions.
Our data shows a large variety of challenges in high quality wide angle 3D video synthesis.
}
\label{fig:captured_data_viz}
\end{figure*}

%% file: figures/fig_camera_rig.tex
\begin{figure}[ht]

\centering
\includegraphics[width=1\linewidth]{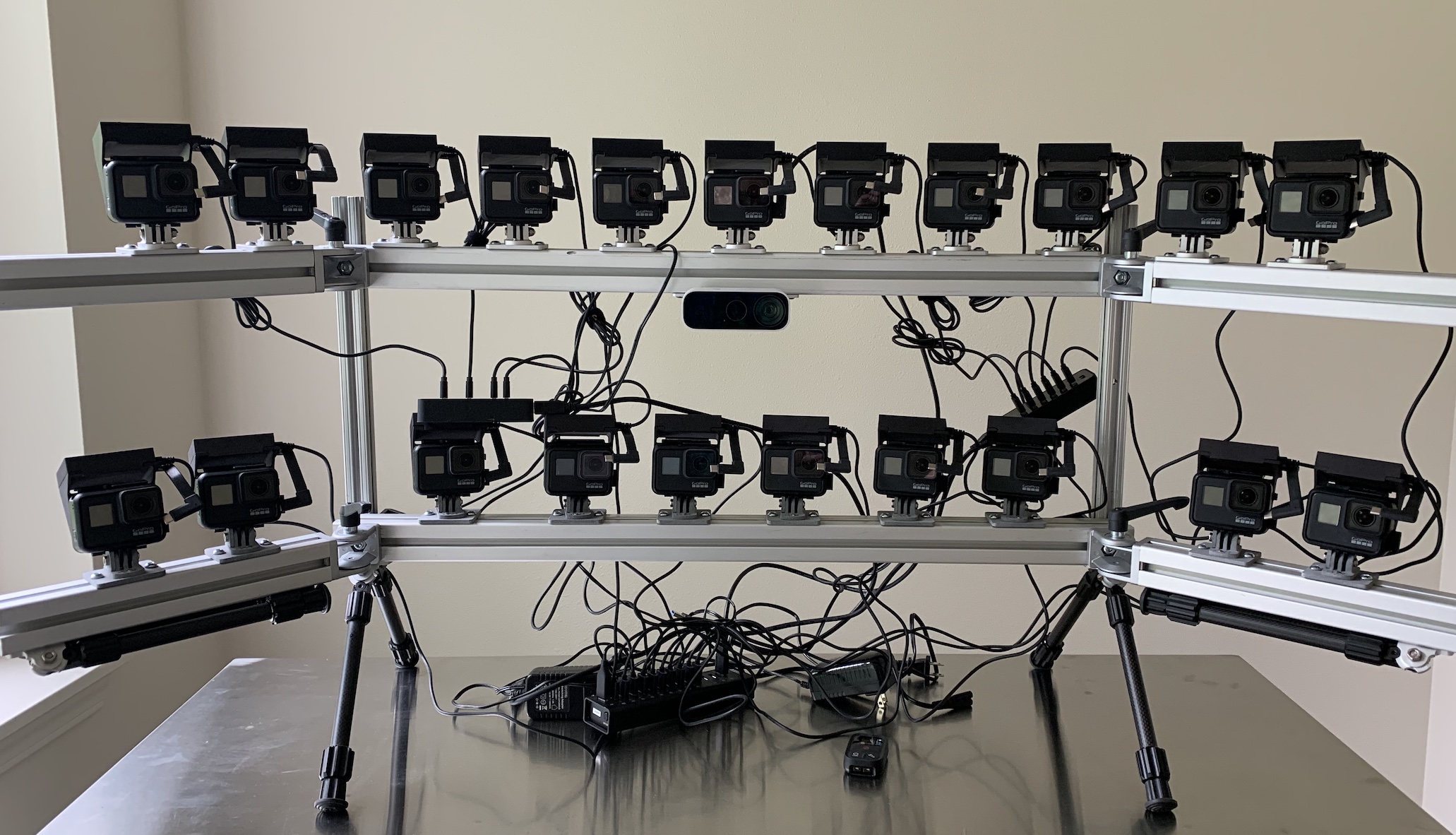}
\caption{\textbf{Our multi-view capture setup} using synchronized GoPro Black Hero 7 cameras.}
\label{fig:camera_rig}

\end{figure}

%% file: figures/tab_ablations_all.tex
\begin{table}[htp!]
\caption{
\textbf{Quantitative comparison} of our proposed method to baselines of existing methods and radiance field baselines trained at 200K iterations on a 10-second sequence. %
DyNeRF-IS$^\star$ uses both sampling strategies (ISG and IST) and thus runs for \emph{more} iterations: 250K iterations of ISG, followed by 100K of IST; it is shown here only for completeness.
}
\resizebox{\linewidth}{!}{
\small
\centering
\begin{tabular}{@{}lccccc@{}}
\toprule
  Method & PSNR $\uparrow$ & MSE $\downarrow$ & DSSIM $\downarrow$ & LPIPS $\downarrow$ & FLIP $\downarrow$ \\ %
 \midrule
MVS & 19.1213 & 0.01226 & 0.1116 & 0.2599 & 0.2542 \\
NeuralVolumes & 22.7975 & 0.00525 & 0.0618 & 0.2951  & 0.2049 \\ %
LLFF  & 23.2388 & 0.00475 & 0.0762 & 0.2346 & 0.1867 \\
NeRF-T & 28.4487 & 0.00144 & 0.0228 & 0.1000 & 0.1415 \\ %
DyNeRF$^\dagger$ & 28.4994 & 0.00143 & 0.0231 & 0.0985 & 0.1455 \\ %
DyNeRF-ISG & 29.4623 & 0.00113 & 0.0201 & 0.0854 & 0.1375 \\ %
DyNeRF-IST & \textbf{29.7161} & \textbf{0.00107} & \textbf{0.0197} & 0.0885 & \textbf{0.1340} \\ %
DyNeRF-IS$^\star$ & 29.5808 & 0.00110 & \textbf{0.0197} & \textbf{0.0832} & 0.1347 \\
\bottomrule
\end{tabular}
}
\label{tab:ablations_methods}
\end{table}

%% file: figures/tab_model_size.tex
\begin{table}[htp!]
\caption{
\textbf{Comparison in model storage size} of our method (DyNeRF) to alternative solutions.
For HEVC, we use the default GoPro 7 video codec.
For JPEG, we employ a compression rate that maintains the highest image quality.
For NeRF, we use a set of the original NeRF networks \cite{Mildenhall20eccv} reconstructed frame by frame.
For HEVC, PNG and JPEG, the required memory may vary within a factor of 3 depending on the video appearance.
For NeuralVolumes (NV), it only accounts the neural network size without counting its dependency on additional input streams.
For NeRF, NeuralVolume and DyNeRF, the required memory is constant. 
All calculation are based on 10 seconds of 30 FPS videos captured by 18 cameras.
}
\resizebox{\linewidth}{!}{
\small
\centering
\begin{tabular}{@{}ccccccc@{}}
\toprule
          & HEVC & PNG & JPEG & NeRF & NV & DyNeRF \\ \midrule
Size (MB) &  1,406  &  21,600  & 3,143 & 1,080 & 773 & \textbf{28} \\ \bottomrule
\end{tabular}
}
\label{tab:model_size}
\end{table}

%% file: figures/tab_ablations_all_separate.tex
\begin{table}[htp!]
\caption{
\textbf{Ablation studies on the latent code dimension} on a sequence of 60 consecutive frames.
Codes of dimension 8 are insufficient to capture sharp details, while codes of dimension 8,192 take too long to be processed by the network.
We use 1,024 for our experiments, which allows for high quality while converging fast.
*Note that with a code length of 8,192 we cannot fit the same number of samples in the GPU memory as in the other cases, so we report a score from a later iteration when roughly the same number of samples have been used.
}
\resizebox{\linewidth}{!}{
\small
\centering
\begin{tabular}{@{}lccccc@{}} 
\toprule 
Dimension & PSNR $\uparrow$ & MSE $\downarrow$ & DSSIM $\downarrow$ & LPIPS $\downarrow$ & FLIP $\downarrow$ \\ %
 \midrule

 8 & 26.4349 & 0.00228 & 0.0438 & 0.2623 & 0.1562 \\ %
 64 & 27.1651 & 0.00193 & 0.0401 & 0.2476 & 0.1653 \\ %
 256 & 27.3823 & 0.00184 & 0.0421 & 0.2669 & \textbf{0.1500} \\ %
 1,024 & \textbf{27.6286} & \textbf{0.00173} & 0.0408 & 0.2528 & 0.1556 \\ %
 8,192* & 27.4100 & 0.00182 & \textbf{0.0348} & \textbf{0.1932} & 0.1616 \\ %
 \bottomrule

\end{tabular}
}
\label{tab:ablations_parameters}
\end{table}

%% file: figures/fig_importance_sampling_compare.tex
\begin{figure*}[h!]
\centering

\includegraphics[width=1\linewidth]{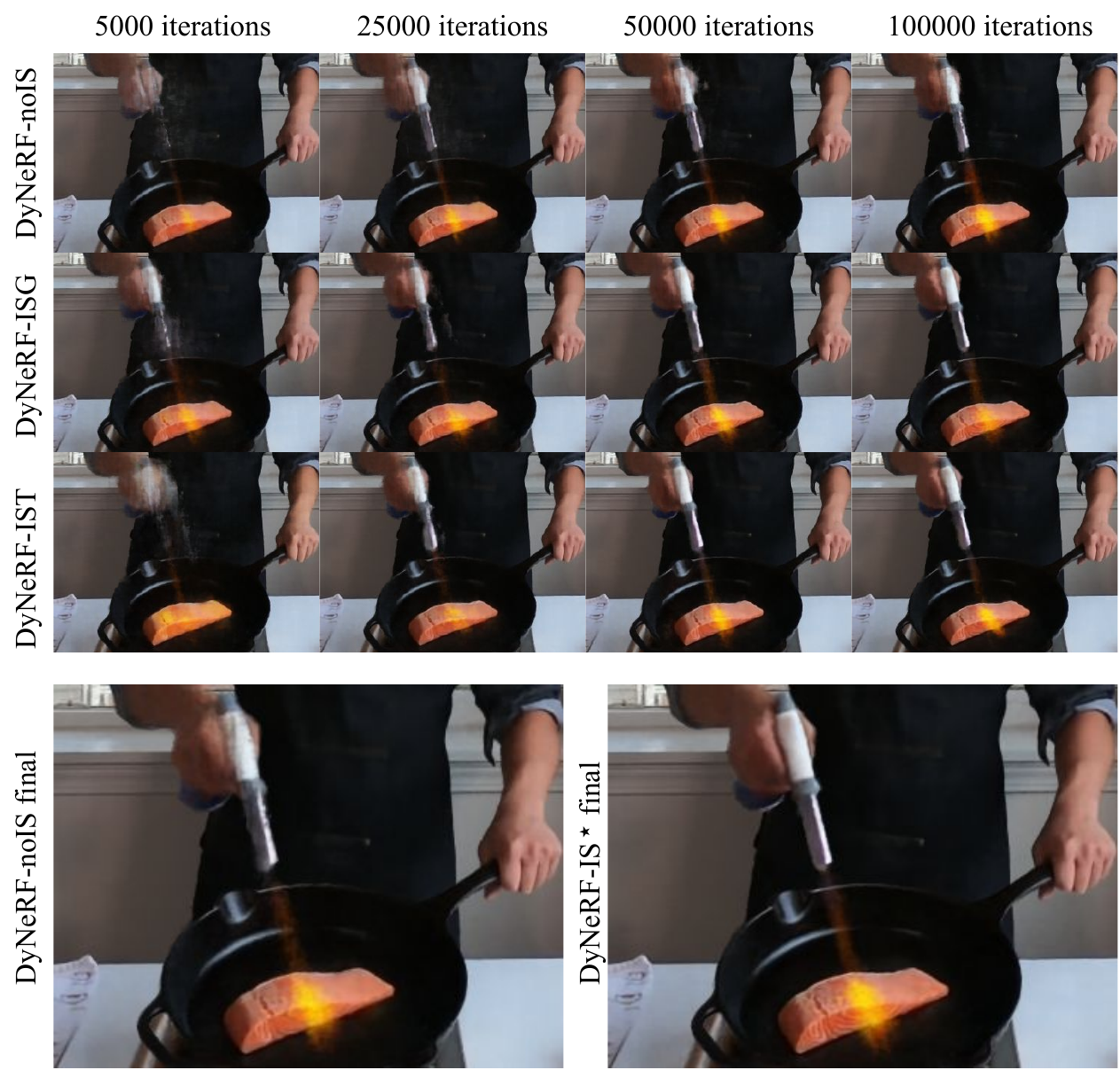}
\vspace{-15pt}
\caption{
\textbf{Comparison of importance sampling strategies over training iterations.}
}
\vspace{-15pt}
\label{fig:importance_sampling_effect}
\end{figure*}

%% file: figures/fig_ablation_method_variants.tex
\begin{figure*}[ht]

\centering

\includegraphics[width=1\linewidth]{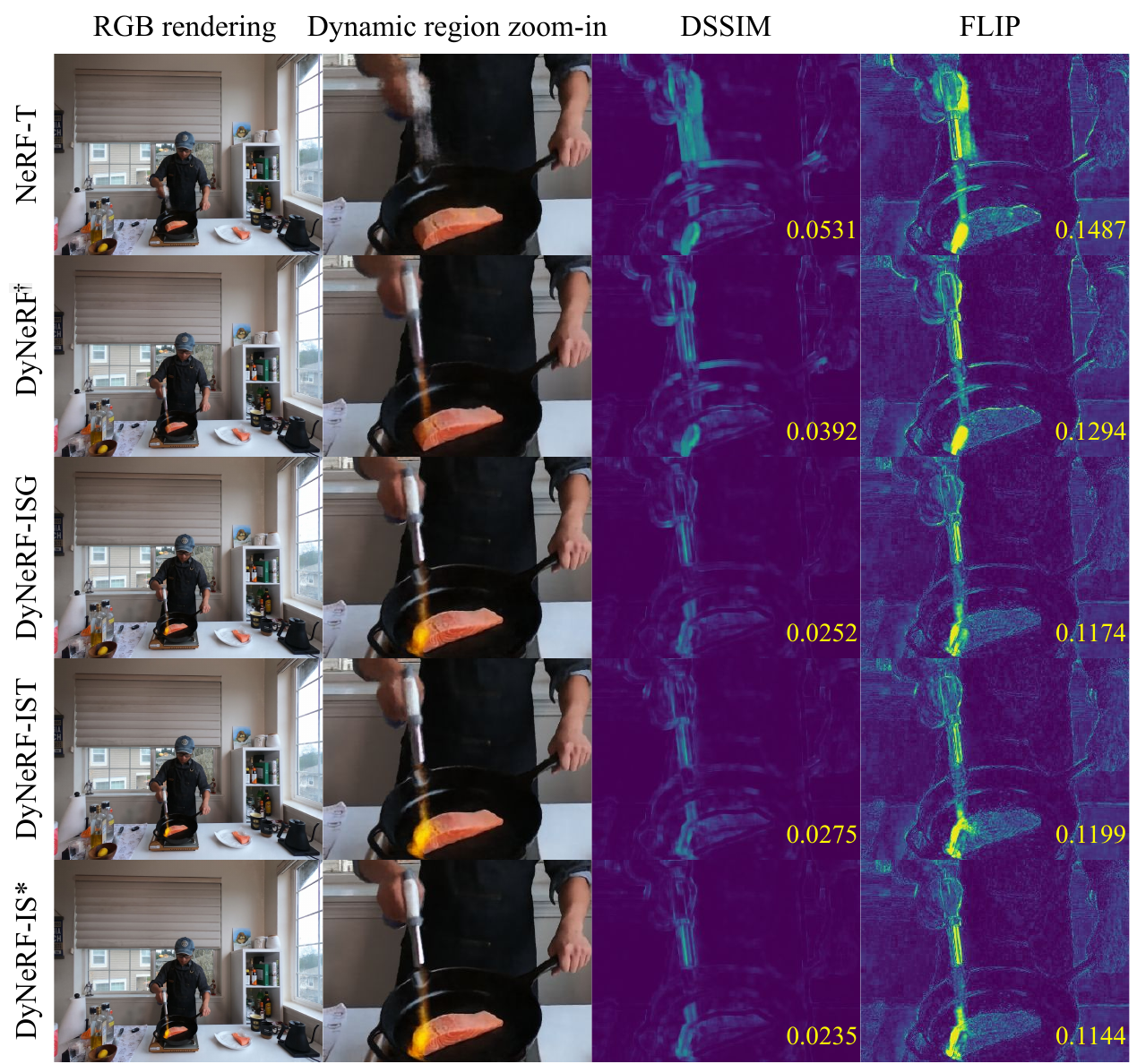}
\vspace{-15pt}
\caption{
\textbf{Qualitative comparisons} of DyNeRF variants on one image of the sequence whose averages are reported in Tab.~\ref{tab:ablations_methods}.
From left to right we show the rendering by each method, then zoom onto the moving flame gun, then visualize DSSIM and FLIP for this region using the \emph{viridis} colormap  (dark blue is 0, yellow is 1, lower is better).
The three hierarchical DyNeRF variants outperform these baselines: DyNeRF-ISG has sharper details than DyNeRF-IST, but DyNeRF-IST recovers more of the flame, while DyNeRF$^\star$ combines both of these benefits.
}
\vspace{-10pt}
\label{fig:ablation_method_variants_full}

\end{figure*}